\documentclass{article}

\usepackage{amsopn}

\usepackage{lipsum}
\usepackage{amsfonts}
\usepackage{graphicx}
\usepackage{epstopdf}
\usepackage{algorithmic}
\ifpdf
  \DeclareGraphicsExtensions{.eps,.pdf,.png,.jpg}
\else
  \DeclareGraphicsExtensions{.eps}
\fi

\usepackage{xcolor}
\definecolor{brightturquoise}{rgb}{0.03, 0.91, 0.87}
\usepackage{hyperref}
\hypersetup{colorlinks,linkcolor={blue},citecolor={green},urlcolor={red}}  
\usepackage{amsmath}
\usepackage{graphicx}
\usepackage{geometry}
\usepackage{lscape}
\usepackage{makecell}
\usepackage{subcaption}
\usepackage{tkz-graph}
\usepackage[export]{adjustbox}
\usepackage{float}

\usepackage{tkz-graph}
\usetikzlibrary{graphs}
\usepackage{tikz-cd} 
\usetikzlibrary{shapes.geometric}

\usepackage[utf8]{inputenc}
\usepackage[english]{babel}

\usepackage{amsthm}

\tikzset{
	basic/.style  = {draw, font=\sffamily, rounded corners=6pt},
	root/.style   = {basic, rounded corners=6pt, thin,align=center, sibling distance=60mm},
	level 1/.style = {basic, rounded corners=6pt, thin,align=center, sibling distance=40mm},
	level 2/.style = {basic, rounded corners=6pt, thin,align=center, sibling distance=40mm},
	level 3/.style = {basic, rounded corners=6pt, thin,align=center, sibling distance=400mm}
	level 4/.style = {basic, rounded corners=6pt, thin,align=center, sibling distance=40mm}
}
\usepackage{enumitem}
\setlist[enumerate]{leftmargin=.5in}
\setlist[itemize]{leftmargin=.5in}

\newtheorem{prop}{Proposition}
\newtheorem{cor}{Corollary}

\theoremstyle{definition}
\newtheorem{definition}{Definition}

\theoremstyle{remark}
\newtheorem{remark}{Remark}

\title{Integer Programming for Learning Directed Acyclic Graphs from Continuous Data}
\author{Hasan Manzour\thanks{Department of Industrial and Systems Engineering, University of Washington
  ({hmanzour@uw.edu}).}
\and Simge K\"u\c{c}\"ukyavuz\thanks{Department of Industrial Engineering and Management Sciences, Northwestern University 
  ({simge@northwestern.edu}).}
\and Ali Shojaie\thanks{Department of Biostatistics, University of Washington
  ({ashojaie@uw.edu}).}}

\newcommand{\ignore}[1]{}

\begin{document}

\maketitle

% REQUIRED
\begin{abstract}
Learning directed acyclic graphs (DAGs) from data is a challenging task both in theory and in practice, because the number of possible DAGs scales superexponentially with the number of nodes. In this paper, we study the problem of learning an optimal DAG from continuous observational data. We cast this problem in the form of a mathematical programming model which can naturally incorporate a super-structure in order to reduce the set of possible candidate DAGs. We use the penalized negative log-likelihood score function with both $\ell_0$ and $\ell_1$ regularizations and propose a new mixed-integer quadratic optimization (MIQO) model, referred to as a layered network (LN) formulation. The LN formulation is a compact model, which enjoys as tight an optimal continuous relaxation value as the stronger but larger formulations under a mild condition. Computational results indicate that the proposed formulation outperforms existing mathematical formulations and scales better than available algorithms that can solve the same problem with only  $\ell_1$ regularization. In particular, the LN formulation clearly outperforms existing methods in terms of computational time needed to find an optimal DAG in the presence of a sparse super-structure. 
\end{abstract}

% REQUIRED
%\begin{keywords}
%  Structural Learning, Directed Acyclic Graph, Integer Programming, Continuous Data
%\end{keywords}

% REQUIRED
%\begin{AMS}
%  68Q25, 68R10, 68U05
%\end{AMS}

\section{Introduction}
The study of Probabilistic Graphical Models (PGMs) is an essential topic in modern artificial intelligence \cite{koller2009probabilistic}. A PGM is a rich framework that represents the joint probability distribution and dependency structure among a set of random variables in the form of a graph. Once learned from data or constructed from expert knowledge, PGMs can be utilized for probabilistic reasoning tasks, such as prediction; see \cite{koller2009probabilistic,lauritzen1996} for  comprehensive reviews of PGMs. 

Two most common classes of PGMs are \textit{Markov networks} (undirected graphical models) and \textit{Bayesian networks} (directed graphical models). A Bayesian Network (BN) is a PGM in which the conditional probability relationships among random variables are represented in the form of a Directed Acyclic Graph (DAG). BNs use the richer language of directed graphs to model probabilistic influence among random variables that have clear directionality; they are particularly popular in practice, with applications in genetics \cite{zhang2013integrated}, biology \cite{markowetz2007inferring}, machine learning \cite{koller2009probabilistic}, and causal inference \cite{spirtes2000causation}.

Learning BNs is a central problem in machine learning. An essential part of this problem entails learning the DAG structure that accurately represents the (hypothetical) joint probability distribution of the BN. Although one can form the DAG based on expert knowledge, acquisition of knowledge from experts is often costly and nontrivial. Hence, there has been considerable interest in learning the DAG directly from observational data \cite{bartlett2017integer, chickering2002optimal, cussens2017polyhedral, cussens2017bayesian, han2016estimation,park2017bayesian, raskutti2013learning, spirtes2000causation}. 

Learning the DAG which best explains observed data is an NP-hard problem \cite{chickering1996learning}.
Despite this negative theoretical result, there has been interest in developing methods for learning DAGs in practice.  There are two main  approaches for learning DAGs from observational data: constraint-based and score-based. In constraint-based methods, such as the well-known PC-Algorithm \cite{spirtes2000causation}, the goal is to learn a completed partially DAG (CPDAG) consistent with conditional independence relations inferred from the data. Score-based methods, including the approach in this paper, aim to find a DAG that maximizes a score that measures how well the DAG fits the data.

Existing DAG learning algorithms can also be divided into methods for discrete and continuous data. Score-based methods for learning DAGs from discrete data typically involve a two-stage learning process. In stage 1, the score for each candidate parent set (CPS) of each node is computed. In stage 2, a search algorithm is used to maximize the global score, so that the resulting graph is acyclic. Both of these stages require exponential computation time \cite{xiang2013lasso}. For stage 2, there exist elegant exact algorithms based on dynamic programming \cite{eaton2007exact, koivisto2012advances, koivisto2004exact, parviainen2009exact, perrier2008finding, silander2012simple, yuan2011learning}, A$^\star$ algorithm \cite{yuan2013learning, yuan2011learning}, and integer-programming \cite{bartlett2013advances, bartlett2017integer,cussens2010maximum,cussens2012bayesian, cussens2017polyhedral,cussens2017bayesian}. The A$^\star$ algorithm identifies the optimal DAG by solving a shortest path problem in an implicit state-space search graph, whereas integer programming (IP) directly casts the problem as a constrained optimization problem. Specifically, the variables in the IP model indicate whether or not a given parent set is assigned to a node in the network. Hence, the problem involves $m2^{m-1}$ binary variables for $m$ nodes. To reduce the number of binary variables, a common practice is to limit the cardinality of each parent set \cite{bartlett2017integer, cussens2017polyhedral}, which can lead to suboptimal solutions.

A comprehensive empirical evaluation of A$^{\star}$ algorithm and IP methods for discrete data {is} conducted in \cite{malone2014predicting}. The results show that the relative efficiency of these methods varies due to the intrinsic differences between them. In particular, state-of-the-art IP methods can solve instances up to 1,000 CPS per variable regardless of the number of nodes, whereas A$^{\star}$ algorithm works for problems with up to 30 nodes, even with tens of thousands of CPS per node.  

While statistical properties of DAG learning from continuous data have been extensively studied \cite{ghoshal2016information, loh2014high, raskutti2013learning, solus2017consistency, van2013ell}, the development of efficient computational tools for learning the optimal DAG from continuous data remains an open challenge. In addition, despite the existence of elegant exact search algorithms for discrete data, the literature on DAG learning from continuous data has primarily focused on approximate algorithms based on coordinate descent \cite{fu2013learning, han2016estimation} and non-convex continuous optimization \cite{zheng2018dags}.
To our knowledge, \cite{park2017bayesian} and \cite{xiang2013lasso} provide the only exact algorithms for learning medium to large DAGs from continuous data. An IP-based model using the topological ordering of variables is proposed in \cite{park2017bayesian}. An A$^{\star}$-lasso algorithm for learning an optimal DAG from continuous data with an $\ell_1$ regularization is developed in \cite{xiang2013lasso}. A$^{\star}$-lasso incorporates the lasso-based scoring method within dynamic programming to avoid an exponential number of parent sets and uses the A$^{\star}$ algorithm to prune the search space of the dynamic programming method. %Also, related is the sparsest permutation algorithm of  \cite{raskutti2013learning}; however, this algorithm is computationally intensive and does not scale to larger problems. \hm{I suggest to remove the last paragraph. I feel we need to emphasize TO and A*-lasso here as we compared with them.}

Given the state of existing algorithms for DAG learning from continuous data, there is currently a gap between theory and computation: While statistical properties of exact algorithms can be rigorously analyzed \cite{loh2014high, van2013ell}, it is much harder to assess the statistical properties of approximate algorithms \cite{aragam2015concave, fu2013learning, han2016estimation, zheng2018dags} that offer no optimality guarantees \cite{koivisto2012advances}. 
This gap becomes particularly noticeable in cases where the statistical model is identifiable from observational data. In this case,  the optimal score from exact search algorithms is guaranteed to reveal the true underlying DAG when the sample size is large. Therefore, causal structure learning from observational data becomes feasible \cite{loh2014high, peters2013identifiability}. 

In this paper, we focus on DAG learning for an important class of BNs for continuous data, where causal relations among random variables are linear. More specifically, we consider DAGs corresponding to linear structural equations models (SEMs). In this case, network edges are associated with the coefficients of regression models corresponding to linear SEMs. Consequently, the score function can be explicitly encoded as a penalized negative log-likelihood function with an appropriate choice of regularization \cite{park2017bayesian, shojaie2010penalized, van2013ell}. Hence, the process of computing scores (i.e., stage 1) is completely bypassed, and a single-stage model can be formulated \cite{xiang2013lasso}. Moreover, in this case, IP formulations only require a polynomial, rather than exponential, number of variables, because each variable can be defined in the space of arcs instead of parent sets. Therefore, cardinality constraints on the size of parent sets, which are used in earlier methods to reduce the search space and may lead to suboptimal solutions, are no longer necessary.  

\paragraph{Contributions} In this paper, we develop tailored exact DAG learning methods for continuous data from linear SEMs,  and make the following contributions: 
\begin{itemize}
	\item[--] We develop a mathematical framework that can naturally incorporate prior structural knowledge, when available. Prior structural knowledge can be supplied in the form of an undirected and possibly cyclic graph (super-structure). An example is the skeleton of the DAG, obtained by removing the direction of all  edges in a graph. Another example is the moral graph of the DAG, obtained by adding an edge between pairs of nodes with common children and removing the direction of all  edges \cite{spirtes2000causation}. The skeleton and moral graphs are particularly important cases, because they can be consistently estimated from observational data under proper assumptions \cite{kalisch2007estimating, loh2014high}. Such prior information limits the number of possible DAGs and improves the computational performance. %\hm{I think we can say " ... can be consistency estimated from observational data \textbf{under proper conditions}"}       
	\item[--] We discuss three mathematical formulations, namely, cutting plane (CP), linear ordering (LO), and topological ordering (TO) formulations, for learning an optimal DAG, using both $\ell_0$ and $\ell_1$-penalized likelihood scores. We also propose a new mathematical formulation to learn an optimal DAG from continuous data, the \textit{layered network} (LN) formulation, and establish that other DAG learning formulations entail a smaller continuous relaxation feasible region compared to that of the continuous relaxation of the LN formulation (Propositions \ref{Prop3: LNLO} and \ref{Prop3: LNTO}). Nonetheless, all formulations attain the same optimal continuous relaxation objective function value under a mild condition (Propositions \ref{Prop4: Root} and \ref{Prop5: BB}). 
Notably, the number of binary variables and constraints in the LN formulation  solely depend on the number of edges in the super-structure (e.g., moral graph). Thus, the performance of the LN formulation substantially improves in the presence of a sparse super-structure. The LN formulation has a number of other advantages; it is a compact formulation in contrast to the CP formulation; its relaxation can be solved much more efficiently compared with the LO formulation; and it requires  fewer  binary variables and explores fewer branch-and-bound nodes  than the TO formulation. Our empirical results affirm the computational advantages of the LN formulation. They also demonstrate that the LN formulation can find a graph closer to the true underlying DAG. These improvements become more noticeable in the presence of a prior super-structure (e.g., moral graph).  %for the same DAG are symmetric solutions to the associated IP. Consequently, B\&B requires exploring multiple symmetric formulations as it branches on fractional TO variables (in contrast to TO) .  . %This suggests that the LN formulation is computationally more efficient because its LP relaxation is either easier to solve due to fewer number of constraints while it still gives as tight lower bound as an LP relaxation of other models. 
	\item[--] We compare the IP-based method and the A$^{\star}$-lasso algorithm for the case of $\ell_1$ regularization. As noted earlier, there is no clear winner among A$^\star$-style algorithms and IP-based models for DAG learning from discrete data \cite{malone2014predicting}. Thus, a wide range of approaches based on dynamic programming, A$^{\star}$ algorithm, and IP-based models have been proposed for discrete data. In contrast, for DAG learning from continuous data with complete super-structure, the LN formulation remains competitive with the state-of-art A$^{\star}$-lasso algorithm for small graphs, whereas it performs better for larger problems.\ Moreover, {LN performs} substantially better when a sparse super-structure is available. %There are three reasons for this advantage: 
	%(i) an IP-based model for DAG learning from continuous data directly defines the variables on the space of arcs. Therefore, it does not require an exponential number of variables (in contrast to IP-based models for discrete data). Thus, the hurdle of the exponential number of CPS is rectified by an explicit objective function; 
	%(ii) 
	This is mainly because the LN formulation directly defines the variables based on the super-structure, whereas the A$^{\star}$-lasso algorithm cannot take advantage of the prior structural knowledge as effectively as the LN formulation. %IP-based models. This is particularly important in the case of LN formulation because the number of constraints and variables depend on the super-structure
	%; and (iii) algorithmic advances in integer optimization in conjunction with hardware improvements have resulted in a 200 billion factor speedup in solving IP problems \citep{bertsimas2016best}.  
%	\hm{I think we can remove points (i) and (iii) and only mention (ii) here. I discussed all points in more detail when comparing A*-lasso and LN.} 

\end{itemize}

In Section~\ref{Sec: SEMs}, we outline the necessary preliminaries, define the DAG structure learning problem, and present a general framework for the problem. Mathematical formulations for DAG learning are presented in Section \ref{Sec: Math models}. The strengths of different optimization problems are discussed in Section \ref{Sec: LP}. Empirical results are presented in  Sections \ref{Sec: Computational} and \ref{sec:simAstar}. We end the paper with a summary and discussion of future research in Section \ref{Sec: Conclusion}.

%%%%%%%%%%%%%%%%%%%%%%%%%%%%%%%%%%%
\section{Penalized DAG Estimation with Linear SEMs} \label{Sec: SEMs}
%%%%%%%%%%%%%%%%%%%%%%%%%%%%%%%%%%%
The causal effect of (continuous) random variables in a DAG $\mathcal{G}_0$ can be described by SEMs that represent each variable as a (nonlinear) function of its parents. The general form of these models is given by \cite{pearl2009causal}
\begin{equation} 
X_k = f_k\big(pa^{\mathcal{G}_0}_k , \delta_k \big), \quad  k = 1, \dots , m, \label{NSLM}
\end{equation}
where $X_k$ is the random variable associated with node $k$; $pa^{\mathcal{G}_0}_k$ denotes the parents of node $k$ in $\mathcal{G}_0$, i.e., the set of nodes with arcs pointing to node $k$;  $m$ is the number of nodes; and latent random variables, $\delta_k$ represent the unexplained variation in each node. 

An important class of SEMs is defined by linear functions, $f_k(\cdot)$, which can be described by $m$ linear regressions of the form 
\begin{equation} \label{LSLM}
X_k = \sum_{j \in pa^{\mathcal{G}_0}_k} \beta_{jk} X_j + \delta_k, \quad k=1,\dots, m,
\end{equation}
where $\beta_{jk}$ represents the effect of node $j$ on $k$ for $j \in pa^{\mathcal{G}_0}_k$.
%Let $\sigma^2_k$ denote the variance of $\delta_k$. 
In the special case where the random variables are Gaussian, Equations~\eqref{NSLM} and \eqref{LSLM} are equivalent, in the sense that $\beta_{jk}$ are coefficients of the linear regression model of $X_k$ on $X_j$, $j \in pa^{\mathcal{G}_0}_k$, and $\beta_{jk} = 0$ for $j \notin pa^{\mathcal{G}_0}_k$ \cite{shojaie2010penalized}. 
However, estimation procedures proposed in this paper are not limited to Gaussian random variables and apply more generally to linear SEMs \cite{loh2014high, shojaie2010penalized}. 

Let $\mathcal{M} = (V, E)$ be an undirected and possibly cyclic super-structure graph with node set $V=\{1,2,\dots,m\}$ and edge set $E \subseteq V \times V$. 
From $\mathcal{M}$, generate a bi-directional graph $\overrightarrow{\mathcal{M}} = (V, \overrightarrow{E})$ where $\overrightarrow{E} =\{(j,k), (k,j) | (j,k) \in E\}$.
Throughout the paper, we refer to directed edges as \emph{arcs} and to undirected edges as \emph{edges}. 

Consider $n$ i.i.d.\ observations from the linear SEM \eqref{LSLM}. Let $\mathcal{X}=(\mathcal{X}_1, \dots , \mathcal{X}_m)$ be the $n \times m$ data matrix with $n$ rows representing i.i.d.\ samples, and $m$ columns representing random variables. 
The linear SEM \eqref{LSLM} can be compactly written as 
\begin{equation} 
\mathcal{X} = \mathcal{X}{B} + \Delta,  \label{Matrix SEMs} 
\end{equation} 
where ${B} = [\beta] \in {\mathbb R}^{m \times m}$ is a matrix with $\beta_{kk}=0$ for $k=1,\dots,m$ and $\beta_{jk}=0$ for all $(j,k) \notin \overrightarrow{E}$; $\Delta$ is the $n\times m$ noise matrix.  More generally, $B$ defines a directed graph $\mathcal{G}(B)$ on $m$ nodes such that arc $(j,k)$ appears in  $\mathcal{G}(B)$ if and only if $\beta_{jk} \neq 0$. 

Let $\sigma^2_k$ denote the variance of $\delta_k; k ={1,2,, \dots, m}$. We assume that all noise variables have equal variances, i.e., $\sigma_k=\sigma$. This condition implies the identifiability of DAGs from Gaussian \cite{peters2013identifiability} and non-Gaussian \cite{loh2014high} data. Under this condition, the negative log likelihood for linear SEMs with Gaussian or non-Gaussian noise is proportional to %\as{do we have identifiability for non-Gaussian too?} \hm{Yes, \cite{loh2014high} state that "As a side corollary, we establish that the DAG structure of a linear SEM is identifiable when-ever the additive errors are homoscedastic, which generalizes a recent result derived only for Gaussian variables (Peters and Buhlmann,2013)." I think we need to slightly revise this paragraph. }
\begin{equation}\label{eqn:lklhd}
	l_n(\beta) =\frac{1}{2}\text{tr}\{(I-{B})(I-{B})^TS\},
\end{equation}
where $S = \mathcal{X}^T\mathcal{X}$ and $I$ is the identity matrix \cite{loh2014high}. 

In practice, we are often interested in learning \textit{sparse} DAGs. Thus, a regularization term is used to obtain a sparse estimate. For the linear SEM \eqref{LSLM}, the optimization problem corresponding to the penalized negative log-likelihood with super-structure $\mathcal{M}$ (PNL$\mathcal{M}$) for learning sparse DAGs is given by
\begin{subequations} \label{eq:PNLMform}
	\begin{align}
	\textbf{PNL$\mathcal{M}$} \quad	&\underset{B \in {\mathbb {R}}^{m \times m}}{\min} \quad l_n(\beta) + \lambda \phi({B}) \label{Eq: Opt}, \\
	\text{s.t.,} \, \, & \mathcal{G}(B) \, \, \text{induces a DAG from} \, {\overrightarrow{\mathcal{M}}}. \label{Eq: DAG const}
	\end{align}
\end{subequations}  
The objective function \eqref{Eq: Opt} consists of two parts: the quadratic loss function, $l_n(\beta)$, from \eqref{eqn:lklhd}, and the regularization term, $\phi({B})$. Popular choices for $\phi(B)$ include $\ell_1$-regularization or lasso \cite{tibshirani1996regression}, $\phi({B})= \sum_{(j,k) \in E^{\rightarrow}} |\beta_{jk}|$, and $\ell_0$-regularization, $\phi({B}) = \sum_{(j,k) \in \overrightarrow{E}} \mathbf 1(\beta_{jk})$, where $\mathbf 1(\beta_{jk})=1$ if $\beta_{jk}\ne 0$, and 0 otherwise. 
%\sum_{(j,k) \in \overrightarrow{E}} \textbf{{1}}_{\{\beta_{jk} \neq 0\}}$, where the function $\textbf{1}_{\{x \neq 0\}}$ takes value 1 if $x \neq 0$ and 0 otherwise. %The objective function is convex with an $\ell_1$ penalty and is non-convex with an $\ell_0$ penalty. 
The tuning parameter $\lambda$ controls the degree of regularization. 
The constraint \eqref{Eq: DAG const} stipulates that the resulting directed subgraph (digraph) has to be an induced DAG from $\overrightarrow{\mathcal{M}}$. 

When the super-structure $\mathcal{M}$ is a complete graph, PNL$\mathcal{M}$ reduces to the classical PNL. In this case, the  consistency of \textit{sparse} PNL for DAG learning from Gaussian data with an $\ell_0$ penalty follows from an analysis similar to \cite{van2013ell}. In particular, we have
\[
%	\sum_{j=1}^{m}\sum_{k=1}^{m} [\hat{\beta}_{jk}(\lambda_n)-{\beta}_{jk}^{0}]^2 = O_p(\log(n)n^{-1}) 
%\quad (n \rightarrow \infty), \quad $$ 
pr({\hat{\mathcal{G}}_n= \mathcal{G}^0}) \rightarrow 1 \quad (n \rightarrow \infty),
\]
%where $\hat{\beta}_{jk}$ and 
where $\hat{\mathcal{G}}_n$ is the estimate of the true structure ${\mathcal{G}^0}$. An important advantage of the PNL estimation problem is that its consistency does not require the (strong) faithfulness assumption \cite{peters2013identifiability, van2013ell}. 

The mathematical model \eqref{eq:PNLMform} incorporates a super-structure $\mathcal{M}$ (e.g., moral graph) into the PNL model. When $\mathcal{M}$ is the moral graph, the consistency of sparse PNL$\mathcal{M}$ follows from the analysis in \cite{loh2014high}, which studies the consistency of the following two-stage framework for estimating sparse DAGs: (1) infer the moral graph from the support of the inverse covariance matrix; and (2) choose the best-scoring induced DAG from the moral graph. The authors investigate conditions for identifiability of the underlying DAG from observational data and establish the consistency of the two-stage framework. %In particular, the moral graph is an important case because it can consistently be estimated using the algorithms for estimation of inverse covariance matrix \cite{loh2014high}. The consistency of sparse DAG estimation using PNL$\mathcal{M}$ follows from the consistency of PNL estimation and the consistency of the super-structure $\mathcal{M}$ (e.g., moral graph).

While PNL and PNL$\mathcal{M}$ enjoy desirable statistical properties for linear SEMs with Gaussian \cite{van2013ell} and non-Gaussian \cite{loh2014high} noise, the computational challenges associated with these problems have not been fully addressed. This paper aims to bridge this gap. 

%%%%%%%%%%%%%%%%%%%%%%%%%%%%%%%%%%
\section{Mathematical Formulations} \label{Sec: Math models}
%%%%%%%%%%%%%%%%%%%%%%%%%%%%%%%%%% 
Prior to presenting mathematical formulations for solving PNL$\mathcal{M}$, we discuss a property of DAG learning from continuous data that distinguishes it from the corresponding problem for discrete data. To present this property in Proposition~\ref{Prop1: Cycle}, we need a new definition. 

\begin{figure}
	\centering	
	\includegraphics*[scale=1]{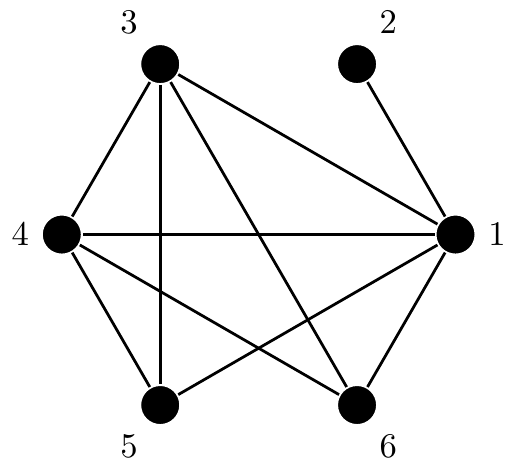}
	\includegraphics*[scale=1]{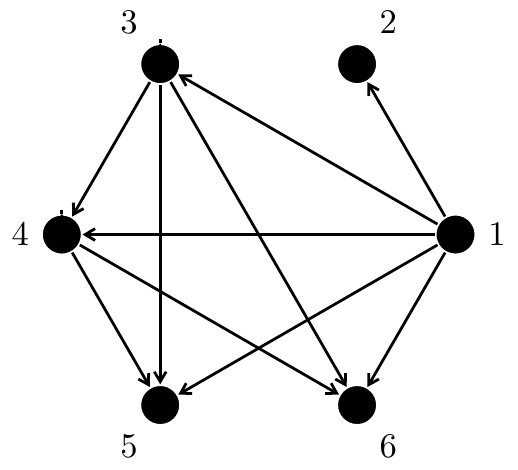}
	\caption{A super-structure graph $\mathcal{M}$ (left) and a tournament (right) with six nodes which does not contain any cycles.}
	\label{fig: tournument}
\end{figure}

\begin{definition}\label{def:1}
A  tournament is a directed graph obtained by specifying a direction for each edge in the super-structure graph $\mathcal{M}$ (see Figure~\ref{fig: tournument}).  
\end{definition}

\begin{prop}
	{\it There exists an optimal solution  $\mathcal{G}(B)$ to   PNL$\mathcal{M}$ \eqref{eq:PNLMform} with an $\ell_0$ (or  an $\ell_1$) regularization that is a  cycle-free  tournament.} \label{Prop1: Cycle} 
\end{prop} 

All proofs are given in Appendix I. Proposition~\ref{Prop1: Cycle} implies that for DAG learning from continuous variables, the search space reduces to acyclic  tournament structures. This is a far smaller search space when compared with the super-exponential $O\Big(m!2^{m \choose 2}\Big)$ search space of DAGs for discrete variables. However, one has to also identify the optimal $\beta$ parameters. This  search for optimal $\beta$ parameters is critical, as it further reduces the super-structure to the edges of the DAG by removing the edges with zero $\beta$ coefficients. 

A solution method based on brute force enumeration of all tournaments requires $m!\gamma$ computational time when $\mathcal{M}$ is complete, where $\gamma$ denotes the computational time associated with solving PNL$\mathcal{M}$ given a known tournament structure. This is because when $\mathcal{M}$ is complete, the total number of tournaments  {(equivalently the total number of permutations)} is $m!$. However, when $\mathcal{M}$ is incomplete, the number of DAGs is fewer than $m!$. The topological search space is $m!$ regardless of  the structure of $\mathcal{M}$ and several topological orderings can correspond to the same DAG. The TO formulation \cite{park2017bayesian} is based on this search space. In Section~\ref{sec:LN}, we discuss a search space based on the layering of a DAG, which uniquely identifies a DAG, and propose the corresponding Layered Network (LN) formulation, which effectively utilizes the structure of $\mathcal{M}$. We first discuss existing mathematical formulations for the PNL$\mathcal{M}$ optimization problem \eqref{eq:PNLMform} in the next section. Given the desirable statistical properties of  $\ell_0$ regularization \cite{van2013ell} and the fact that existing mathematical formulations  are given for $\ell_1$ regularization, we present the formulations for  $\ell_0$  regularization. %and briefly discuss the formulations for $\ell_1$ regularization.\ 
We outline the necessary notation below.

\noindent \textbf{Indices}\\
$V = \{1,2,\dots,m\}$:  index set of random variables\\
$\mathcal{D}= \{1,2,\dots,n\}$: index set of samples \vspace{0.1in}\\
\noindent \textbf{Input} \\
$\mathcal{M}=(V,E)$: an undirected super-structure graph (e.g., the moral graph)\\
$\overrightarrow{\mathcal{M}}=(V,\overrightarrow{E})$: the bi-directional graph corresponding to the undirected graph $\mathcal{M}$ \\
$\mathcal{X} = (\mathcal{X}_1, \dots, \mathcal{X}_m)$, where $\mathcal{X}_v = (x_{1v}, x_{2v}, \dots, x_{nv})^{\top}$ and $x_{dv}$ denotes $d$th sample ($d \in \mathcal{D}$) of random variable $X_v$  \\
$\lambda:$ tuning parameter (penalty coefficient) \vspace{0.1in}\\
\noindent \textbf{Continuous optimization variables} \\
$\beta_{jk}$: weight of arc $(j, k)$ representing the regression coefficients $\forall (j,k) \in \overrightarrow{E}$ \vspace{0.1in}\\
\noindent\textbf{Binary optimization variables} \\
$z_{jk}=1 \, \, \text{if arc} \, \,  (j, k) \, \text{exists in a DAG}; \text{otherwise} \, 0, \, \forall (j,k) \in \overrightarrow{E}$ \\
$g_{jk}=1 \, \, \text{if} \, \, \beta_{jk} \neq 0; \, \text{otherwise}  \, 0, \, \forall (j,k) \in \overrightarrow{E}$ \vspace{0.1in}

\subsection{Existing Mathematical Models}\label{sec:existing}
The main source of difficulty in solving PNL$\mathcal{M}$ is due to the acyclic nature of DAG imposed by the constraint in \eqref{Eq: DAG const}. %There are several ways to encode such a constraint, which we discuss in the next subsection.   
A popular technique for ensuring acyclicity is to use cycle elimination constraints, which were first introduced in the context of the Traveling Salesman Problem (TSP) {in \cite{dantzig1954solution}.} 
%Note that all existing mathematical formulations are given for $\ell_1$-regularization. Here, we extend these formulations to $\ell_0$-regularization.  

Let $\mathcal{C}$ be the set of all possible cycles and $\mathcal{C}_A \in \mathcal{C}$ be the set of arcs defining a cycle and define $F(\beta, g):= n^{-1} \sum_{k\in V}\sum_{d\in \mathcal{D}} \Big(x_{dk}-\sum_{(j,k) \in \overrightarrow{E}} \beta_{jk}x_{dj}\Big)^2 + \lambda \sum_{(j,k) \in \overrightarrow{E}}  g_{jk}$. Then, the $\ell_0$-PNL$\mathcal{M}$ model can be formulated as
\begin{subequations}
	\begin{alignat}{3}
	\label{CP-obj} \quad  \textbf{$\ell_0$-CP} \quad   \underset{}{\min}\quad & \, F(\beta, g) \\
	& \label{CP-con1} -M g_{jk} \leq \beta_{jk} \leq M g_{jk},  \quad && \forall (j,k) \in \overrightarrow{E},\\
	 &\label{CP-con2} \sum_{(j,k ) \in \, \mathcal{C}_A} g_{jk} \leq |\mathcal{C}_A|-1, \quad &&  \forall \mathcal{C}_A \in \mathcal{C}, \\
	&\label{CP-con3}  g_{jk} \in \{0,1\},\quad && \forall (j,k) \in \overrightarrow{E}.
	\end{alignat}
\end{subequations}

Following \cite{park2017bayesian}, the objective function \eqref{CP-obj} is an expanded version of $l_n(\beta)$ in PNL$\mathcal{M}$ (multiplied by $2n^{-1}$) with an $\ell_0$ regularization. The constraints in \eqref{CP-con1} stipulate that $\beta_{jk} \neq 0$ only if $z_{jk}=1$, where $M$ is a sufficiently large constant. The constraints in \eqref{CP-con2} rule out all cycles. Note that for $|\mathcal{C}_A|=2$, constraints in \eqref{CP-con2} ensure that at most one arc exists among two nodes. The last set of constraints specifies the binary nature of the decision vector $g$. Note that $\beta$ variables are continuous and unrestricted; however, in typical applications, they can be bounded by a finite number $M$. This formulation requires $|\overrightarrow{E}|$ binary variables and an exponential number of constraints. A cutting plane method \cite{nemhauser1988integer} that adds the cycle elimination inequalities as needed is often used to solve this problem. We refer to this formulation as the {\it cutting plane} (CP) formulation. 

\begin{remark}\label{remark:1}
For a complete {super-structure} $\mathcal{M}$, it suffices to impose the set of constraints in \eqref{CP-con2} only for cycles of size 2 and 3 given by
\begin{subequations}
	\begin{alignat*}{2}
	g_{ij} +g_{jk} +g_{ki} \leq 2, \quad & \forall i,j,k \in V, \, i \neq j \neq k, \\
	g_{jk} +g_{kj} \leq 1,\quad & \forall  j,k \in V, \, j \neq k. 
	\end{alignat*}
\end{subequations}
In other words, the CP formulation (both with $\ell_0$ and $\ell_1$ regularizations) needs a polynomial number of constraints for complete {super-structure} $\mathcal{M}$. 
\end{remark}

The second formulation is based on a well-known combinatorial optimization problem, known as \textit{linear ordering} (LO) \cite{grotschel1985acyclic}. Given a finite set $S$ with $q$ elements, a linear ordering of $S$ is a permutation $\mathcal{P} \in S_q$ where $S_q$ denotes the set of all permutations with $q$ elements. In the LO problem, the goal is to identify the best permutation among $m$ nodes. The ``cost" for a permutation $\mathcal{P}$ depends on the order of the elements in a pairwise fashion. Let $p_j$ denote the order of node $j\in V$ in permutation $\mathcal{P}$. Then, for two nodes $j, k \in \{1, \dots , m\}$, the cost is $c_{jk}$ if the order of node $j$ precedes the order of node $k$ ($p_j \prec  p_k$) and is $c_{kj}$ otherwise ($p_j \succ p_k$). A binary variable $w_{jk}$ indicates whether $p_j \prec p_k$. Because $w_{jk} + w_{kj} = 1$ and $w_{jj} = 0$, one only needs ${m\choose 2}$ variables to cast the LO problem as an IP formulation \cite{ grotschel1985acyclic}. 

The LO formulation for DAG learning from continuous data has two noticeable differences compared with the classical LO problem: (i) the objective function is quadratic, and (ii) an additional set of continuous variables, i.e., $\beta$s, is added. Cycles are ruled out by directly imposing the linear ordering constraints. The PNL$\mathcal{M}$ can be formulated as \eqref{eq:LOform}. 
\begin{subequations}\label{eq:LOform}
	\begin{alignat}{3}
	\label{LO-con0} \textbf{$\ell_0$-LO} \quad \min & \quad F(\beta,g), \\
	%& \eqref{CP-con1}\\
	& \label{LO-con1} -M g_{jk} \leq \beta_{jk} \leq M g_{jk},  \quad && \forall(j,k) \in \overrightarrow{E},\\
			& 	\label{LO-con3}  g_{jk}  \leq w_{jk}, \quad && \forall (j,k) \in \overrightarrow{E}, \\
	& 	\label{LO-con2} w_{jk} + w_{kj} = 1, \quad && \forall j, k \in V, j \neq k, \\
	& \label{LO-con4} w_{ij} +w_{jk} + w_{ki} \leq 2, \quad  &&\forall i, j, k \in V, i \neq j \neq k, \\
	& \label{LO-con5} w_{jk} \in \{0,1\}, \quad&& \forall j, k \in V, j \neq k, \\
	& \label{LO-con6} g_{jk} \in  \{0,1\}, \quad &&\forall (j,k) \in \overrightarrow{E}.
	\end{alignat}
\end{subequations}
The interpretation of constraints \eqref{LO-con1}-\eqref{LO-con2} is straightforward. The constraints in \eqref{LO-con3} imply that if node $j$ appears after node $k$ in a linear ordering ($w_{jk}=0$), then there should not exist an arc from $j$ to $k$ ($g_{jk}=0$). The set of inequalities \eqref{LO-con4} implies that if $p_i \prec p_j$ and $p_j \prec p_k$, then $p_i \prec p_k$. This ensures the linear ordering of nodes and removes cycles. 

The third approach for ruling out cycles is to impose a set of constraints such that the nodes follow a topological ordering. A topological ordering is a linear ordering of the nodes of a graph such that the graph contains an arc $(j,k)$ if node $j$ appears before node $k$ in the linear order.  
Define decision variables $o_{rs} \in \{0,1\}$ for all $r, s \in \{1, \dots, m\}$. This variable takes value 1 if topological order of node $r$ (i.e., $p_r$) equals $s$, and 0, otherwise.  If a topological ordering is known, the DAG structure can be efficiently learned in polynomial time \cite{shojaie2010penalized}, but the problem remains challenging when the ordering is not known. The topological ordering prevents cycles in the graph. {This property is used in \cite{park2017bayesian} to model the problem of learning a DAG with $\ell_1$ regularization}. We extend their formulation to $\ell_0$ regularization. The {\it topological ordering} (TO) formulation is given by 
\begin{subequations}
	\begin{alignat}{3}
	\label{TO-obj}  \textbf{$\ell_0$-TO} \quad  \min & \quad  F(\beta, g),\\
	& \label{TO-con1} -M g_{jk} \leq \beta_{jk} \leq M g_{jk},  \quad && \forall(j,k) \in \overrightarrow{E},\\
			& 	\label{TO-con3}  g_{jk}  \leq z_{jk}, \quad && \forall (j,k) \in \overrightarrow{E}, \\
	& 	\label{TO-con2} z_{jk} + z_{kj} \leq 1, \quad && \forall (j,k) \in \overrightarrow{E}, \\
	\label{TO-con4} & z_{jk} - m z_{kj} \leq \sum_{s \in V} s \, (o_{ks} - o_{js}), \quad&& \forall (j,k) \in \overrightarrow{E},\\
	\label{TO-con5} & \sum_{s \in V} o_{rs} =1, \quad && \forall r \in V, \\ 
	\label{TO-con6} & \sum_{r \in V} o_{rs} =1, \quad  &&\forall s \in V,\\
	\label{TO-con7} & z_{jk} \in \{0,1\},\quad&& \forall (j,k) \in \overrightarrow{E},\\
	& \label{TO-con8} o_{rs} \in \{0,1\},  \quad &&\forall \, r, s \in \{1,2, \dots, m\},\\
	&  \label{TO-con9} g_{jk} \in \{0,1\},  \quad&& \forall   (j,k) \in \overrightarrow{E}. 
	\end{alignat}
\end{subequations}

\begin{figure}[]
	\begin{subfigure}[b]{0.45\textwidth}
		\centering{
		\begin{tikzpicture}[scale=1] 
		\SetGraphUnit{2}
		\renewcommand*{\VertexLineColor}{white}
		\renewcommand*{\VertexLightFillColor}{black}
		\renewcommand*{\VertexLineWidth}{1pt}
		\GraphInit[vstyle=Welsh]
		\Vertices{circle}{1,2,3,4}
		\SetUpEdge[style={->,very thick}]
		\Edges(1,2)
		\Edges(2,3)
		\Edges(4,3)
		\end{tikzpicture}}
		\caption{} 	 \label{Fig: L0-Fig1}
	\end{subfigure}
	\hfill
	\begin{subfigure}[b]{0.45\textwidth}
		\centering{
		\begin{tikzpicture}[scale=1] 
		\SetGraphUnit{2}
		\renewcommand*{\VertexLineColor}{white}
		\renewcommand*{\VertexLightFillColor}{black}
		\renewcommand*{\VertexLineWidth}{1pt}
		\GraphInit[vstyle=Welsh]
		\Vertices{circle}{1,2,3,4}
		\SetUpEdge[style={->,very thick}]
		\Edges(1,2)
		\Edges(2,3)
		\Edges(4,3)
		\Edges(2,4)
		\Edges(1,3)
		\Edges(1,4)
		\end{tikzpicture}}
		\caption{}  \label{Fig: L0-Fig2}
	\end{subfigure}
	\caption{The role of the binary decision variables in $\ell_0$ regularization: %Without the {$g$ variables} in $\ell_0$ regularization, TO cannot correctly quantify the objective value associated with the optimal graph (a). 
	Using $z$ (instead of $g$) in the objective function creates a graph similar to (b) and counts the number of arcs instead of the number of non-zero $\beta$s in (b).} \label{Fig: L0-Fig}
\end{figure}
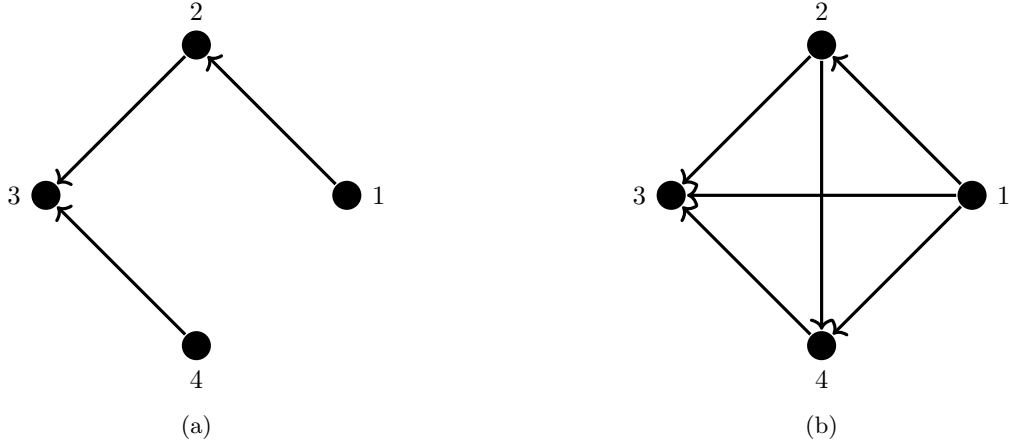

In this formulation, $z_{jk}$ is an auxiliary binary variable which takes value 1 if an arc exists from node $j$ to node $k$. Recall that, $g_{jk}=1$ if $|\beta_{jk}|>0$. The constraints in \eqref{TO-con3} enforce the correct link between $g_{jk}$ and $z_{jk}$, i.e., $g_{jk}$ has to take value zero if $z_{jk}=0$. The constraints in \eqref{TO-con2} imply that there should not exist a bi-directional arc among two nodes. This inequality can be replaced with equality (Corollary~\ref{cor:1}).   The constraints in \eqref{TO-con4} remove cycles by imposing an ordering among nodes. The set of constraints in \eqref{TO-con5}-\eqref{TO-con6} assigns a unique topological order to each node. The last two sets of constraints indicate the binary nature of decision variables $o$ and $z$.

{Corollary~\ref{cor:1}, which is a direct consequence of Proposition~\ref{Prop1: Cycle}, implies that we can use $z_{jk} = 1 -z_{kj} \, \,$ for all $j < k$ and reduce the number of binary variables.}

\begin{cor}\label{cor:1}
	The constraints in \eqref{TO-con2} can be replaced by $z_{jk}+z_{kj}=1, \, \, \forall (j,k) \in \overrightarrow{E}$.
\end{cor}

%Given a constructed DAG for $|\beta_{jk}|\neq 0$, 

In constraints \eqref{TO-con1}-\eqref{TO-con8}, both variables $z$ and $g$ are needed to correctly model the $\ell_0$ regularization term in the objective function (see Figure \ref{Fig: L0-Fig1}). This is because the constraints \eqref{TO-con4} satisfy the transitivity property: if $z_{ij}=1$ for $(i,j) \in \overrightarrow{E}$ and $z_{jk}=1$ for $(j,k) \in \overrightarrow{E}$, then $z_{ik}=1$ for $(i,k) \in \overrightarrow{E}$, since $z_{ij}=1$ implies that $\sum_{s \in V} s (o_{js} - o_{is}) \geq 1$; similarly, $z_{jk}=1$ implies $\sum_{s \in V} s \, (o_{ks} - o_{js})\ge 1$. If we sum both inequalities, we have $\sum_{s \in V} s \, (o_{ks} - o_{is}) \geq 2$, which enforces $z_{ik}=1$. Such a transitivity relation, however, need not hold for the decision vector $g$. In other words, the decision variable $g_{jk}$ is used to keep track of the number of non-zero weights $\beta_{jk}$ associated with the arc $(j,k)$, and the decision vector $z$ is used to remove cycles via the set of constraints in \eqref{TO-con4} by creating an acyclic tournament on the super-structure $\mathcal{M}$. A tournament on super-structure $\mathcal{M}$ assigns a direction for each edge in an undirected super-structure $\mathcal{M}$. %Recall the decision variable $g_{jk}=1$ if $\beta_{jk} \neq 0$. 
In other words, if we were to let $g_{ij}=z_{ij}$ for $(i,j)\in \overrightarrow{E}$ and hence use the decision variable $z$ in the objective, then we would be counting the number of edges, equal to $|E|$, instead of number of non-zero $\beta$ values. %\hmm{That is always equal to $|E|$}. %\as{We next give a definition followed by a proposition.}

\subsection{A New Mathematical Model: The Layered Network (LN) Formulation}\label{sec:LN}
\begin{figure}[]
	%\hspace*{-1in}
		\centering
		\includegraphics[scale=1]{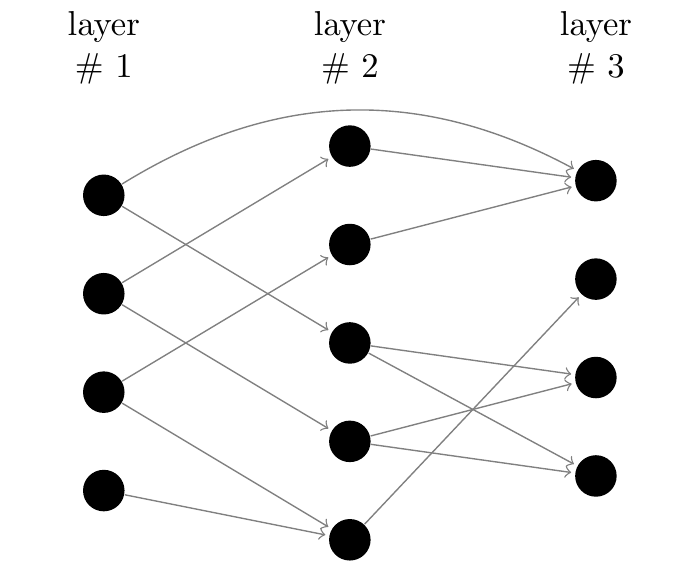}
		\caption{Layered Network encoding of a DAG.}  \label{LNencoding}
	\end{figure}

As an alternative to the existing mathematical formulations, we propose a new formulation for imposing acyclicity constraints that is motivated by the layering of nodes in  DAGs \cite{healy2002branch}. More specifically, our formulation ensures that the resulting graph is a \emph{layered network}, in the sense that there exists no arc from a layer $v$ to layer $u$, where $u<v$. Let $\psi_k$ be the \textit{layer value} for node $k$. One may interpret $\psi_k$ as $\sum_{s=1}^{m} {s \, o_{ks}}$ for all $k \in V$, where the variables $o_{ks}$ are as  defined in the TO formulation. However, note that the notion of $\psi_k$ is more general because $\psi_k$ need not be integer. Figure \ref{LNencoding} depicts the layered network encoding of a DAG. 
 With this notation, our \textit{layered network} (LN) formulation can be written as  
 
\begin{subequations}\label{eq:LNform}
	\begin{alignat}{3}
	 \quad  \label{L-obj} \quad \min & \quad F(\beta,g),\\
	& \label{LN-con1} -M g_{jk} \leq \beta_{jk} \leq M g_{jk},  \quad && \forall (j,k) \in \overrightarrow{E},\\
				& 	\label{LN-con3}  g_{jk}  \leq z_{jk}, \quad&& \forall (j,k) \in \overrightarrow{E}, \\
	\label{LN-con2} & z_{jk} + z_{kj} = 1, \quad \quad &&\forall (j,k) \in \overrightarrow{E}, \\
 &	\label{LN-con4} z_{jk} - (m-1) z_{kj} \leq  \psi_k - \psi_j, \quad && \forall (j,k) \in \overrightarrow{E},\\
	& \label{LN-con5} z_{jk} \in \{0,1\}, \quad&& \forall (j,k) \in \overrightarrow{E},\\
	 &  \label{LN-con6} 1 \leq  \psi_{k} \leq m, \quad &&\forall k \in V, \\
	 	&  \label{LN-con7} g_{jk} \in \{0,1\},  \quad&&   \forall (j,k) \in \overrightarrow{E}. 
	\end{alignat}
\end{subequations}

The interpretation of the constraints \eqref{LN-con1}-\eqref{LN-con3} is straightforward. The constraints in \eqref{LN-con4} ensure that the graph is a layered network. The last set of constraints indicates the continuous nature of the decision variable $\psi$ and gives the tightest valid bound for $\psi$. It suffices to consider any real number for  layer values $\psi$  as long as layer values of any two nodes differ by at least one if there exists an arc between them. Additionally, LN uses a tighter inequality compared to TO, by replacing $m$ with parameter $m-1$ in \eqref{LN-con4}. This is because the difference between the layer values of two nodes can be at most $m-1$ for a DAG with $m$ nodes.\ The next proposition establishes the validity of the LN formulation.

\begin{prop}
\textit{An optimal solution to \eqref{eq:LNform} is an optimal solution to \eqref{eq:PNLMform}.} \label{Prop2: LNProof}
\end{prop}

The LN formulation highlights a desirable property of the layered network representation of a DAG in comparison to the topological ordering representation. Let us define nodes in layer 1 as the set of nodes that have no incoming arcs in the DAG, nodes in layer 2 as the set of nodes that have incoming arcs only from nodes in the layer 1, layer 3 as the set of nodes that have incoming arcs from layer 2 (and possibly layer 1), etc.\ (see Figure \ref{LNencoding}). The \textit{minimal layer number} of a node is the length of the \emph{longest} directed path from any node in layer 1 to that node. For a given DAG, there is a unique {minimal layer number}, but not a unique topological order. As an example, Figure \ref{Fig: L0-Fig1} has three valid topological orders: (i) 1,2,4,3, (ii) 1,4,2,3, and (iii) 4,1,2,3. In contrast, it has a unique layer representation, 1,2,3,1.

There is a one-to-one correspondence between minimal layer numbering and a DAG. However, the solutions of the LN formulation, i.e., $\psi$ variables (layer values), do not necessarily correspond to the minimal layer numbering. This is because the LN formulation does not impose additional constraints to enforce a minimal numbering and can output solutions that are not minimally numbered. %Such constraints can be avoided in the LN model because we do not branch on those continuous $\psi$ variables. 
However, because branch-and-bound does not branch on continuous variables, alternative (non-minimal) feasible solutions for the $\psi$ variables do not impact the branch-and-bound process. On the contrary, we have multiple possible representations of the same DAG with topological ordering. Because topological ordering variables are binary, the branch-and-bound method applied to the TO formulation explores multiple identical DAGs as it branches on the topological ordering variables. This enlarges the size of the branch-and-bound tree and increases the computational burden.       

Layered network representation also has an important practical implication: Using this representation, the search space can be reduced to the total number of ways we can layer a network (or equivalently the total number of possible minimal layer numberings) instead of the total number of topological orderings. When the super-structure $\mathcal{M}$ is complete, both quantities are the same, and equal to $m!$. Otherwise, a brute-force search for finding the optimal DAG  has computational time $\mathcal{L}{\gamma}$, where $\mathcal{L}$ denotes the total number of minimal layered numberings, and $\gamma$ is the computational complexity of solving PNL$\mathcal{M}$ given a known tournament structure.

\ignore{
\begin{figure}[]
	\begin{subfigure}[b]{0.45\textwidth}
		\begin{tikzpicture}[scale=1]
		\SetGraphUnit{2}
		\renewcommand*{\VertexLineColor}{white}
		\renewcommand*{\VertexLightFillColor}{black}
		\renewcommand*{\VertexLineWidth}{1pt}
		\GraphInit[vstyle=Welsh]
		\Vertices{circle}{1,2,3,4}
		\SetUpEdge[style={->,very thick}]
		\Edges(1,2)
		\Edges(2,3)
		\Edges(4,3)
		\end{tikzpicture}
		\caption{}
	\end{subfigure}
	\hspace{0.1in}
	\begin{subfigure}[b]{0.45\textwidth}
		\begin{tikzpicture}[scale=1]
		\SetGraphUnit{2}
		\renewcommand*{\VertexLineColor}{white}
		\renewcommand*{\VertexLightFillColor}{black}
		\renewcommand*{\VertexLineWidth}{1pt}
		\GraphInit[vstyle=Welsh]
		\Vertices{circle}{1,2,3,4}
		\SetUpEdge[style={->,very thick}]
		\Edges(1,2)
		\Edges(2,3)
		\Edges(4,3)
		\Edges(1,3)
		\Edges(1,4)
		\end{tikzpicture}
		\caption{}
	\end{subfigure}
	\caption{Illustration of the Layered Network formulation: (a) nodes 1, 2, 3, 4 are placed in layers 1, 2, 3, 1, respectively; (b) nodes 1, 2, 3, 4 are placed in layers 1, 2, 3, 2, respectively.} \label{Fig. layer}
\end{figure}
}

We close this section by noting that a set of constraints similar to \eqref{LN-con4} was introduced in \cite{cussens2010maximum} for learning pedigree. However, {the formulation in \cite{cussens2010maximum}} requires an exponential number of variables. In more recent work on DAGs for discrete data, Cussens and colleagues have focused on a tighter formulation for removing cycles, known as cluster constraints \cite{cussens2012bayesian, cussens2017polyhedral, cussens2017bayesian}. To represent the set of cluster constraints, variables have to be defined according to the parent set choice leading to an exponential number of variables. Thus, such a representation is not available in the space of arcs. %Later in Proposition~\ref{Prop5: BB} we shed light on why LN formulation performs well for learning DAG from continuous data. %\as{the manual refs to props etc can become problematics}

%%%%%%%%%%%%%%%%%%%
\subsubsection{Layered Network with $\ell_1$ regularization}\label{sec:LNL1}
%%%%%%%%%%%%%%%%%%%
Because of its convexity, the structure learning literature has utilized the $\ell_1$-regularization for learning DAGs from continuous variables \cite{han2016estimation, park2017bayesian, shojaie2010penalized, xiang2013lasso, zheng2018dags}. The LN formulation with $\ell_1$-regularization can be written as   
\begin{subequations}
	\begin{align}
	   \label{LN-L1-Obj}  \underset{}{\min} & \, \, \frac{1}{n} \sum_{i\in I}\sum_{k\in V} (x_{ik}- \sum_{(j,k) \in E ^{\rightarrow}} \beta_{jk}x_{ij})^2 +\lambda \sum_{(j,k) \in \overrightarrow{E}}  |\beta_{jk}|, \\
	 	& \label{LN-con0} -M z_{jk} \leq \beta_{jk} \leq M z_{jk}  \quad& \forall (j,k) \in \overrightarrow{E},\\
	 		 & \nonumber \eqref{LN-con4}-\eqref{LN-con6}. &
	\end{align}
\end{subequations} 

\begin{remark}\label{remark:3}
For a complete super-structure $\mathcal{M}$, $\psi_k = \sum_{j \in V \setminus k} z_{jk} \, \forall k \in V$. Thus, the LN formulations (both $\ell_0$ and $\ell_1$) can be encoded without $\psi$ variables by writing \eqref{LN-con4} as 
\begin{equation*}
z_{jk} - (m-1) z_{kj} \leq   \sum_{j \in V \setminus k } z_{jk}  -  \sum_{k \in V \setminus j} z_{kj}  \quad  \forall \, j,k \in V \quad j \neq k.
\end{equation*}    
\end{remark}

\begin{remark}\label{remark:2}
	CP and LO formulations reduce to the same formulation for $\ell_1$ regularization when the super-structure $\mathcal{M}$ is complete by letting $w_{ij}=g_{ij}$ in formulation \eqref{eq:LOform} for all $(j,k)\in \overrightarrow{E}$.
\end{remark}

\begin{table}[t]
	\fontsize{7}{12}\selectfont
	\caption{The number of binary variables and the number of constraints}
	\begin{tabular}{llllllllll}		
		\hline
		& \multicolumn{4}{c}{Incomplete (moral) $\mathcal{M}$} &  & \multicolumn{4}{c}{Complete (moral) $\mathcal{M}$}   \\  \cline{2-5} \cline{7-10}
		& CP & LO & TO & LN & & CP & LO & TO & LN  \\  \hline 
		$\#$ Binary Vars ($\ell_0$) & $|\overrightarrow{E}|$ & $|\overrightarrow{E}|+ {m\choose 2}$ & $m^2 +|E|+|\overrightarrow{E}|$ &  $|E|+|\overrightarrow{E}|$ &  &  $2{m\choose 2}$ &  ${m\choose 2}$ & $m^2+ 3{m\choose 2}$ & $3{m\choose 2}$  \vspace{0.1in} \\ \hline
		$\#$ Binary Vars ($\ell_1$) & $|E|$ & ${m\choose 2}$ & $m^2 +|E|$ &  $|E|$ &  &  ${m\choose 2}$ &  ${m\choose 2}$ & $m^2+ {m\choose 2}$ & ${m\choose 2}$  \vspace{0.1in} \\  \hline 
		$\#$ Constraints \\ (both $\ell_0$ and $\ell_1$) & \textit{Exp} &   2${m\choose 3}$ & $|\overrightarrow{E}|+2m$ & $|\overrightarrow{E}|$ & &    2${m\choose 3}$ & $2{m\choose 3}$ & ${m\choose 2}+ 2m$ &${m\choose 2}$ \vspace{0.1in} \\ \hline 
	\end{tabular}
	\label{tbl:nvars}
\end{table}

An advantage of the $\ell_1$-regularization for DAG learning is that all models (CP, LO, TO and LN) can be formulated without decision variables $g_{jk}$, since counting the number of non-zero $\beta_{jk}$ is no longer necessary.

Table~\ref{tbl:nvars} shows the number of binary variables and the number of constraints associated with cycle prevention constraints in each model. Evidently, $\ell_0$ models require additional binary variables compared to the corresponding $\ell_1$ models. Note that the number of binary variables and constraints for the LN formulation solely depend on the number of edges in the super-structure $\mathcal{M}$. This property is particularly desirable when the {super-structure} $\mathcal{M}$ is {sparse}.\ The LN formulation requires the fewest number of constraints among all models. 
The LN formulation also requires fewer binary variables than the TO formulation. More importantly, different topological orders for the same DAG are symmetric solutions to the associated TO formulation. Consequently, branch-and-bound requires exploring multiple symmetric formulations as it branches on fractional TO variables. As for the LO formulation, the number of constraints is $O(m^3)$ which makes its continuous relaxation cumbersome to solve in the branch-and-bound process. The LN formulation is compact, whereas the CP formulation requires an exponential number of constraints for incomplete {super-structure} $\mathcal{M}$. The CP formulation requires fewer binary variables for $\ell_0$ formulation than  LN; both formulations need the least number of binary variables for $\ell_1$ regularization. 

In the next section, we discuss the theoretical strength of these mathematical formulations and provide a key insight on why the LN formulation performs well for learning DAGs from continuous data.  

\section{Continuous Relaxation} \label{Sec: LP}
One of the fundamental concepts in IP is relaxations, wherein some or all constraints of a problem are loosened. Relaxations are often used to obtain a sequence of easier problems which can be solved efficiently yielding bounds and approximate, not necessarily feasible, solutions for the original problem. Continuous relaxation is a common relaxation obtained by relaxing the binary variables of the original mixed-integer quadratic program (MIQP) and allowing them to take real values. Continuous relaxation is at the heart of branch-and-bound methods for solving MIQPs.  
An important concept when comparing different MIQP formulations is the strength of their continuous relaxations. 

\begin{definition}\label{def:2}
A formulation $A$ is said to be \emph{stronger} than formulation $B$ if $\mathcal{R}(A) \subset \mathcal{R} (B)$ where $\mathcal{R}(A)$ and $\mathcal{R}(B)$ correspond to the feasible regions of continuous relaxations of $A$ and $B$, respectively. 
\end{definition}

\begin{figure}[]
\begin{center}
	\begin{tikzpicture}[scale=0.4]
	
	\draw[->] (-4,-4)->(6,-4); 
	\draw[->] (-4,-4)->(-4,4); 
	
	\foreach \Point in {(0,2), (0,4), (2,0), (2,4), (0,-2), (0,-4), (2,0), (2,-4), (0,0), (-2,0), (-2,4), (-2,-2), (-2,2), (2,2), (-2, -4), (4,2), (4,0), (4,-2), (4, -4), (4,4), (2,-2)}{
		\node [gray] at \Point {\textbullet};}
	
	\foreach \Point in {(0,0), (0,2), (2,2), (-2,0), (0, -2), (2,0), (2, -2)}{
		\node at \Point {\textbullet};}
	
	\node (pol) [draw, thick, blue!40!blue,rotate=60,minimum size=2.4cm,regular polygon, regular polygon sides=7] at (0.5,0) {}; 	
	 
	\draw[-] (0,-2)--(-2,0); 
	\draw[-] (0,-2)--(2,-2); 
	\draw[-] (2,-2)--(2,0); 
	\draw[-] (2,0)--(2,2); 
	\draw[-] (2,2)--(0,2); 
	\draw[-] (0,2)--(-2,0); 
		\draw[red] (-2,0)--(0,3); 
			\draw[red] (0,3)--(2,2); 
				\draw[red] (2,2)--(3,0); 
					\draw[red] (3,0)--(2,-2); 
						\draw[red] (2,-2)--(0,-2.5); 
												\draw[red] (0,-2.5)--(-2,0); 
	%\draw[-] (0,-2)--(-1.5,-1); 
	%\draw[-] (-1.5,-1)--(-1.5,1); 
	
	\end{tikzpicture}
\end{center}
\caption{Continuous relaxation regions of three IP models. The tightest model (convex hull) is represented by the black polygon strictly inside other polygons. The blue and red polygons represent valid yet weaker formulations. The black points show the feasible integer points for all three formulations.}
\end{figure}
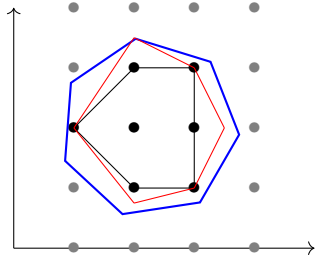

\begin{prop}
{\it The LO formulation is stronger than the LN formulation, that is, \\ $\mathcal{R}(LO) \subset \mathcal{R}(LN)$.} \label{Prop3: LNLO}
\end{prop}

\begin{prop}
{\it When the parameter $m$ in \eqref{TO-con4} is replaced with $m-1$, the TO formulation is stronger than the LN formulation, that is,  $\mathcal{R}(TO) \subset \mathcal{R}(LN)$.} \label{Prop3: LNTO}
\end{prop}

These propositions are somewhat expected because the LN formulation uses the fewest number of constraints. Hence, the continuous relaxation feasible region of the LN formulation is loosened compared to the other formulations. The next two results justify the advantages of the LN formulation. 

\begin{prop}
\it{Let $\beta^{\star}_{jk}$ denote the optimal coefficient associated with an arc $(j,k) \in \overrightarrow{E}$ from \eqref{eq:PNLMform}. For both $\ell_0$ and $\ell_1$ regularizations, the initial continuous relaxations of the LN formulation attain as tight an optimal objective function value as the LO, CP, TO formulations if $M \geq 2  \underset{(j,k) \in \overrightarrow{E}}{\max} \,  |\beta^{\star}_{jk}|$.} \label{Prop4: Root}
\end{prop}

Proposition \ref{Prop4: Root} states that although the LO and TO formulations are tighter than the LN formulation with respect to the feasible region of their continuous relaxations, the continuous relaxation of all models attain the same objective function value (root relaxation). 

\begin{prop}
\it{For the same variable branching in the branch-and-bound process, the continuous relaxations of the LN formulation for both $\ell_0$ and $\ell_1$ regularizations attain as tight an optimal objective function value as LO, CP and TO, if $M \geq 2  \underset{(j,k) \in \overrightarrow{E}}{\max} \,  |\beta^{\star}_{jk}|$.} \label{Prop5: BB}
\end{prop}

Proposition~\ref{Prop5: BB} is at the crux of this section. It shows that not only does the tightness of the optimal objective function value of the continuous relaxation   hold for the root relaxation, but it also holds throughout the branch-and-bound process under the specified condition on $M$, if the same branching choices are made. Thus, the advantages of {the LN formulation} are due to the fact that it is a compact formulation that entails the fewest number of constraints, while attaining the same optimal objective value of continuous relaxation as tighter models.    

In practice, finding a tight value for $M$ is difficult. {Our computational results show that the approach suggested in \cite{park2017bayesian} to obtain a value of $M$, which is explained in Section~\ref{Sec: Computational} and used in our computational experiments, always satisfies the condition in Proposition~\ref{Prop5: BB} across all  generated instances. }%A more conservative choice of $\lambda$ practically always satisfies the condition in Proposition ~\ref{Prop5: BB}}.

\section{Comparison of MIQP Formulations} \label{Sec: Computational}
We present numerical results comparing the proposed LN formulation with existing approaches. Experiments are performed on a cluster operating on UNIX  with Intel Xeon E5-2640v4 2.4GHz. All MIQP formulations are implemented in the Python programming language. Gurobi 8.0 is used as the MIQP solver. A time limit of $50m$ (in seconds), where $m$ denotes the number of nodes, is imposed across all experiments after which runs are aborted. Unless otherwise stated, an MIQP optimality gap of $0.001$ is imposed across all experiments; the gap is calculated by $\frac{UB-LB}{UB}$ where \textit{UB} denotes the objective value associated with the best feasible integer solution (incumbent) and \textit{LB} represents the best obtained lower bound during the branch-and-bound process. 

For CP, instead of incorporating all constraints given by \eqref{CP-con2}, we begin with no constraint of type \eqref{CP-con2}. Given an integer solution with cycles, we detect a cycle and impose a new cycle prevention constraint to remove the detected cycle. Depth First Search (DFS) can detect a cycle in a directed graph with complexity $O(|V|+|E|)$. Gurobi Lazy Callback is used, which allows adding cycle prevention constraints in the branch-and-bound algorithm, whenever an integer solution with cycles is found. The same approach is used by \cite{park2017bayesian}. Note that Gurobi solver follows a branch-and-cut implementation and adds many general-purpose and special-purpose cutting planes.

To select the $M$ parameter in all formulations we use the proposal of \cite{park2017bayesian}. Specifically, given $\lambda$, we solve each problem without cycle prevention constraints. We then use the upper bound $M = 2 \underset{(j,k) \in \overrightarrow{E}}{\max} \, |\beta_{jk}|$. %This gives the same $M$ for the mathematical models for a fixed value of $\lambda$. 
The results provided in \cite{park2017bayesian} computationally confirm that this approach gives a large enough value of $M$. We also confirmed the validity of this choice across all our test instances. 

\subsection{Synthetic datasets} \label{sec:synth-data}
%We first conduct our experiments with synthetic datasets. 
We use the \texttt{R} package \texttt{pcalg} to generate random Erd\H{o}s-R\'enyi graphs. Firstly, we create a DAG using \texttt{randomDAG} function and assign random arc weights (i.e., $\beta$) from a uniform distribution, $\mathcal{U}[0.1, 1]$. This {\it ground truth} DAG is used to assess the quality of estimates. Next, the resulting DAG and random  coefficients  are input to the \texttt{rmvDAG} function, which uses linear regression as the underlying model, to generate multivariate data (columns of matrix $\mathcal X$) with the standard normal error distribution. %The DAG used to generate the multivariate data is 
%assumed to be the true structure while it may not be an optimal solution for the objective function. This is particularly important for $\ell_1$ regularization because of the lack of consistency proof. We take this DAG 
%considered as the \textit{ground truth}, and is used to assess the quality of estimates from optimization models. 

We consider $m\in\{10,20,30,40\}$ nodes and $n \in \{100, 1000\}$ samples. The average outgoing degree of each node,  denoted by $d$, is set to 2. We generate 10 random graphs for each setting ($m$, $n$, $d$). The raw observational data, $\mathcal{X}$, for the datasets with $n=100$ is the same as first 100 rows of the datasets with $n=1000$. 

%We consider penalty coefficients $\lambda \in \{0.1, 1\}$. 

We consider two types of problem instances: (i) a set of instances for which the moral graph corresponding to the true DAG is available; (ii) a set of instances with a complete undirected graph, i.e., assuming no prior knowledge. The first class of problems is referred to as \textit{moral} instances, whereas the second class is called \textit{complete} instances. The raw observational data, $\mathcal{X}$, for moral and complete instances are the same. The function \texttt{moralize(graph)} in the \texttt{pcalg} R-package is used to generated the moral graph from the true DAG. The moral graph can also be (consistently) estimated from data using penalized estimation procedures with polynomial complexity %\as{give refs}
\cite{kalisch2007estimating, loh2014high}. However, since the quality of the moral graph equally affects all optimization models, the true moral graph is used in our experiments. 

We use the following IP-based metrics to measure the quality of a solution: Optimality gap (MIQP GAP), computation time in seconds (Time), Upper Bound (UB), Lower Bound (LB), computational time of root continuous relaxation (Time LP), and the number of explored nodes in the branch-and-bound tree. 

We also evaluate the quality of the estimated DAGs by comparing them with the ground truth. To this end, we use the average structural Hamming distance $(\mathrm{SHD})$, as well as true positive (TPR) and false positive rates (FPR). These criteria evaluate different aspects of the quality of the estimated DAGs: $\mathrm{SHD}$ counts the number of differences (addition, deletion, or arc reversal) required to transform predicted DAG to the true DAG; TPR is the number of correctly identified arcs divided by the total number of true arcs, $P$; FPR is the number of incorrectly identified arcs divided by the total number of negatives (non-existing arcs), $N$. For brevity, TPR and FPR plots are presented in Appendix II.

\subsection{Comparison of  $\ell_0$ formulations}\label{sec:L0sims}
\begin{figure*}[]
%\hspace*{-1in}
	\begin{subfigure}[t]{0.49\textwidth}
		\centering
			\includegraphics[scale=0.22]{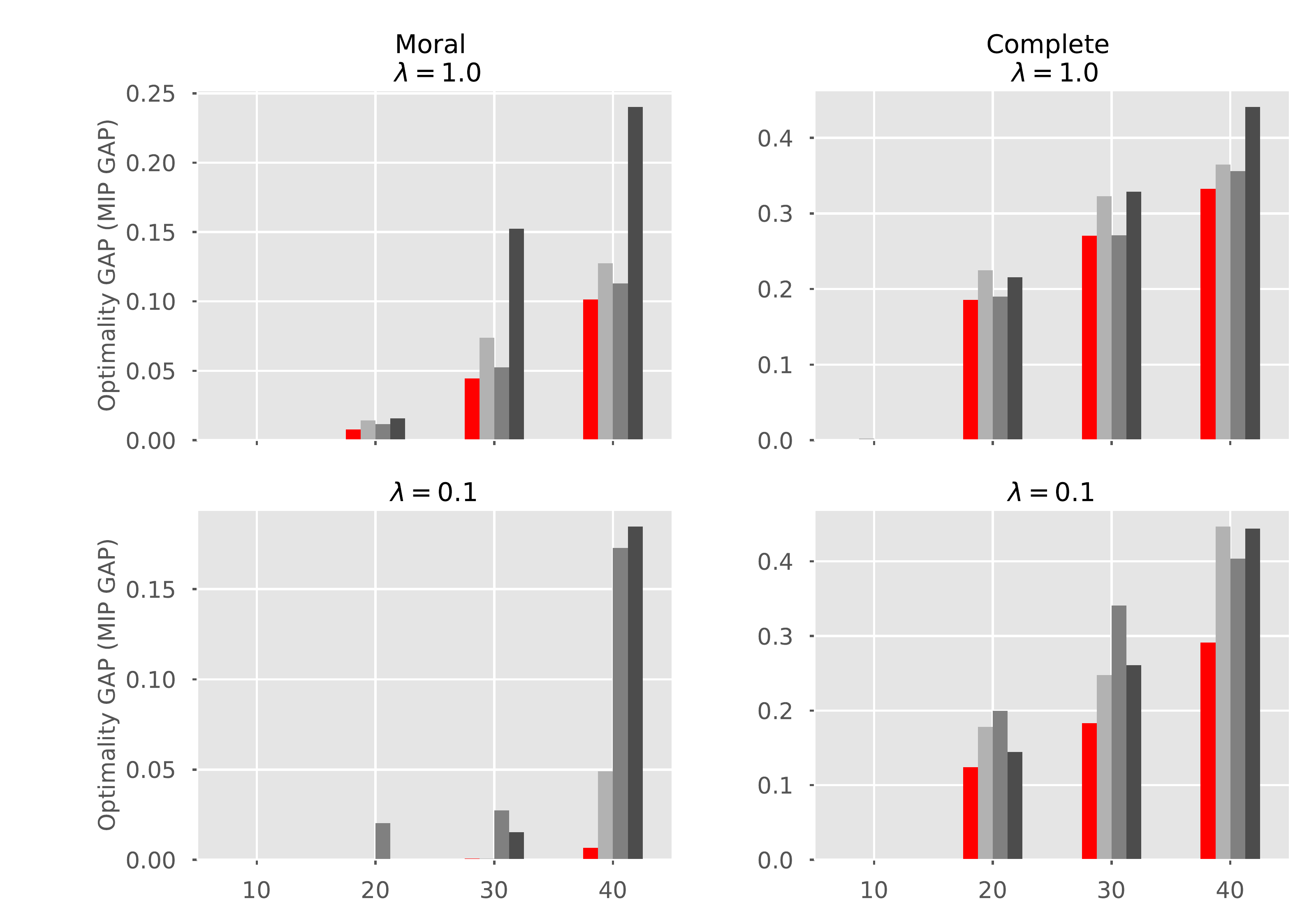}
		\caption{Optimality GAPs for MIQPs}
	\end{subfigure}%
	~ 
	\begin{subfigure}[t]{0.49\textwidth}
		\centering
		\includegraphics[scale=0.22]{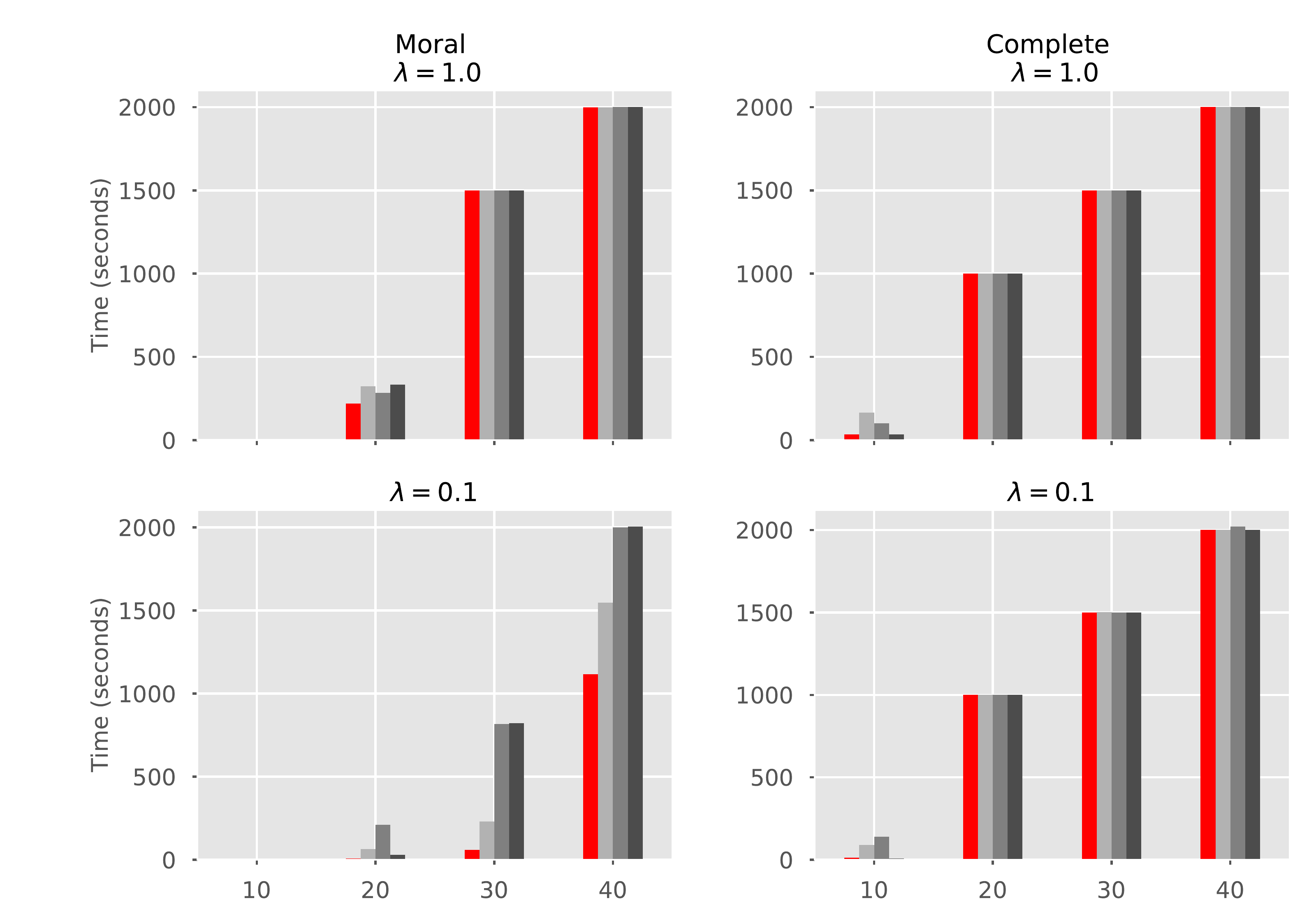}
		\caption{Time (in seconds) for MIQPs}
	\end{subfigure}
	~
%	\hspace*{-1in}
		\begin{subfigure}[t]{0.49\textwidth}
			\centering
			\includegraphics[scale=0.22]{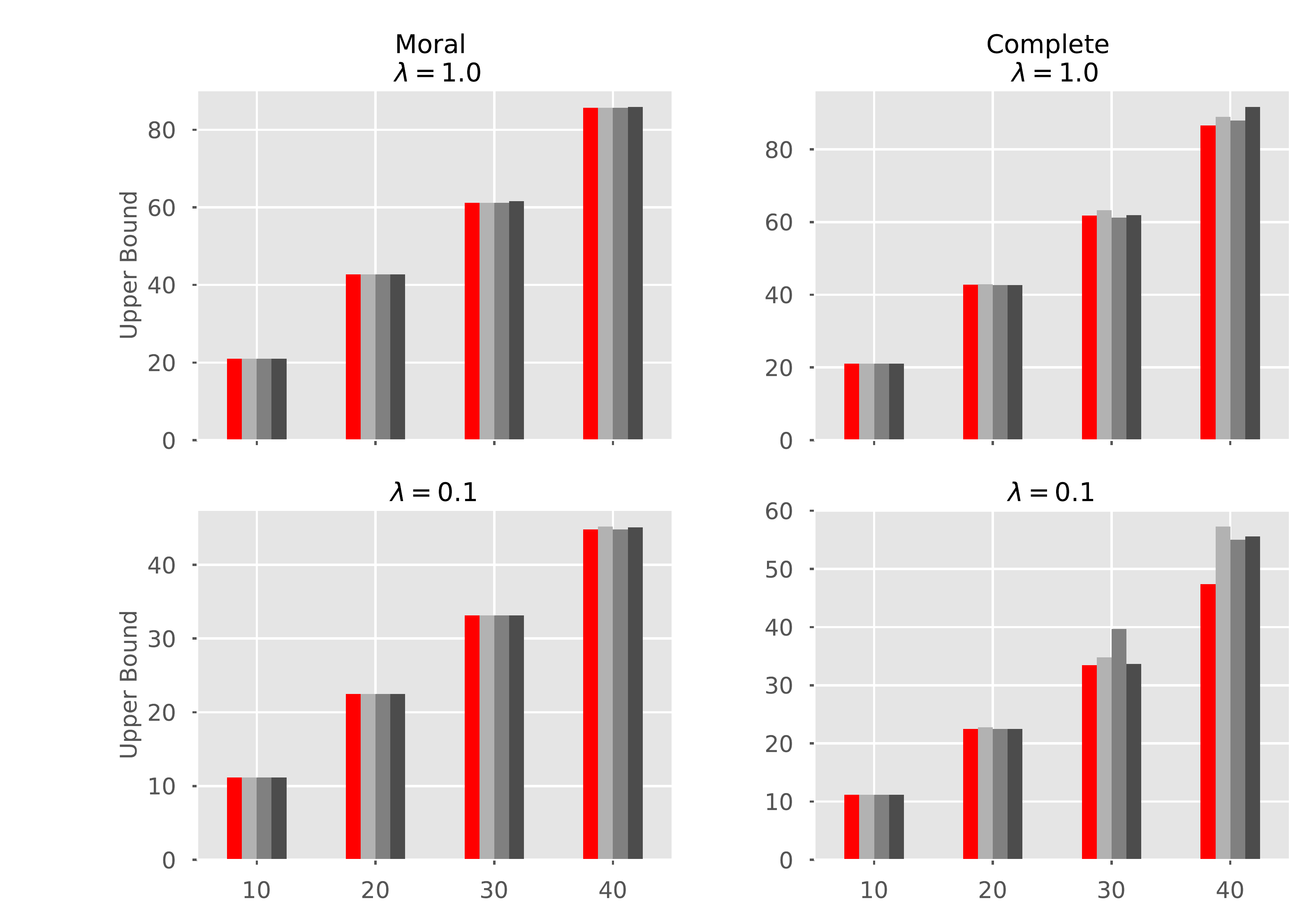}
			\caption{Best upper bounds for MIQPs}
		\end{subfigure}
			~ 
			\begin{subfigure}[t]{0.49\textwidth}
				\centering
				\includegraphics[scale=0.22]{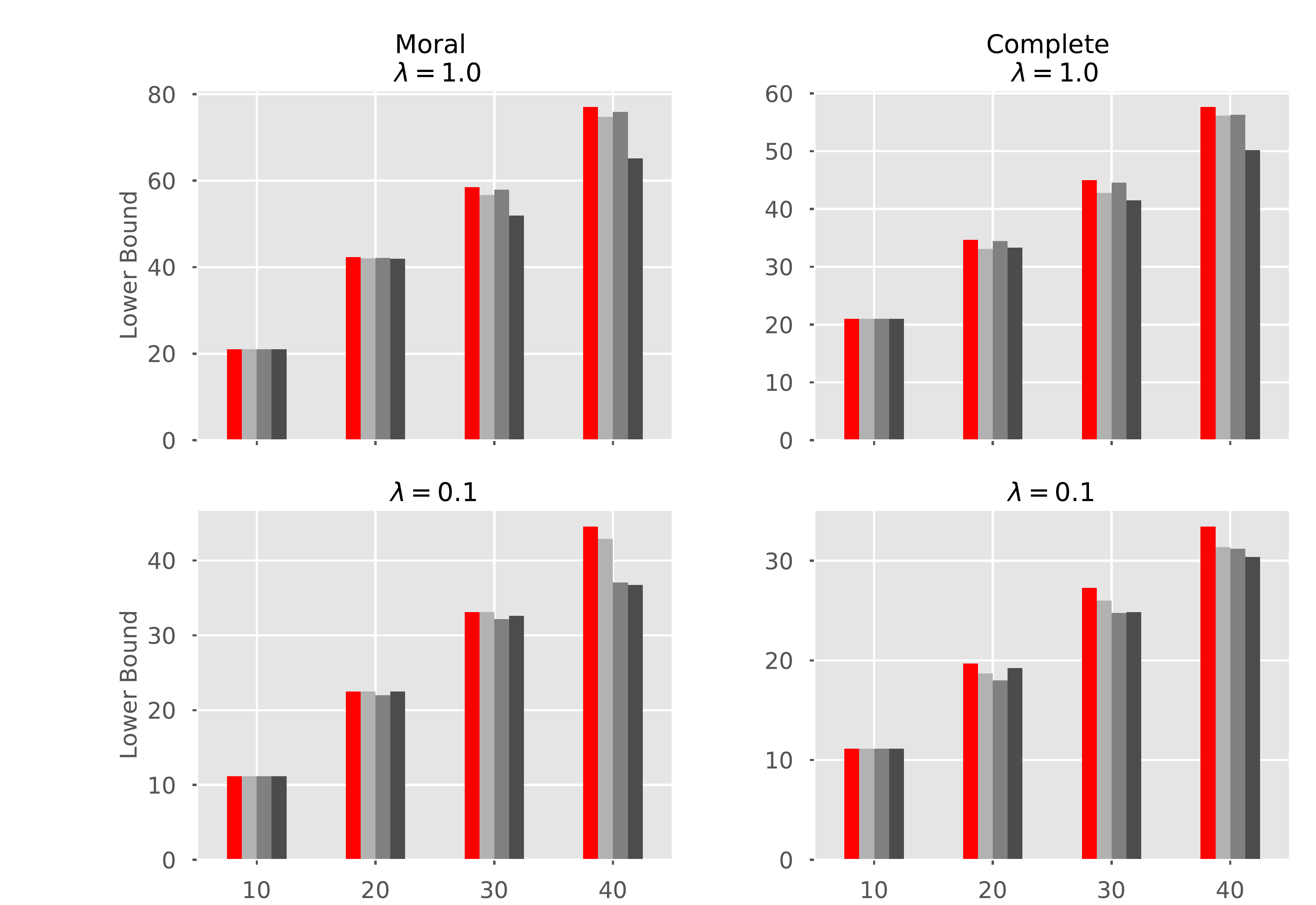}
				\caption{Best lower bounds for MIQPs}
			\end{subfigure}
			~
%				\hspace*{-1in}
					\begin{subfigure}[t]{0.49\textwidth}
						\centering
						\includegraphics[scale=0.22]{L01000Time_seconds_}
						\caption{Time (in seconds) for continuous root relaxation}
					\end{subfigure}
					~ 
%					\hspace{0.2in}
					\begin{subfigure}[t]{0.49\textwidth}
						\centering
						\includegraphics[scale=0.22]{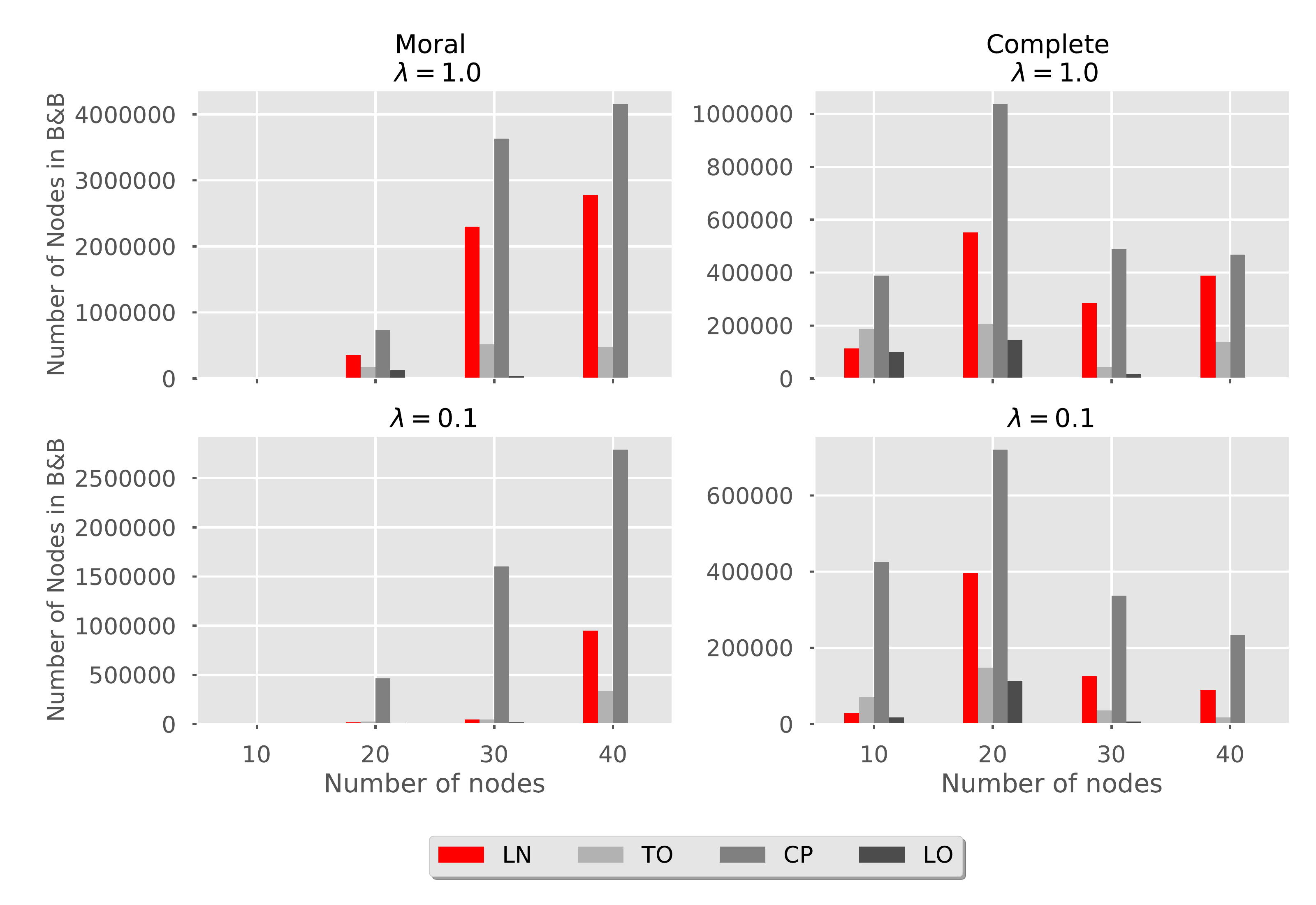}
						\caption{Number of explored nodes in B\&B tree}
					\end{subfigure}
	\caption{Optimization-based measures for MIQPs for $\ell_0$ regularization with the number of samples $n=1000$.}
	\label{Figure: IP_1000}
\end{figure*}

Figure~\ref{Figure: IP_1000} reports the average metrics {across} %\as{why over? didn't you say you generated 10?} \hm{Exactly 10. I removed the word ``over".  By over I meant ``across".} 
10 random graphs for $\ell_0$ formulations with $n=1000$. %We use red color to highlight the results for the LN formulation. 
The LO formulation fails to attain a reasonable solution for one graph (out of 10) with $m=40$ and $\lambda \in \{0.1,1\}$. This is due to the large computation time for solving its continuous relaxation. We excluded these two instances from LO results.     

Figure~\ref{Figure: IP_1000}(a) shows that the LN formulation outperforms other formulations in terms of the average optimality gap across all number of nodes $m \in \{10,20,30,40\}$ and regularization parameters, $\lambda \in \{0.1,1\}$. The difference becomes more pronounced for moral instances. For moral instances, the number of binary variables and constraints for LN solely depends on the size of moral graph. Figure~\ref{Figure: IP_1000}(b) also indicates that the LN formulation requires the least computational time for small instances, whereas all models hit the time limit for larger instances. 

Figures~\ref{Figure: IP_1000}(c)-(d) show the performance of all methods in terms of their upper and lower bounds. For easier instances (e.g., complete instances with $m \in \{10,20\}$ and moral instances), all methods attain almost the same upper bound. Nonetheless, LN performs better in terms of improving the lower bound. For more difficult instances, LN  outperforms other methods in terms of attaining a smaller upper bound (feasible solution) and a larger lower bound.

Figures~\ref{Figure: IP_1000}(e)-(f) show the continuous relaxation time of all models, and the number of explored nodes in the branch-and-bound tree, respectively. The fastest computational time for the continuous relaxation is for the TO formulation followed by the LN formulation. However, the number of explored nodes provides more information about the performance of  mathematical formulations. In small instances, i.e., $m=10$, where an optimal solution is attained, the size of the branch-and-bound tree for the LN formulation is smaller than the TO formulation. This is because the TO formulation has a larger number of binary variables, leading to a larger branch-and-bound tree. On the other hand, for large instances, the number of explored nodes in the LN formulation is larger than the TO formulation. This implies that the LN formulation explores more nodes in the branch-and-bound tree given a time limit. This may be because continuous relaxations of the LN formulation are easier to solve in comparison to the continuous relaxations of the TO formulation in the branch-and-bound process. As stated earlier, the branch-and-bound algorithm needs to explore multiple symmetric formulations in the TO formulation as it branches on fractional topological ordering variables. This degrades the performance of the TO formulation. The LO formulation is very slow because its continuous relaxation becomes cumbersome as the number of nodes, $m$, increases. Thus, we can see a substantial decrease in the number of explored nodes in branch-and-bound trees associated with the LO formulation. The CP formulation is implemented in a cutting-plane fashion. Hence, its number of explored nodes is not directly comparable with other formulations.     

Figures~\ref{Figure: IP_1000}(a)-(f) show the importance of incorporating available structural knowledge (e.g., moral graph). The average optimality gap and computational time are substantially lower for moral instances compared to complete instances. Moreover, the substantial difference in the optimality gap elucidates the importance of incorporating structural knowledge. Similar results are obtained for $n=100$ samples; see Appendix II.  

We next discuss the performance of different methods in terms of estimating the true DAG. The choice of tuning parameter $\lambda$, the number of samples $n$, and the quality of the best feasible solution (i.e., upper bound) influence the resulting DAG. Because our focus in this paper is on computational aspects, we fixed the values of $\lambda$ for a fair comparison between the formulations, and used  $\lambda=0.1$ based on results in preliminary experiments. Thus, we focus on the impact of sample size as well as the quality of the feasible solution {in the explanation of our results}.  

\begin{figure*}[]
	\centering
%	\hspace*{-0.4in}
	\begin{subfigure}[t]{0.49\textwidth}
		\centering
		\includegraphics[scale=0.22]{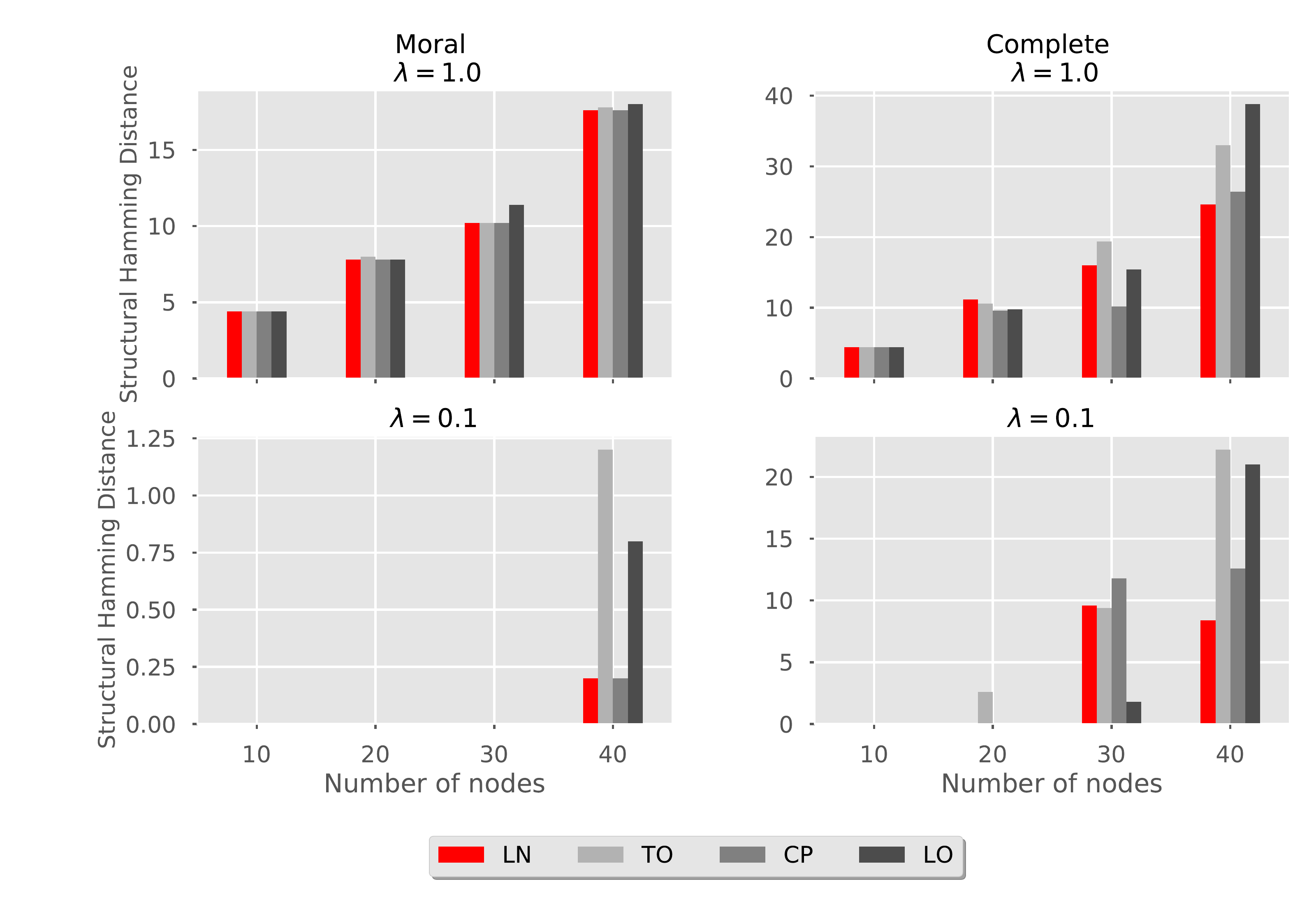} 
		\caption{$n = 1000$}
	\end{subfigure}%
	~ 
	\begin{subfigure}[t]{0.49\textwidth}
		\centering
		\includegraphics[scale=0.22]{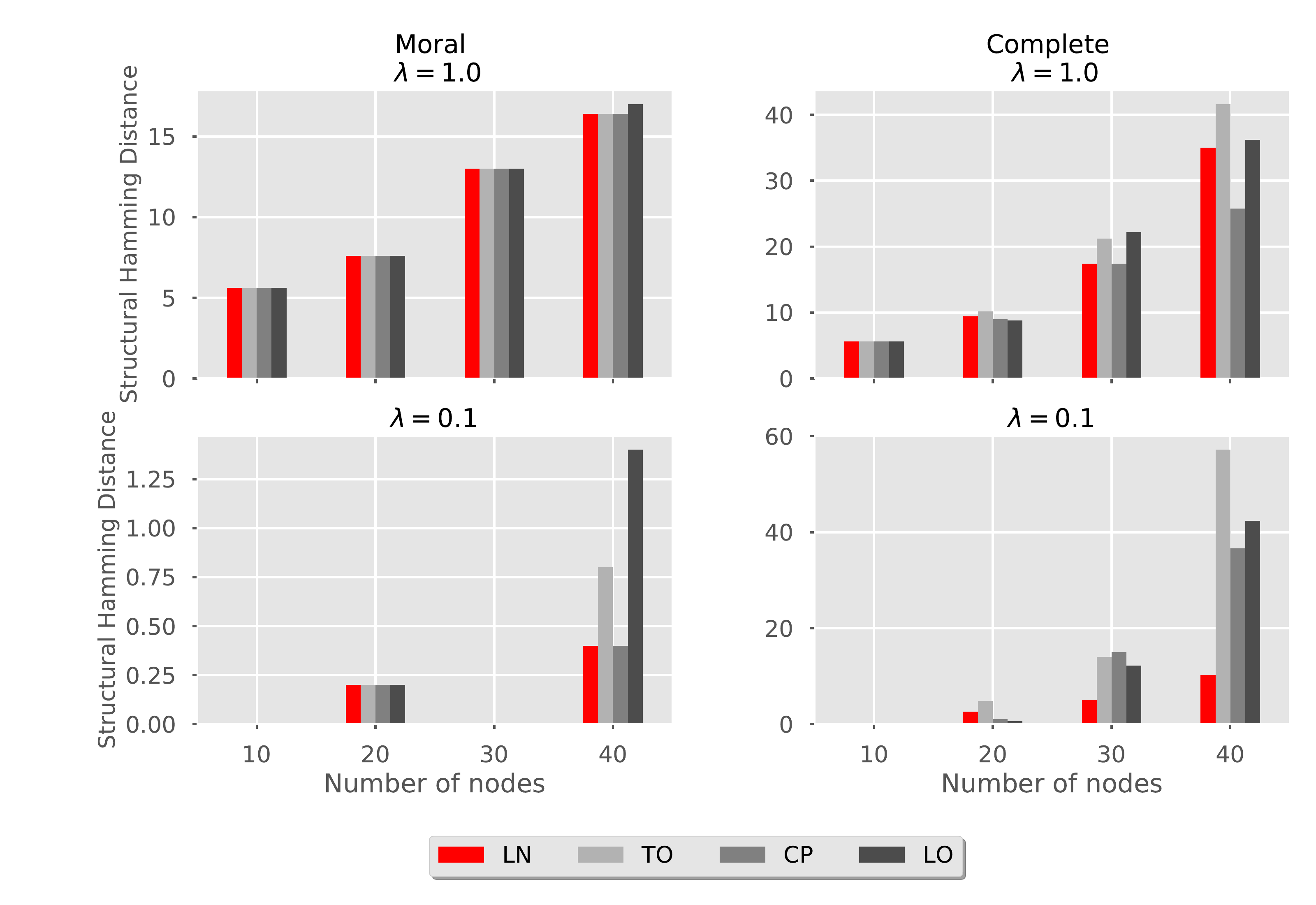}
		\caption{$n = 100$}
	\end{subfigure}
	~
%	\hspace*{-1in}
%	\begin{subfigure}[t]{0.6\textwidth}
%		\centering
%		\includegraphics[scale=0.25]{Fig/L0_1000/TruePositiveRate(TPR)}
%		\caption{True Positive Rate (TPR) for MIQPs}
%	\end{subfigure}
%	~ 
%	\begin{subfigure}[t]{0.5\textwidth}
%		\centering
%		\includegraphics[scale=0.25]{Fig/L0_100/TruePositiveRate(TPR)}
%		\caption{True Positive Rate(TPR) for MIQPs}
%	\end{subfigure}
%	~
%	\hspace*{-1in}
%	\begin{subfigure}[t]{0.6\textwidth}
%		\centering
%		\includegraphics[scale=0.25]{Fig/L0_1000/FalsePositiveRate(FPR)}
%		\caption{False Positive Rate(FPR) for MIQPs}
%	\end{subfigure}
%	~ 
%	\hspace{0.2in}
%	\begin{subfigure}[t]{0.5\textwidth}
%		\centering
%		\includegraphics[scale=0.25]{Fig/L0_100/FalsePositiveRate(FPR)}
%		\caption{False Positive Rate(FPR) for MIQPs}
%	\end{subfigure}
	\caption{Structural Hamming Distance (SHD) of MIQP estimates with $\ell_0$ regularization. }%(a) $n=1000$ samples; (b) $n=100$ samples.}
	\label{Figure: shdGraph_1000}
\end{figure*}

Figures~\ref{Figure: shdGraph_1000}(a)-(b) show the SHDs for all formulations for $n=1000$ and $n=100$, respectively. Comparing Figure~\ref{Figure: shdGraph_1000}(a) with Figure~\ref{Figure: shdGraph_1000}(b), we observe that the SHD tends to increase as the number of samples decreases. As discussed earlier, when $n \rightarrow \infty $, penalized likelihood likelihood estimate with an $\ell_0$ regularization ensures identifiability in our setting  \cite{peters2013identifiability, van2013ell}. However, for a finite sample size, identifiability may not be guaranteed. Moreover, the appropriate choice of $\lambda$ for $n=100$ may be different than the corresponding $\lambda$ for $n=1000$.  

Figure~\ref{Figure: shdGraph_1000}(a) shows that all methods learn the true DAG with $\lambda=0.1$, and given a moral graph for $m \in \{10, 20, 30\}$. In addition, SHD is negligible for LN and CP formulations for $m=40$. However, we observe a substantial increase in SHD (e.g., from 0.2 to near 10 for LN) for complete graphs. %Figure \ref{Figure: Graph_1000} (c)-(f) demonstrate similar patters for TRP and FPR. 
These figures indicate the importance of incorporating available structural knowledge (e.g., a moral graph) for better estimation of  the true DAG.

While, in general,  LN  performs well  compared with other formulations, we do not expect to see a clear dominance in terms of  accuracy of  DAG estimation either due to finite samples or the fact that none of the methods could attain a global optimal solution for larger instances. On the contrary, we retrieve the true DAG for smaller graphs for which optimal solutions are obtained. As pointed out in \cite{park2017bayesian}, a slight change in the objective function value could significantly alter the estimated DAG. Our results corroborate this observation.

\subsection{Comparison of $\ell_1$ formulations}\label{sec:L1sims}
\begin{figure*}[]
	%\hspace*{-0.2in}
	\begin{subfigure}[t]{0.49\textwidth}
		\centering
		\includegraphics[scale=0.22]{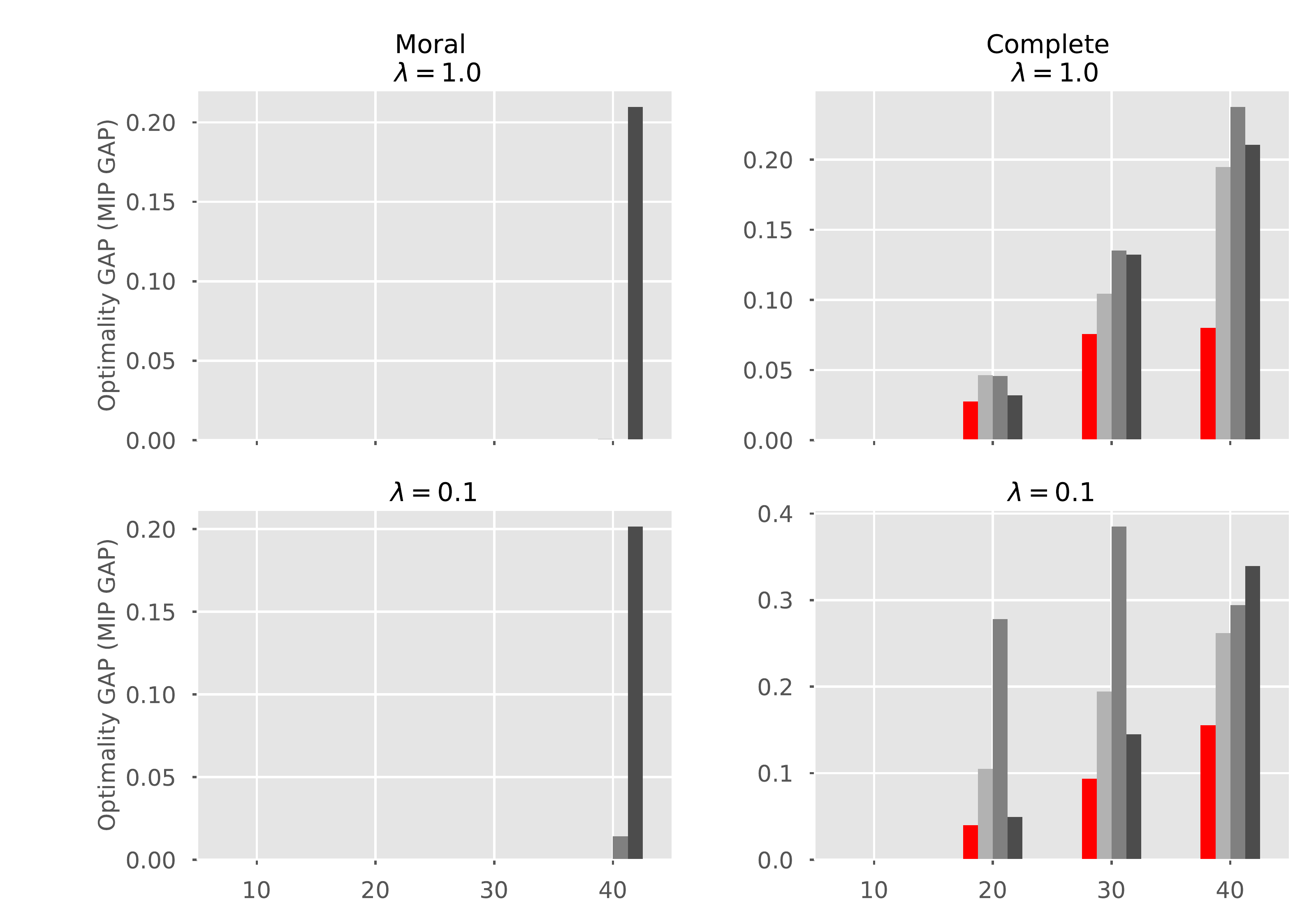}
		\caption{Optimality GAPs for MIQPs}
	\end{subfigure}%
	~ 
	%\hspace*{-0.5in}
	\begin{subfigure}[t]{0.49\textwidth}
		\centering
		\includegraphics[scale=0.22]{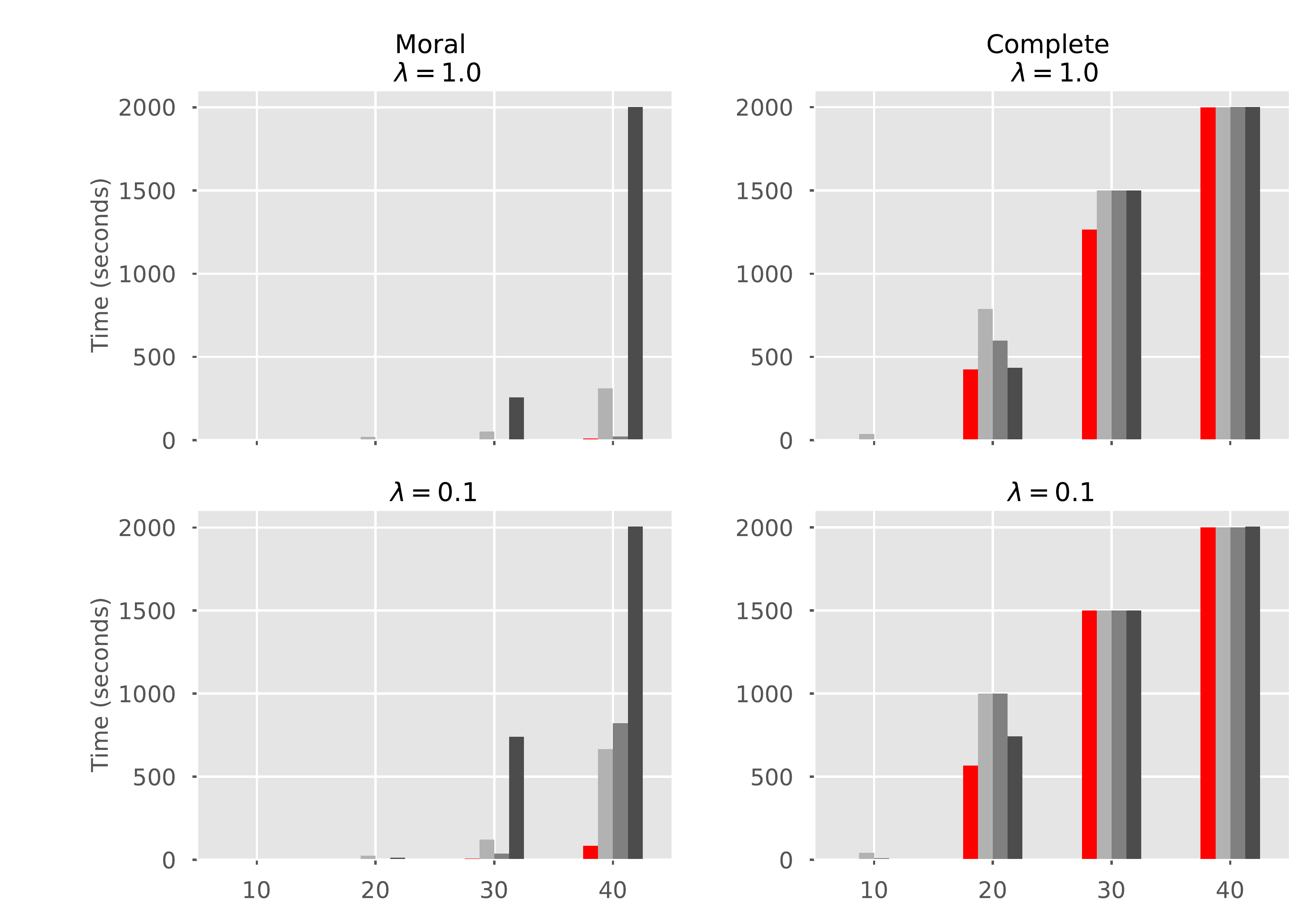}
		\caption{Time (in seconds) for MIQPs}
	\end{subfigure}
	~
	%\hspace*{-1in}
	\begin{subfigure}[t]{0.49\textwidth}
		\centering
		\includegraphics[scale=0.22]{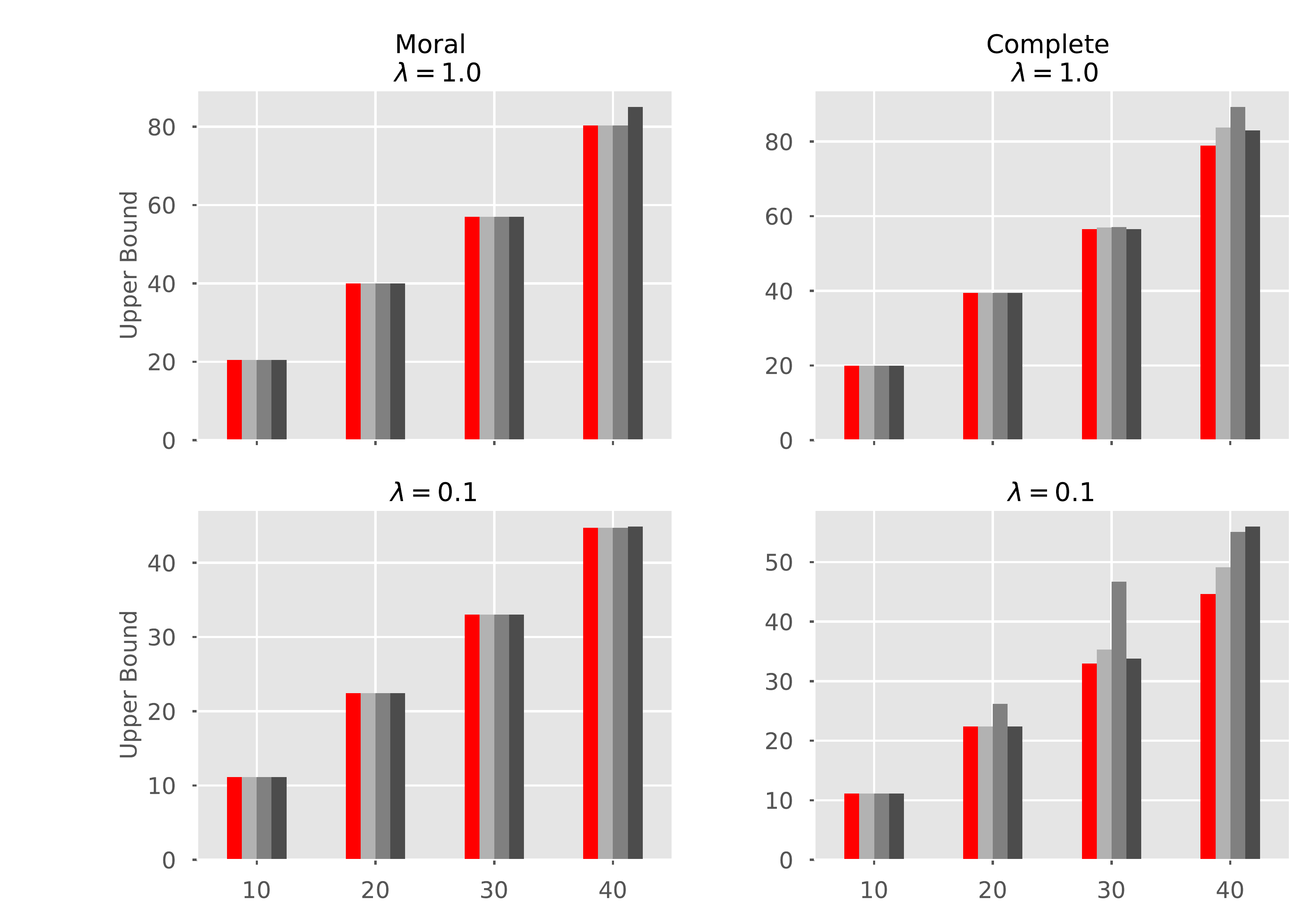}
		\caption{Best upper bounds for MIQPs}
	\end{subfigure}
	~ 
	\begin{subfigure}[t]{0.49\textwidth}
		\centering
		\includegraphics[scale=0.22]{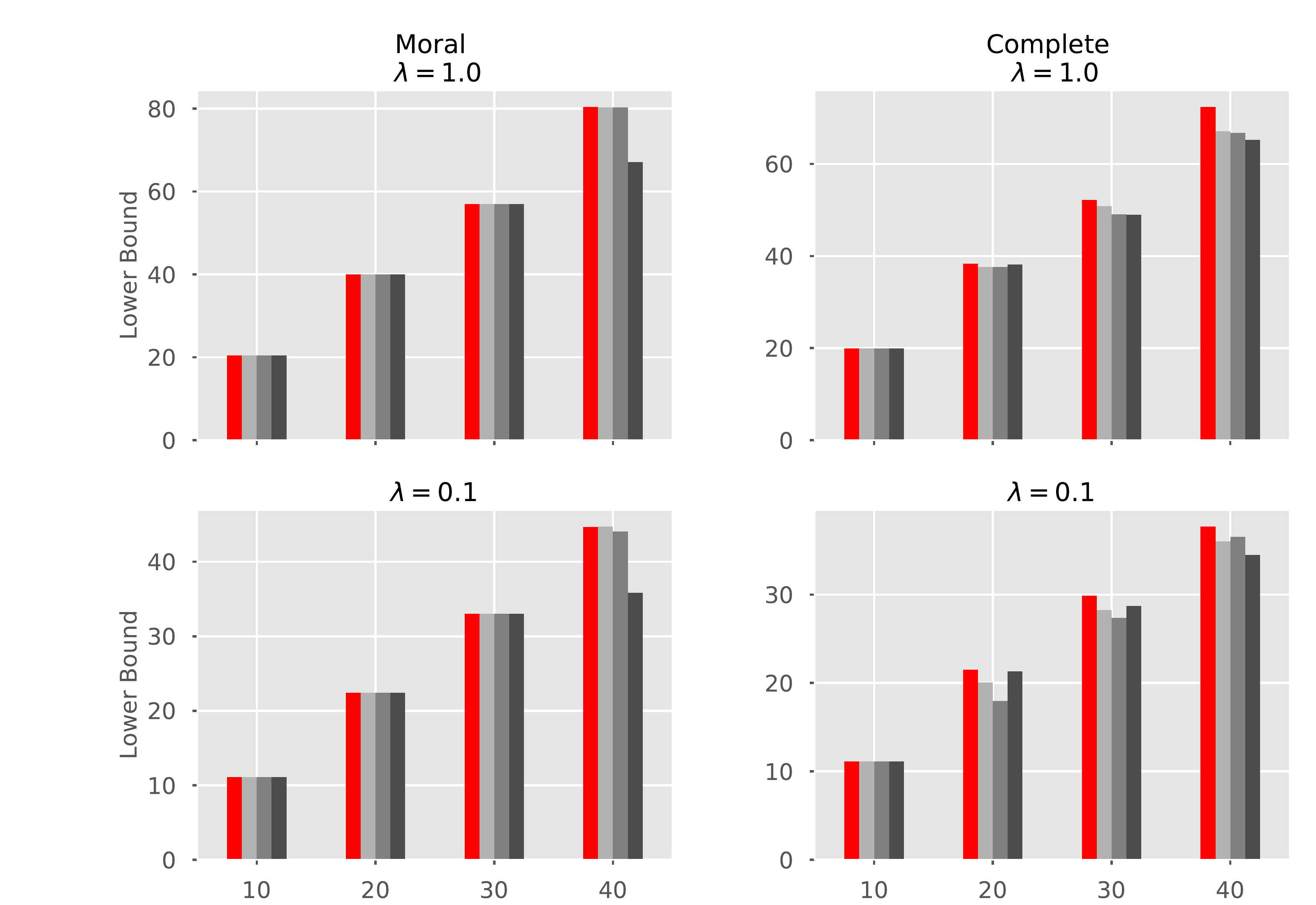}
		\caption{Best lower bounds for MIQPs}
	\end{subfigure}
	~
	%\hspace*{-1in}
	\begin{subfigure}[t]{0.49\textwidth}
		\centering
		\includegraphics[scale=0.22]{L11000Time_seconds_}
		\caption{Time (in seconds) for continuous root relaxation}
	\end{subfigure}
	~ 
	%\hspace{0.2in}
	\begin{subfigure}[t]{0.49\textwidth}
		\centering
		\includegraphics[scale=0.22]{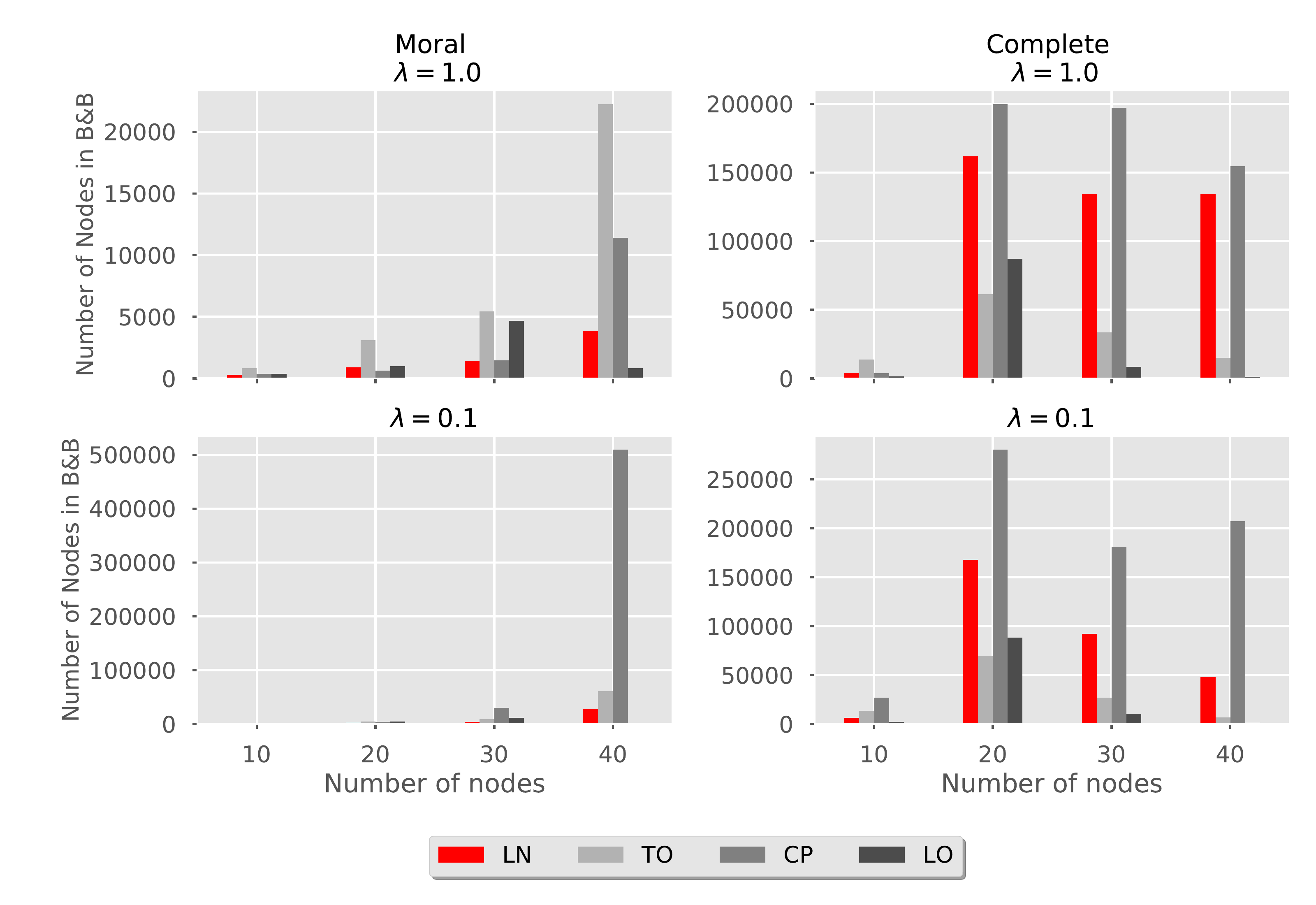}
		\caption{Number of explored nodes in B\&B tree}
	\end{subfigure}
	
	\caption{Optimization-based measures for MIQPs for $\ell_1$ regularization with the number of samples $n=1000$.}
	\label{Figure: IP_L1_1000}
\end{figure*}

Figure~\ref{Figure: IP_L1_1000} shows various average metrics {across} 10 random graphs for $\ell_1$ regularization with $n = 1000$ samples. Figure~\ref{Figure: IP_L1_1000}(a) shows that the LN formulation clearly outperforms other formulations in terms of average optimality gap across all number of nodes, $m \in \{10,20,30,40\}$, and regularization parameters, $\lambda \in \{0.1,1\}$. Moreover, Figure~\ref{Figure: IP_L1_1000}(b) shows that the LN formulation requires significantly less computational time in moral instances, and in complete instances with $m \in \{10, 20\}$ compared to other methods. In complete instances, all methods hit the time limit for $m \in \{30, 40\}$. Figures~\ref{Figure: IP_L1_1000}(c)-(f) can be interpreted similar to the Figures~\ref{Figure: IP_1000}(c)-(f) for $\ell_0$ regularization. Similar to $\ell_0$ regularization, Figures~\ref{Figure: IP_L1_1000}(a)-(b) demonstrate the importance of incorporating structural knowledge (e.g., a moral graph) for $\ell_1$ regularization. Similar results are observed for $n=100$ samples; see Appendix II.

As expected, the DAG estimation accuracy with $\ell_1$ regularization is inferior to the $\ell_0$ regularization. This is in part due to the bias associated with the $\ell_1$ regularization, which could be further controlled with, for example, adaptive $\ell_1$-norm regularization \cite{zou2006adaptive}. Nonetheless, formulations for $\ell_1$ regularization require less computational time and are easier to solve than the corresponding formulations for $\ell_0$ regularization.

\begin{figure*}[]
	%\hspace*{-1in}
	\begin{subfigure}[t]{0.49\textwidth}
		\centering
		\includegraphics[scale=0.22]{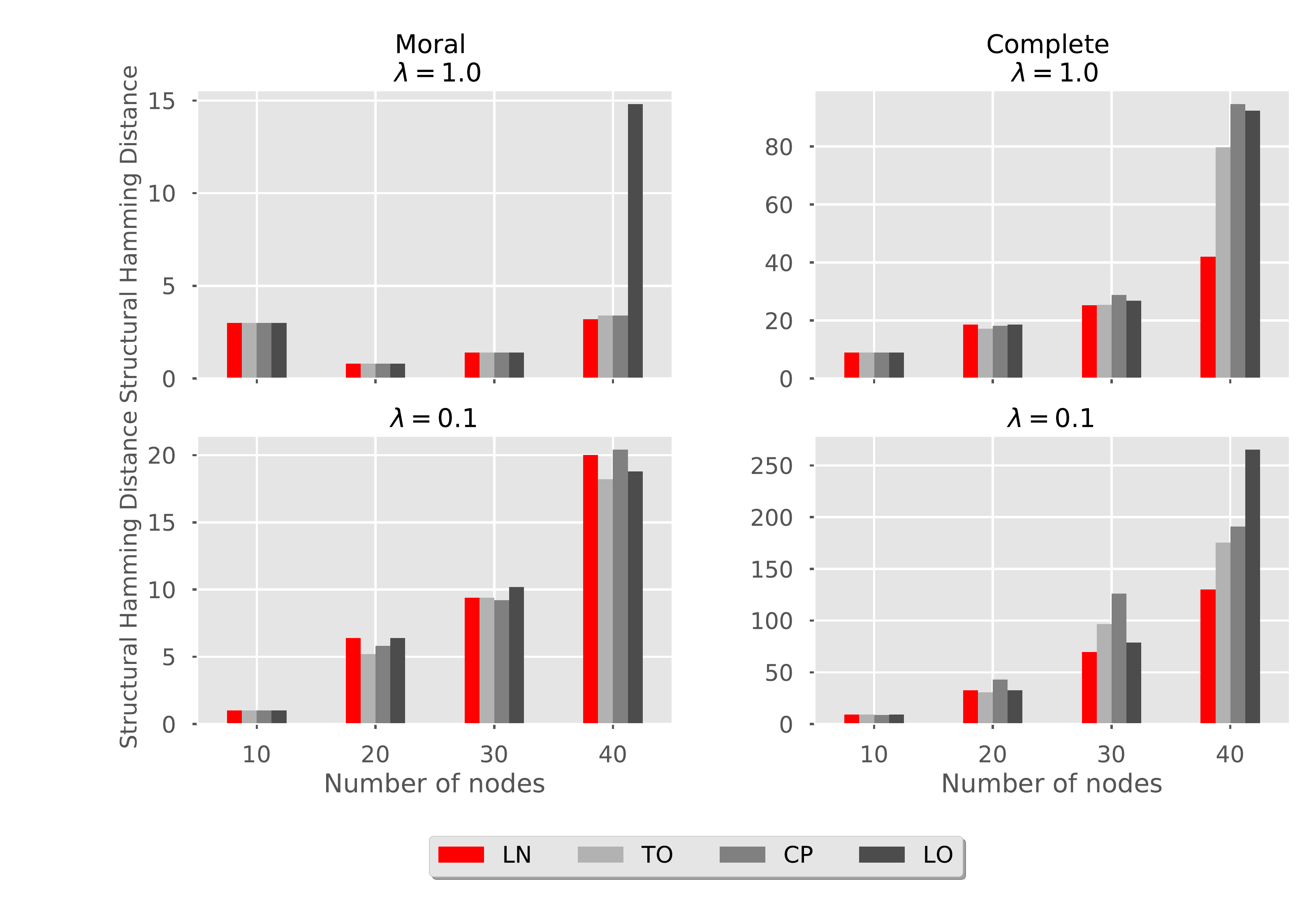}
		\caption{$n=1000$ samples}
	\end{subfigure}%
	~ 
	\begin{subfigure}[t]{0.49\textwidth}
		\centering
		\includegraphics[scale=0.22]{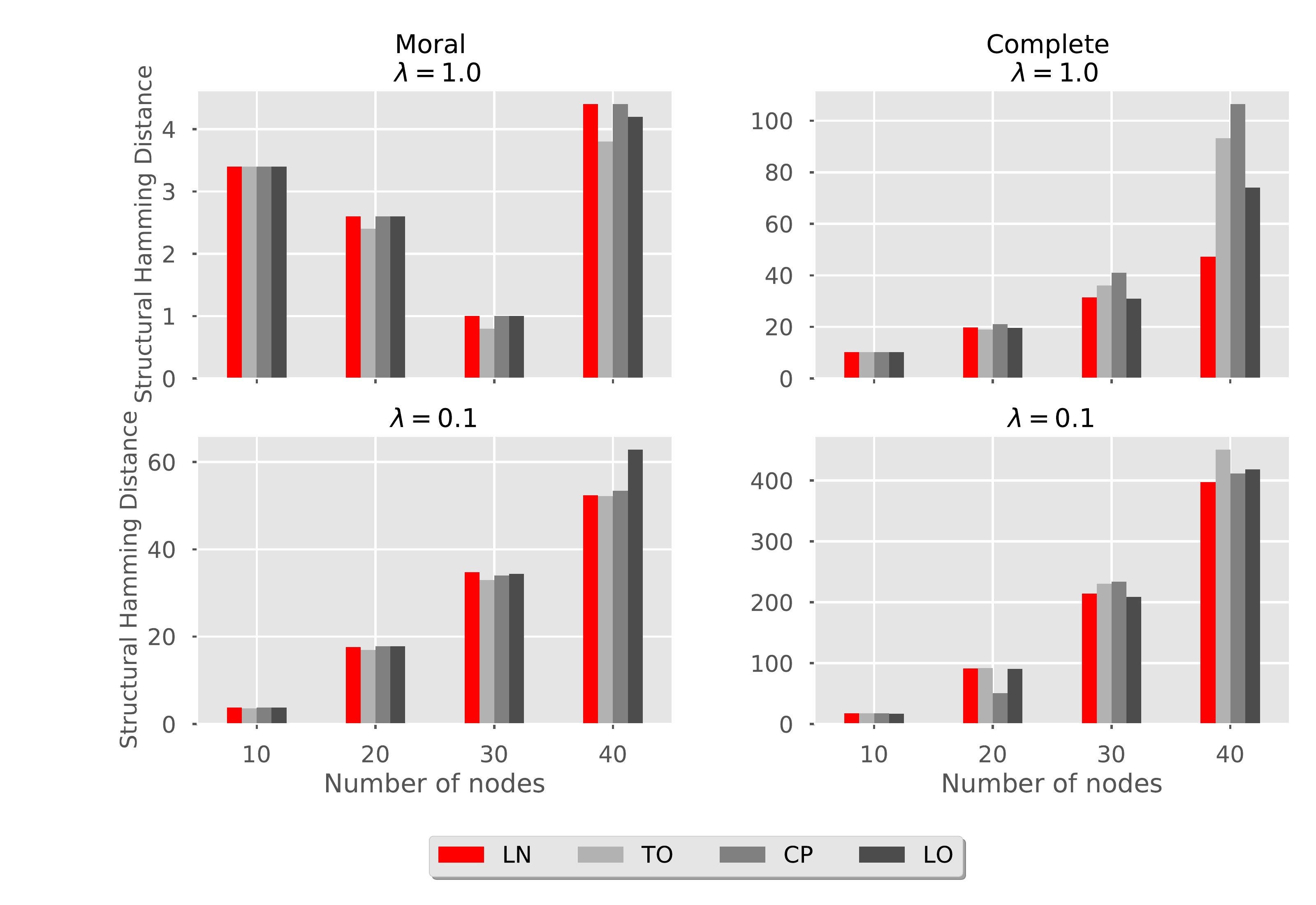}
		\caption{$n=100$ samples}
	\end{subfigure}
	~
%	\hspace*{-1in}
%	\begin{subfigure}[t]{0.6\textwidth}
%		\centering
%		\includegraphics[scale=0.25]{Fig/L1_1000/TruePositiveRate(TPR)}
%		\caption{True Positive Rate (TPR) for MIQPs}
%	\end{subfigure}
%	~ 
%	\begin{subfigure}[t]{0.5\textwidth}
%		\centering
%		\includegraphics[scale=0.25]{Fig/L1_100/TruePositiveRate(TPR)}
%		\caption{True Positive Rate (TPR) for MIQPs}
%	\end{subfigure}
%	~
%	\hspace*{-1in}
%	\begin{subfigure}[t]{0.6\textwidth}
%		\centering
%		\includegraphics[scale=0.25]{Fig/L1_1000/FalsePositiveRate(FPR)}
%		\caption{False Positive Rate (FPR) for MIQPs}
%	\end{subfigure}
%	~ 
%	\hspace{0.2in}
%	\begin{subfigure}[t]{0.5\textwidth}
%		\centering
%		\includegraphics[scale=0.25]{Fig/L1_100/FalsePositiveRate(FPR)}
%		\caption{False Positive Rate (FPR) for MIQPs}
%	\end{subfigure}
%	
	\caption{Structural Hamming Distance (SHD) of MIQP estimates with $\ell_1$ regularization.}
	\label{Figure: shdGraph_L1_100}
\end{figure*}

\section{Comparison with the A$^{\star}$-lasso algorithm}\label{sec:simAstar}
In this section, we compare the LN formulation with A$^{\star}$-lasso \cite{xiang2013lasso}, using the   MATLAB code made available by the authors. For this comparison, the same true DAG structures are taken from \cite{xiang2013lasso} and the strength of arcs ($\beta$) are chosen from $\mathcal{U}[-1,-0.1] \cup \mathcal{U}[0.1,1]$. The number of nodes in the 10 true DAGs varies from $m=6$ to $m=27$ (see Table \ref{Table: Astar comparison}). The true DAG and resulting random $\beta$ coefficients are used to generate $n=500$ samples for each column of data matrix $\mathcal{X}$. %We consider 10 graphs with the number of nodes ranging from $m=6$ to $m=27$ (see, Table \ref{Table: Astar comparison}). 

A time limit of six hours is imposed across all experiments after which runs are aborted. %To be consistent with the model in \cite{xiang2013lasso}, we removed the $n^{-1}$ multiplier from the first term in the objective function \eqref{CP-obj}. 
In addition, for a fairer comparison with  A$^{\star}$-lasso, we do not impose an MIQP gap termination criterion of 0.001 for  LN and  use the  Gurobi  default optimality gap criterion of 0.0001. 

%The DAG used to generate the multivariate data is 
%assumed to be the true structure while it may not be an optimal solution for the objective function. This is particularly important for $\ell_1$ regularization because of the lack of consistency proof. We take this DAG 
%considered as the \textit{ground truth}, and is used to assess the quality of estimates from optimization models. 

%The random Erd{o}s-Renyi graphs have the number of nodes $m\in\{10,20,30,40\}$ and the number of samples $n \in \{100, 1000\}$. The average outgoing degree of each node in these graphs denoted by $d$ is set to be 2. We generate 10 random graphs for each setting ($m$, $n$, $d$). The raw observational data (i.e., $\mathcal{X}$) for the datasets with $n=100$ is the same as first 100 rows of the datasets with $n=1000$. We consider penalty coefficients $\lambda \in \{0.1, 1\}$. 

Similar to synthetic data described in Section \ref{sec:synth-data}, we consider two cases: (i) \textit{moral} instances and (ii) \textit{complete} instances. For the former case, the moral graph is constructed from the true DAG as done   in Section \ref{sec:synth-data}. The raw observational data (i.e., $\mathcal{X}$) for moral and complete instances are the same. %The function \texttt{moralize(graph)} in the \texttt{pcalg} R-package is used to generated the moral graph from the true DAG. The moral graph can also be (consistently) estimated from data using penalized estimation procedures with polynomial complexity %\as{give refs}

We compare the LN formulation with A$^{\star}$-lasso using $\ell_1$ regularization. Note that A$^{\star}$-lasso cannot solve the model with $\ell_0$ regularization. Furthermore, the original A$^{\star}$-lasso algorithm assumes no super-structure. Therefore, to enhance its performance, we 
%We conduct our comparison between A$^{\star}$-lasso and the LN formulation on both complete and moral instances. To this end, we 
modified the MATLAB code for  A$^{\star}$-lasso in order to incorporate the moral graph structure, when available. We consider two values for $\lambda \in \{0, 0.1\}$ for our comparison. As $\lambda$ decreases, identifying an optimal DAG becomes more difficult. Thus, it is of interest to evaluate the computational performance of both methods for $\lambda=0$ (i.e., no regularization) to assess the performance of these approaches on difficult cases (see, e.g., the computational results in \cite{zheng2018dags} and the statistical analysis in \cite{loh2014high} for $\lambda=0$). %The value $\lambda=\ln(n)$ is equivalent to using the Bayesian Information Criterion (BIC) score, which leads to desirable model selection performance as the number of samples increases. 
We note that model selection methods (such as Bayesian Information Criterion \cite{SS2018}) can be used  to identify the best value of $\lambda$. However, in this section, our focus is to  evaluate the computational performance of these approaches for a given $\lambda$ value. 

%Therefore, it is of interest to evaluate the performance of models  For example, the model with $\lambda=0$ can consistently estimate the true DAG \cite{loh2014high} although \cite{loh2014high} has also discussed the consistency of that model with $\lambda>0$ for sparse DAG learning.  
%$I feel we need to mention this in this section, as well. For example a model selection requires testing different $\lambda$ values for each instance. Notably, I feel $\lambda=0$ is indeed an important case. For example, the model with $\lambda=0$ can consistently estimate the true DAG \cite{loh2014high} although \cite{loh2014high} has also discussed the consistency of that model with $\lambda>0$ for sparse DAG learning. Also, \cite{zheng2018dags} used $\lambda=0$ in one of their experiments. Specially, for our purpose $\lambda=0$ makes more sense because it makes the problem harder to solve.}

%\as{how is the moral graph calculated in this case? also, are the data simulated in this case?? what about the graphs??? also, are you using L0 or L1 for LN?} \hm{From true DAGs from \cite{xiang2013lasso}. Construct the moral graph from the True DAG. The beta is simulated. The true DAG is known. We \textit{cannot} use A* for L0 model. Everything is only for L1.}

Table~\ref{Table: Astar comparison} shows the solution times (in seconds) of A$^{\star}$-lasso versus the LN formulation for complete and moral instances. For the LN formulation, if the algorithm cannot prove optimality within the 6-hour time limit, we stop the algorithm and report, in parentheses, the optimality gap at termination.  For  complete instances with $\lambda=0$, the results highlight that for small instances (up to 14 nodes) A$^{\star}$-algorithm performs better, whereas the LN formulation outperforms A$^{\star}$-lasso for larger instances. In particular, we see that the LN formulation attains the optimal solution for the Cloud data set in 810.47 seconds and it obtains a feasible solution that is provably within 99.5\% of the optimal objective value for Funnel and Galaxy data sets.  For moral instances, we observe significant improvement in the computational performance of the LN formulation, whereas the improvement in A$^{\star}$-lasso is marginal in comparison. This observation highlights the fact that dynamic programming-based approaches cannot effectively utilize the  super-structure knowledge,  whereas an IP-based approach, particularly the LN formulation, can significantly reduce the computational times. For instance, LN's computational time for the Cloud data reduces from  $\sim810$ seconds to less than two seconds when the moral graph is provided.
  In contrast,  the reduction in computational time for A$^{\star}$-algorithm with the moral graph is negligible. %The computation time for LN depends on the number of edges in the moral graph. For instance, although both Insurance and Factors datasets have 27 {nodes}, the Factors dataset is much harder to solve because its moral graph has more edges. 
For $\lambda = 0.1$, the problem is easier to solve. Nevertheless, LN performs well in comparison to A$^\star$-lasso and the performance of the LN formulation improves for moral instances. 

For DAG learning from discrete data, an IP-based model (see e.g., \cite{jaakkola2010learning}) outperforms A$^\star$ algorithms when a cardinality constraint on the number of the parent set for each node is imposed; A$^\star$ tends to perform better if such constraints are not enforced. This is mainly because an IP-based model for discrete data requires an exponential number of variables which becomes cumbersome if such cardinality constraints are not permitted. In contrast, for DAG learning from continuous data with linear SEMs, our results show that an IP-based approach does not have such a limitation because variables are encoded in the space of arcs (instead of parent sets). That is why LN performs well even for complete instances (i.e., no restriction on the cardinality of parent set).           
%\as{what about the quality of estimated graphs with A*?} \hm{That requires a careful model selection for $\lambda$ parameter. Our goal (at least in this section) is to determine the global score function. TO be more specific, for small instances, A* and IP models give the same score. Thus, the quality of DAGs are comparable. Our measure is score value. For larger instances, we attain the optimal score in less computational time (or IP attains a better score function given a time limit). If we review related IP papers e.g., Cussens and colleagues, they never compare with true DAG. They only compare with optimal score. One can always do a more complete model selection.}

There are several fundamental advantages of IP-based modeling, particularly the LN formulation, compared to A$^{\star}$-lasso: (i) The variables in IP-based models (i.e., TO, LN, and CP) depend on the super-structure. Therefore, these IP-based models can effectively utilize the prior knowledge to reduce the search space, whereas A$^{\star}$-lasso cannot utilize the super-structure information as effectively. This is particularly important for the LN formulation, as the number of variables and constraints depend only on the super-structure; (ii) all IP-based models can incorporate both $\ell_0$ and $\ell_1$ regularizations, whereas A$^{\star}$-lasso can solve the problem only with $\ell_1$ regularization; (iii) all IP-based methods in general enjoy the versatility to incorporate a wide variety of structural constraints, whereas A$^{\star}$-lasso and dynamic programming approaches cannot accommodate many structural assumptions. For instance, a modeler may prefer restricting the number of arcs in the DAG, this is achievable by imposing a constraint on an IP-based model, whereas one cannot impose such structural knowledge on A$^{\star}$-lasso algorithm; (iv) A$^{\star}$-lasso is based on dynamic programming; therefore, one cannot abrupt the search with the aim of achieving a feasible solution. On the other hand, one can impose a time limit or an optimality gap tolerance to stop the search process in a branch-and-bound tree. The output is then  a feasible solution to the problem, which provides an upper bound as well as a lower bound which guarantees the quality of the feasible solution; (v) algorithmic advances in integer optimization alone (such as faster continuous relaxation solution, heuristics for better upper bounds, and cutting planes for better lower bounds) have resulted in 29,000 factor speedup in solving IPs \cite{bertsimas2016best} using a branch-and-bound process. Many of these advances have been implemented in powerful state-of-the-art optimization solvers (e.g., Gurobi), but they cannot be used in  dynamic programming methods, such as  A$^{\star}$-lasso.

\begin{figure*}[]
%	\begin{subfigure}[t]{0.49\textwidth}
%		\centering
% \includegraphics*[scale=0.22]{Fig/UB_LB/Funnel.pdf}
%		\caption{Funnel dataset with complete graph}
%	\end{subfigure}%
%	~
%		\begin{subfigure}[t]{0.49\textwidth}
%			\centering
%	 \includegraphics*[scale=0.22]{Fig/UB_LB/Funnel.pdf}
%			\caption{Funnel dataset with moral graph}
%		\end{subfigure}% 
%		
%		%\hspace*{-1in}	
%			\begin{subfigure}[t]{0.49\textwidth}
%				\centering
%			 %\includegraphics*[scale=0.25]{Fig/UB_LB/Galaxy.pdf}
%				%\caption{Galaxy dataset with complete graph}
%			\end{subfigure}%
%	~		
%				\begin{subfigure}[t]{0.49\textwidth}
%					\centering
%									%\includegraphics*[scale=0.25]{Fig/UB_LB/Insurance.pdf} 
%					%\caption{Factors dataset with complete graph}
%				\end{subfigure}%
%				
%		\hspace*{-1in}			
					\begin{subfigure}[t]{0.49\textwidth}
						\centering
			\includegraphics*[scale=0.22]{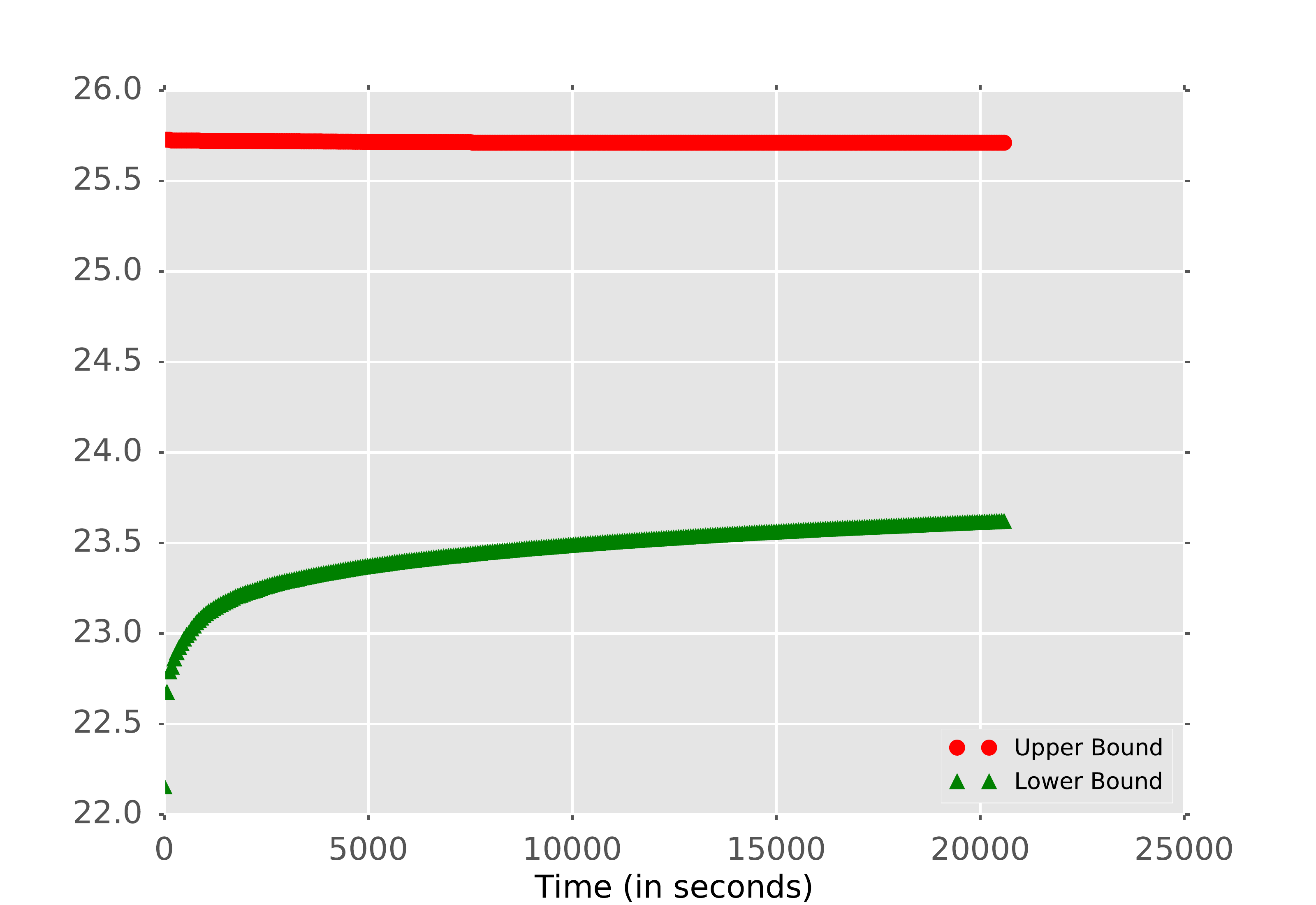}
						\caption{Factors dataset with complete graph}
					\end{subfigure}
					~
						\begin{subfigure}[t]{0.49\textwidth}
							\centering
							\includegraphics*[scale=0.22]{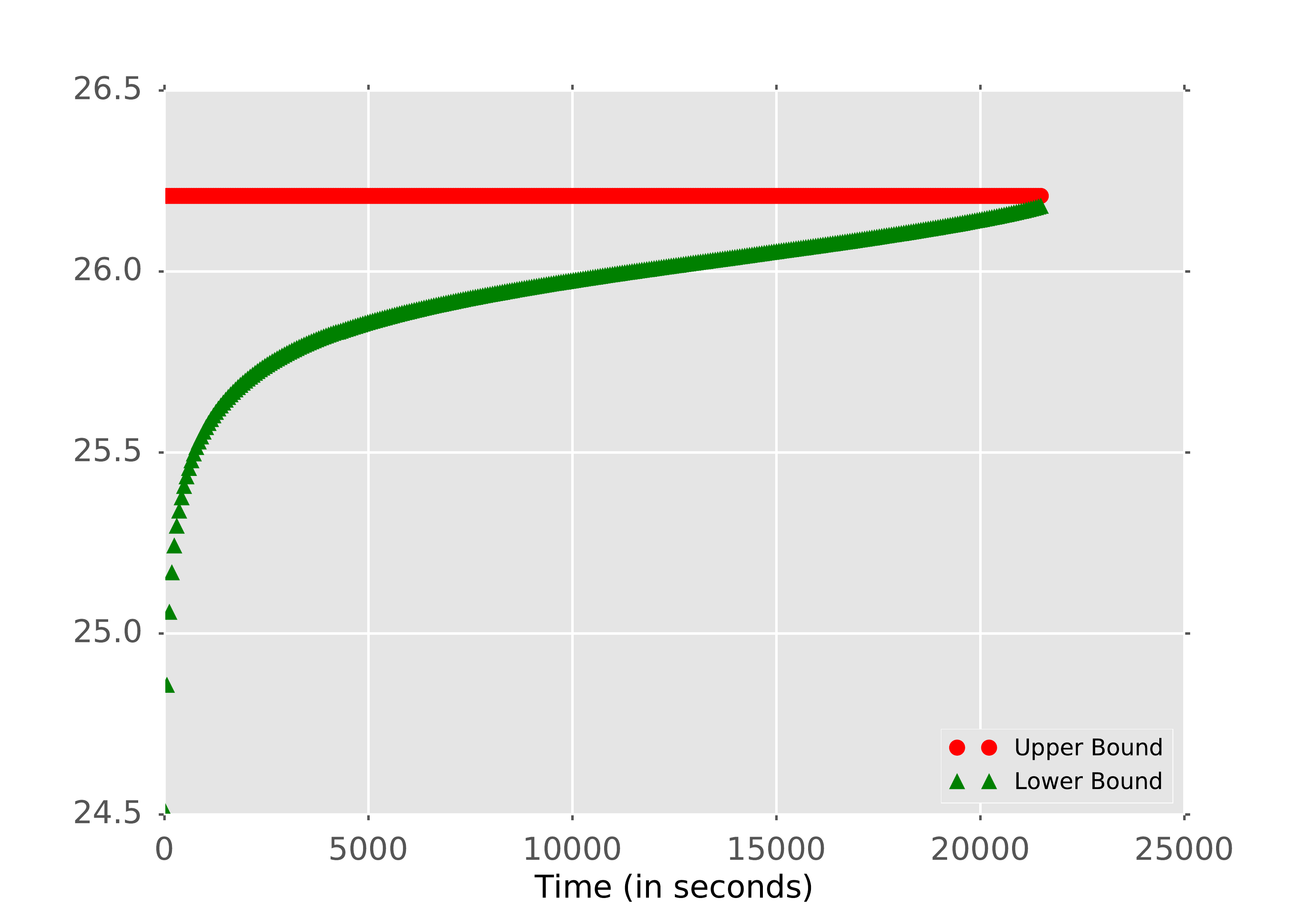} 
							\caption{Factor dataset with {moral graph}}
						\end{subfigure}%
					\caption{The progress of upper bound versus lower bound in the branch-and-bound tree for the LN formulation with no regularization (i.e., $\lambda=0$).}
				\label{fig: upperbound}
\end{figure*}

Figures \ref{fig: upperbound}(a)-(b) illustrate the progress of upper bound versus lower bound in the branch-and-bound process for the Factor dataset and highlight an important practical implication of an IP-based model: such models often attain high quality upper bounds (i.e., feasible solutions) in a short amount of time whereas the rest of the time is spent to close the optimality gap by increasing the lower bound. %Figure \ref{fig: upperbound} (a-c) are for complete instances whereas \ref{fig: upperbound} (d) is for Factors dataset with moral instances. 

The results for the moral graph in {Figure} \ref{fig: upperbound}(b) highlight another important observation. That is, providing the moral graph can significantly accelerate the progress of the lower bound  in the branch-and-bound process. In other words, information from the moral graph helps close the optimality gap more quickly.   

%\begin{table}[]
%	\caption{Computational performance of LN versus A$^{\star}$-algorithm {for $\ell_1$ regularization} with $\{\lambda=0, \log(500)\}$. Numbers in parentheses show the optimality gaps associated with the LN formulation. %\as{why 0??? also what are the instances?} \hm{Instances are dsep, Asia, etc ... Complete instance and Moral instances are those graphs (i.e., dsep, Asia) with complete super-structure or moral super-structure, respectively.}}
%	}
%	\begin{tabular}{lll|lll|lll}	
%\hline
%\multicolumn{ 2}{c}{} & & & \multicolumn{2}{c}{Complete instances} & & \multicolumn{2}{c}{Moral instances} \\ \hline 
%Instances & \textit{m} &  $|\mathcal{M}|$ &  & {A$^{\star}$-lasso} & LN & & {A$^{\star}$-lasso} & LN   \\ \hline
%dsep & 6 & 16 &  & 0.0261 & 0.388 &  & 0.017 & 0.378 \\ 
%Asia & 8 & 40 &  & 0.152 & 0.887 &  & 0.016 & 0.401 \\ 
%Bowling & 9 & 36 &  & 0.467 & 1.31 &  & 0.018 & 0.554 \\ 
%Insurancesmall & 15 & 76 &  & 547.17 & 613.543 &  & 391 & 2.719 \\ 
%Rain & 14 & 70 &  & 101.33 & 246.25 &  & 95.15 & 1.752 \\ 
%Cloud & 16 & 58 &  & 18839 & 810.47 &  & 4433.29 & 1.421 \\ 
%Funnel & 18 & 62 &  & 6 hrs & 6 hrs (0.002) &  & \multicolumn{1}{l}{ 6 hrs} & 1.291 \\ 
%Galaxy & 20 & 76 &  & 6 hrs & 6 hrs (0.005) &  & \multicolumn{1}{l}{ 6 hrs} & 1.739 \\ 
%Insurance & 27 & 168 &  & 6 hrs & 6 hrs (0.162) &  & \multicolumn{1}{l}{ 6 hrs} & 131.74 \\ 
%Factors & 27 & 310 &  & 6 hrs & 6 hrs (0.081) &  & \multicolumn{1}{l}{6 hrs} & \multicolumn{1}{l}{6 hrs (0.001)} \\ \hline 
%	\end{tabular}
%	\label{Table: Astar comparison}
%\end{table}

\begin{table}[htbp]
	%\fontsize{4}{20}\selectfont
		\begin{center}
			\caption{Computational performance of LN versus A$^{\star}$-algorithm {with $\ell_1$ regularization} for $\lambda \in \{0, 0.1\}$}
			\resizebox{\textwidth}{!}{
	\begin{tabular}{lll|lll|llll||lll|lll}
		\hline 
	%	&  & & & \multicolumn{5}{c}{$\lambda=0$}  &  & \multicolumn{5}{c}{$\lambda=\ln(500)$} \\ \hline
		&  &  &  & \multicolumn{2}{c|}{Moral  $\lambda=0$} & &  \multicolumn{2}{c}{Complete  $\lambda=0$}  & & & \multicolumn{2}{c|}{Moral  $\lambda=0.1$} &  & \multicolumn{2}{c}{Complete  $\lambda=0.1$}   \\ \hline
		\multicolumn{1}{l}{Graphs (Data sets)} & \textit{m} &  $|\mathcal{M}|$ &  & \multicolumn{1}{l}{A$^\star$-lasso} & \multicolumn{1}{l|}{LN} &  &  \multicolumn{1}{l}{A$^\star$-lasso} & \multicolumn{1}{l}{LN} & & & \multicolumn{1}{l}{A$^\star$-lasso}  &  \multicolumn{1}{l|}{LN} & & \multicolumn{1}{l}{A$^\star$-lasso}  & \multicolumn{1}{l}{LN} \\ \hline
		dsep & 6 & 16 &  & 0.017 & 0.378 &  & 0.0261 & 0.388 & & & 0.455 & 0.025 &  & 0.429 & 0.108 \\ 
		Asia & 8 & 40 &  & 0.016 & 0.401 &  & 0.152 & 0.887 & & & 0.195 & 0.071 &  & 0.191 & 0.319 \\ 
		Bowling & 9 & 36 &  & 0.018 & 0.554 &  & 0.467 & 1.31 & & & 0.417 & 0.225 &  & 0.489 & 0.291 \\ 
		Insurancesmall & 15 & 76 &  & 391 & 2.719 &  & 547.171 & 613.543 & & & 2.694 & 1.135 &  & 3.048 & 0.531 \\ 
		Rain & 14 & 70 &  & 119.15 & 1.752 &  & 101.33 & 246.25 & & & 51.737 & 0.632 &  & 69.404 & 3.502 \\ 
		Cloud & 16 & 58 &  & 4433.29 & 1.421 &  & 18839 & 810.471 & & & 1066.08 & 0.426 &  & 2230.035 & 7.249 \\ 
		Funnel & 18 & 62 &  & 6 hrs & 1.291 &  & 6 hrs & 6 hrs (.002) & & & 6 hrs & 0.395 &  & 6 hrs & 3.478 \\ 
		Galaxy & 20 & 76 &  & 6 hrs & 1.739 &  & 6 hrs & 6 hrs (.005) &  & & 6 hrs & 0.740 &  & 6 hrs & 9.615
		 \\ 
		Insurance & 27 & 168 &  & 6 hrs & 131.741 &  & 6 hrs & 6 hrs (.162) & & & 6 hours & 12.120 &  & 6 hrs & 6 hrs (.031)
		 \\ 
		Factors & 27 & 310 &  & 6 hrs & 6 hrs (.001) &  & 6 hrs & 6 hrs (.081) & & & 6 hours & 55.961 &  & 6 hrs & 6 hrs (.01)
		 \\ \hline 
	\end{tabular}}
			\label{Table: Astar comparison}
		\end{center}
	\end{table}
%\end{table}

\section{Conclusion} \label{Sec: Conclusion}

In this paper, we study the problem of learning an optimal DAG from continuous observational data using  a score function, where the causal effect among the random variables is linear. We cast the problem as a mathematical program and use a penalized negative log-likelihood score function with both $\ell_0$ and $\ell_1$ regularizations. The mathematical programming framework can naturally incorporate a wide range of structural assumptions. For instance, it can incorporate a super-structure (e.g., skeleton or moral graph) in the form of an undirected and possibly cyclic graph. Such super-structures can be estimated from observational data. We review three mathematical formulations: cutting plane (CP), topological ordering (TO), and Linear Ordering (LO), and propose a new mixed-integer quadratic optimization (MIQO) formulation, referred to as the layered network (LN) formulation. We establish that the continuous relaxations of all models attain the same optimal objective function value under a mild condition. Nonetheless, the LN formulation is a compact formulation in contrast to CP, its relaxation can be solved much more efficiently compared to LO, and enjoys a fewer number of binary variables and traces a fewer number of branch-and-bound nodes than TO. 

%Our computational results highlight the following conclusions. 

%\begin{itemize}
%	\item The LN formulation performs better than other MIQP formulations (i.e., CP, TO, LO) in terms of the optimality gap, upper and lower bounds, and computational time. In particular, the LN formulation gives the best performance when a super-structure (e.g., moral graph) is available. This is primarily because the LN formulation is the only formulation in which the number of constraints and variables solely depends on the number of edges in the super-structure. 
%	\item  In contrast to A$^{\star}$-style algorithms, a mathematical model has the versatility to incorporate various structural assumptions and different regularizations terms (e.g., $\ell_0$ and $\ell_1$). Moreover, optimization solvers often attain optimal solutions in short amount of time while the rest of the time is spent on proving the optimality. Such framework allows a user to impose a wide range of termination criteria (e.g., time limit, optimality gap) to stop the algorithm and collect the best available solution with an optimality performance guarantee. In particular, LN performs better than A$^{\star}$-algorithm as the number of nodes increases when there is no prior structural knowledge available. When a moral graph is provided, LN formulation can effectively use these information to reduce the search space and reduce the computational time substantially.  
%\end{itemize}

Our numerical experiments indicate that these advantages result in considerable improvement in the performance of LN compared to other MIQP formulations (CP, LO, and TO). These improvements are particularly pronounced when a sparse super-structure is available, because LN is the only formulation in which the number of constraints and binary variables solely depend on the super-structure. Our numerical experiments also demonstrate that the LN  formulation has a number of advantages over the A$^{\star}$-lasso algorithm, especially when a sparse super-structure is available.   

At least two future research avenues are worth exploring. First, one of the difficulties of estimating DAGs using mathematical programming techniques is the constraints in \eqref{CP-con1}. The big-$M$ constraint is often very loose, which makes the convergence of branch-and-bound process slow. It is of interest to study  these constraints in order to improve the lower bounds obtained from continuous relaxations.  %There has been recent developments to improve these type of constraints in variable selection problem based on  perspective-relaxation \cite{dong2015regularization} as well as semidefinite optimization problems \cite{atamturk2019rank}. It is of interest to see how the discussed formulations can be tightened based on these recent ideas. 
	%\item Given the observation that LN attains high quality solutions in rather short amount of time, it is of great interest to identify other termination criterion instead of MIP GAP which are contingent on probabilistic guarantees of a solution. To this end, the distance between objective function \textit{after early stopping} and the global objective function can be described in a probabilistic manner. Such framework allows a modeler to terminate the solver after the termination criteria is met and obtain a solution with a probabilistic guarantees. 
	%\item The layered network representation can open the opportunity to develop new approximate techniques based on searching in the space of layered networks instead of topological ordering. This is particularly important if a super-structure is provided because the search space can substantially reduced. The challenge is how to effectively search in the space of layered networks.   
	 Second, in many real-world applications, the underlying DAG has special structures. For instance, the true DAG may be a polytree \cite{dasgupta1999learning}. Another example is a hierarchical structure. In that case, it is natural to learn a DAG such that it satisfies the hierarchy among different groups of random variables. This problem has important applications in discovering the genetic basis of complex diseases (e.g., asthma, diabetes, atherosclerosis). 
 
%\newpage
%\section*{Acknowledgments}
%\as{We would like to acknowledge the assistance of volunteers in putting
%together this example manuscript and supplement.}

\bibliographystyle{plain}
\bibliography{ref}

%\\overrightarrow{E}liographystyle{alpha}
%\\overrightarrow{E}liography{ref}

\newpage
\section*{Appendix I}

\noindent \textbf{PROOF OF PROPOSITION \ref{Prop1: Cycle}} \\
Let $({\hat{\mathbf{\beta}}}, \hat{z})$  be an optimal solution for \eqref{eq:PNLMform} with an optimal objective value $F(\hat{\beta})$. %The $\hat{z}$ is simply used to encode all the arcs in an optimal DAG. 
Let us refer to the DAG structure corresponding to this optimal solution by $DAG(V, \hat{E}^{\rightarrow})$. Suppose that for some $(j,k)$, we have $\hat{z}_{jk} +\hat{z}_{kj} = 0$. To prove the proposition, we construct an optimal solution which satisfies $z_{jk} +z_{kj} = 1$ for all pairs of $(j,k)$ and meets the following conditions: (i) this corresponding DAG (tournament) is cycle free (ii) this tournament has the same objective value, i.e., $F(\hat{\beta})$, as an optimal DAG. 

Select a pair of nodes, say $p, q \in \mathcal{M}, p \neq q$ from $DAG(V, \hat{E})$ for which $\hat{z}_{pq} +\hat{z}_{qp} = 0$. If there is a directed path from $p$ to $q$ (respectively $q$ to $p$), then we can add the following arc $(p, q)$ (respectively, $(q,p)$). This arc does not create a cycle in the graph. 
If there is no directed path between $p$ and $q$, we can add an arc in either direction. In all cases, set $\beta$ value corresponding to the added arc to zero. We repeat this process for all pairs of nodes with  $\hat{z}_{pq} +\hat{z}_{qp} = 0$.

We can add such arcs without creating any cycle. This is because if we cannot add an arc in either direction, it implies that we should have a directed path from $p$ to $q$ and a directed path from $q$ to $p$ in graph $DAG(V, \hat{E})$ which is a contradiction because it implies a directed cycle in an optimal DAG. Note that in each step, we maintain a DAG. Hence, by induction we conclude that condition (i) is satisfied. The pair of nodes can be chosen arbitrarily. 

Since in the constructed solution we set $\beta$ for the added arcs as zero, the objective value does not change. This satisfies condition (ii), and completes the proof. \hfill  $\square$ \\ 

\noindent \textbf{PROOF OF PROPOSITION \ref{Prop2: LNProof}} \\
First we prove that \eqref{LN-con4} removes all cycles. Suppose, for contradiction, that a cycle of size $p \geq 2$ is available and represented by $(1,2, \dots, p, 1)$. This implies $z_{j+1, j} =0 \, \text{and} \, z_{j, j+1}=1, \, \forall j=\{1, \dots, p-1\}$, and $z_{p,1}=1, z_{1,p}=0$. Then, 
\begin{eqnarray*}
	&&1= z_{12}-mz_{21} \leq \psi_2 - \psi_1, \\
	&&1= z_{23}-mz_{32} \leq \psi_3 - \psi_2, \\
	&&\vdots \\
	&&1= z_{p-1,p}-mz_{p,p-1} \leq \psi_p - \psi_{1}, \\
	&&1= z_{p,1}-mz_{1,p} \leq \psi_1 - \psi_p. 
\end{eqnarray*}
\raggedbottom

\noindent We sum the above inequalities and conclude $p \leq 0$, a contradiction. 

To complete the proof, we also need to prove that any DAG is feasible for the LN formulation. To this end, we know that each DAG has a topological ordering. For all the existing arcs in the DAG, substitute $z_{jk}=1$ and assign a topological ordering number to the variables $\psi_k, k \in V$ in LN. Then, the set of constraints in \eqref{LN-con4} is always satisfied. %This shows that each DAG is indeed feasible for the LN formulation.  This completes the proof.  
\hfill  $\square$  \\

\noindent \textbf{PROOF OF PROPOSITION \ref{Prop3: LNLO}} \\
The LO formulation is in $w$-space whereas LN is in $(z, \psi)$-space. Hence, we first construct a mapping between these decision variables.

Given a feasible solution $w_{jk}$ for all $ j, k  \in V, \, j \neq k$ in the LO formulation, we define $z_{jk} = w_{jk}$,  $(j,k) \in \overrightarrow{E}$ and $\psi_j = \sum_{\ell \in V\backslash{\{j\}}} w_{\ell j}$, $j \in V$. Let $\textbf{1}({x \geq 0})$ be a function which takes value 1 if $x \geq 0$ and 0 otherwise. Given $z_{jk}$ for all $(j,k) \in \overrightarrow{E}$ and $\psi_j$ for $j \in V$ in the LN formulation, we map $w_{jk}= z_{jk}$ for all $(j,k) \in \overrightarrow{E}$ and  $w_{jk}= \textbf{1}(\psi_k - \psi_j + 1 \geq 0)$ for all $(j,k) \notin \overrightarrow{E}$. %where $S$ stands for the sign function. 
Note that $w$-space is defined for all pair of nodes whereas $z$-space is defined for the set of arcs in $\overrightarrow{E}$. In every correct mapping, we have $w_{jk}=z_{jk} \text{,} \, \, \forall  (j,k) \in \overrightarrow{E}$.

Fixing $j$ and $k$ for each $(j,k) \in \overrightarrow{E}$  and summing the left hand-side of inequalities \eqref{LO-con4} over $i \in V\setminus\{j,k\}$ we obtain 
\begin{eqnarray*}
	&& (m-2)  w_{jk} + \sum_{i \in V\backslash{\{k, j\}}} w_{ki} + \sum_{i \in V\backslash{\{j,k\}}}w_{ij} \leq m-2 \quad  \\
	%	&& = 	(m-2) \times w_{ij} + \sum_{k \in V\backslash{\{i\}}}(1- w_{kj}) -  \sum_{k \in V\backslash{\{j\}}}(1- w_{ki}) \\ && \leq m-2 \quad , \\
	&& \equiv 	(m-2)  w_{jk} + \sum_{i \in V\backslash{\{k, j\}}} w_{ki} +  \sum_{i \in V\backslash{\{j,k\}}}(1-w_{ji}) \leq m-2 \quad  \\
	&& \equiv 	m  w_{jk} -1 + \sum_{i \in V\backslash{\{k\}}} w_{ki} -  \sum_{i \in V\backslash{\{j\}}} w_{ji} \leq m-2 \quad  \\
	&& \equiv 	m  w_{jk} -1 - \sum_{i \in V\backslash{\{k\}}} w_{ik} +  \sum_{i \in V\backslash{\{j\}}} w_{ij} \leq m-2 \quad  \\
	%	&& = 	m \times w_{ij} -1 + \sum_{k \in V} w_{ki} -  \sum_{k \in V}w_{kj} \leq m-2 \quad    \\
	&& \equiv 	m  w_{jk} -m+1 \leq  \sum_{i \in V\backslash{\{k\}}} w_{ik} -  \sum_{i \in V\backslash{\{j\}}} w_{ij},   
\end{eqnarray*}
where the equivalences follow from constraints \eqref{LO-con2}. 
Given our mapping, $z_{jk} = w_{jk}$ for all $(j,k) \in \overrightarrow{E}$ and $\psi_j = \sum_{\ell \in V\backslash{\{j\}}} w_{\ell j}$ for $j \in V$, the above set of constraints can be written as
\begin{equation*}
z_{jk} - (m-1) z_{kj}  \leq  \psi_k - \psi_j \quad  \forall \, (j,k) \in \overrightarrow{E}, 
\end{equation*}
\noindent which satisfies \eqref{LN-con4} in the LN formulation. This implies $\mathcal{R}(LO) \subseteq \mathcal{R}(LN)$ for $\ell_1$-regularization. For $\ell_0$-regularization, we need to add that $g_{jk} \leq w_{jk}=z_{jk}\text{,} \, \forall (j,k) \in E ^{\rightarrow}$. This implies $\mathcal{R}(LO) \subseteq \mathcal{R}(LN)$ for $\ell_0$ regularization.

To show strict containment,  we give a point that is feasible to the LN formulation that cannot be mapped to any feasible point in the LO formulation. Consider $m=3$, $z_{13}=1, z_{31}=0, z_{32}=0.5+\epsilon, z_{23}=0.5-\epsilon, z_{12}=z_{21}=0.5$, $\psi=(1,2,2)$, for $0<\epsilon<\frac{1}{6}$, with an appropriate choice of $\beta$. It is easy to check that this is a feasible solution to the LN formulation. Because we must have $w_{ij}=z_{ij}, \forall (i,j) \in \overrightarrow{E}$, we have $w_{13}+w_{32}+w_{21}>2$. Therefore, the corresponding point is infeasible to the LO formulation and this completes the proof. 

  \hfill  $\square$ \\

\noindent \textbf{PROOF OF PROPOSITION  \ref{Prop3: LNTO}} \\
This proof is for TO formulation when the parameter $m$ on \eqref{TO-con4} is replaced with $m-1$.  

In the TO formulation, define the term $\sum_{s \in V} s o_{ks}$ as $\psi_k$ and the term $\sum_{s \in V} s o_{js}$ as $\psi_j$. Further, remove the set of constraints in  \eqref{TO-con5}, \eqref{TO-con6}, and \eqref{TO-con8}. This implies that $\mathcal{R}(TO) \subseteq \mathcal{R}(LN)$. To see strict containment, consider the  point described in the proof of Proposition  \ref{Prop3: LNLO}, which is feasible to the LN formulation. For this point, there can be no feasible assignment of the decision matrix $o$ such that $\psi_{j}=\sum_{s \in V} s o_{js}$, hence $\mathcal{R}(TO) \subset \mathcal{R}(LN)$.

\hfill  $\square$ \\ %Consider a graph with two nodes. Let $\psi_1=0.1$ and $\psi_2=1.9$. Then, solving a set of linear equations implies that \eqref{TO-con4}-\eqref{TO-con6} does not give no solution for $(O_{11}, O_{12}, O_{21}, O_{22})$. Therefore, $P(TO) \subset P(LN)$.  \\     

\noindent \textbf{PROOF OF PROPOSITION \ref{Prop4: Root}} \vspace{0.1in}\\
Let  $\bar{F}^{}(\beta_{X}^\star)$ denote the optimal objective value associated with the continuous relaxation of model $X \in \{CO, LO, LN\}$. \\
 
\noindent \textbf{Part A. $\bar{F}^{}(\beta_{LO}^\star) = \bar{F}^{}(\beta_{LN}^\star)$}. \vspace{0.1in}\\
\noindent Case 1. $\ell_1$ regularization \vspace{0.1in}\\
Suppose (${\beta_{LN}^\star}, z^\star$) is an optimal solution associated with the continuous relaxation of the LN formulation \eqref{L-obj}-\eqref{LN-con6} and  (${\beta_{LO}^\star}, w^\star$) is an optimal solution associated with continuous relaxation of the LO formulation with $\ell_1$-regularization. 

Given Proposition \ref{Prop3: LNLO}, we conclude that $\bar{F}(\beta_{LN}^{\star}) \leq \bar{F}^{}(\beta_{LO}^{\star})$. We prove that $\bar{F}^{}(\beta_{LN}^{\star}) \geq \bar{F}^{}(\beta_{LO}^{\star})$ also holds in an optimal solution. To this end, we map an optimal solution in continuous relaxation of the $\ell_1$-LN formulation to a feasible solution in continuous relaxation of the $\ell_1$-LO formulation with the same objective function. This implies $\bar{F}^{}(\beta_{LN}^{\star}) \geq \bar{F}^{}(\beta_{LO}^{\star})$. 

Given an optimal solution (${\beta_{LN}^\star}, z^\star$) to the continuous relaxation of the $\ell_1$-LN formulation, we construct a feasible solution $(\beta_{LO},w)$ to the continuous relaxation of the LO formulation as \vspace{0.1in} \\
$
w_{jk}=w_{kj}=
\begin{cases}
\frac{1}{2}, \quad z_{jk}^\star > 0, \, (i,j) \in \overrightarrow{E}, \\
0, \quad \text{otherwise},
\end{cases}
$ \vspace{0.1in} \\
and let $\beta_{LO}= \beta_{LN}^*$. 

We now show that this mapping is always valid for the $\ell_1$-LO formulation. Recall that for the $\ell_1$ regularization, we do not have the decision vector $g$ in the formulation. Three set of constraints have to be satisfied in the $\ell_1$-LO formulation.

\noindent $|\beta_{jk}| \leq Mw_{jk}, \quad \forall (j,k) \in \overrightarrow{E}$ \quad \eqref{LO-con1} \\
$w_{jk} +w_{kj}=1, \quad \forall (j,k) \in \overrightarrow{E}$\quad \eqref{LO-con2} \\ 
$w_{ij} +w_{jk} + w_{ki} \leq 2, \quad  \forall (i,j), (j,k), (k,i) \in \overrightarrow{E} \, i \neq j \neq k.$ \quad\eqref{LO-con4}  

The set of constraints \eqref{LO-con1} is trivially satisfied because we set $M \geq 2  \underset{(j,k) \in \overrightarrow{E}}{\max} \, \quad |\beta_{jk}^{\star}|$. The set of constraints \eqref{LO-con2} is trivially satisfied. The set of constraints \eqref{LO-con4}  is satisfied because the left hand side of inequality can take at most $\frac{3}{2}$ given this mapping. Therefore, $\bar{F}^{}(\beta_{LO}^\star) \leq \bar{F}^{}(\beta_{LO}) = \bar{F}^{}(\beta_{LN}^\star)$. This completes this part of the  proof.

\noindent Case 2. $\ell_0$ regularization \vspace{0.1in}\\
Suppose (${\beta_{LN}^\star}, g_{LN}^\star, z^\star$) is an optimal solution associated with a continuous relaxation of the $\ell_0$-LN formulation, and  (${\beta_{LO}^\star}, g_{LO}^\star, w^\star$) is an optimal solution associated with a continuous relaxation of the $\ell_0$-LO formulation. 

Given Proposition \ref{Prop3: LNLO}, $\bar{F}^{}(\beta_{LN}^{\star}, g_{LN}^\star) \leq \bar{F}^{}(\beta_{LO}^{\star}, g_{LO}^\star)$. We now prove that, in an optimal solution $\bar{F}^{}(\beta_{LN}^{\star}, g_{LN}^\star) \geq \bar{F}^{}(\beta_{LO}^{\star}, g_{LO}^\star)$ also holds. 

Given an optimal solution (${\beta_{LN}^\star}, g_{LN}^\star, z^\star$) for the continuous relaxation of the $\ell_0$-LN formulation, we construct a feasible solution $(\beta_{LO}, g_{LO},w)$ for the continuous relaxation of the $\ell_0$-LO formulation as \\
$
w_{jk}=w_{kj}=
\begin{cases}
\frac{1}{2}, \quad z_{jk}^\star > 0, \, (j,k) \in \overrightarrow{E}, \\
0, \quad otherwise,
\end{cases}
$ \vspace{0.1in} \\
and let $\beta_{LO}= \beta_{LN}^*$ and $g_{LO}= g_{LN}^\star$. 

We now show that this mapping is always valid for the LO formulation. Four sets of constraints have to be satisfied in the LO formulation. \\
 
\noindent $|\beta_{jk}| \leq Mg_{jk}, \quad \forall (j,k) \in \overrightarrow{E}$, \eqref{LO-con1} \\
$g_{jk} \leq w_{kj}, \quad \forall (j,k) \in \overrightarrow{E}$ \eqref{LO-con3} \\ 
$w_{jk} +w_{kj}=1, \quad \forall (j,k) \in \overrightarrow{E}$\quad \eqref{LO-con2} \\ 
$w_{ij} +w_{jk} + w_{ki} \leq 2, \quad  \forall i, j, k \in V, i \neq j \neq k,$ \quad \eqref{LO-con4} \\

The set of constraints in \eqref{LO-con2} is satisfied similar to $\ell_1$ case. The proof that constraints \eqref{LO-con1}, \eqref{LO-con3}, and \eqref{LO-con4} are also met is more involved. 
If the $\ell_0$-LN formulation attains a solution for which $g_{ij}^{\star}=g_{jk}^{\star}=g_{ki}^{\star}=1$ (we dropped subscript LN), then our mapping leads to an infeasible solution to the $\ell_0$-LO formulation, because it forces $w_{ij} + w_{jk} + w_{ki} \geq 2$ for $\ell_0$-LO. Next we show that this will not be the case and that our mapping is valid. 
To this end, we show that in an optimal solution for the continuous relaxation of $\ell_0$-LN, we always have $g^{\star}_{jk} \leq \frac{1}{2}\text{,}  \, \forall(j,k) \in \overrightarrow{E}$. Note that in the LN formulation, we have $|\beta_{jk}| \leq M g_{jk}\text{,} \, \forall (j,k) \in E ^{\rightarrow}$ and $g_{jk} \leq z_{jk}$ (we dropped the subscript LN). Suppose that $g^{\star}_{jk} \geq \frac{1}{2}$. In this case, the objective function forces $g^{\star}_{jk}$ to be at most $\frac{1}{2}$. Note that the objective function can reduce $g^{\star}_{jk}$ up to $\frac{1}{2}$ and decreases the regularization term without any increase on the loss function. This is because $\beta_{jk} \leq M g_{jk}\text{,} \, \forall (j,k) \in E ^{\rightarrow}$ can be replaced by $\beta_{jk} \leq M \frac{1}{2}\text{,} \, \forall (j,k) \in E ^{\rightarrow}$ without any restriction on $\beta$ because  $M \geq 2 \underset{(j,k) \in \overrightarrow{E}}{\max} \, |\beta^{\star}_{jk}|$. Therefore, our mapping is valid and implies that $\bar{F}^{}(\beta_{LO}^\star, g^{\star}_{LO}) \leq \bar{F}^{}(\beta_{LO}, g_{LO}) = \bar{F}^{}(\beta_{LN}^\star, g^{\star}_{LN})$.

\noindent \textbf{Part B. $\bar{F}^{}(\beta_{TO}^\star) = \bar{F}^{}(\beta_{LN}^\star)$.} \vspace{0.1in}\\
\noindent  Case 1. $\ell_1$ regularization \vspace{0.1in}\\
Given Proposition \ref{Prop3: LNTO}, we conclude that $F^{}(\beta_{LN}^{\star}) \leq F^{}(\beta_{TO}^{\star})$. We now prove that $\bar{F}^{}(\beta_{LN}^{\star}) \geq \bar{F}^{}(\beta_{TO}^{\star})$ also holds in an optimal solution. We map an optimal solution in continuous relaxation of the $\ell_1$-LN formulation to a feasible solution in continuous relaxation of the $\ell_1$-TO formulation with the same objective value. This implies $\bar{F}^{}(\beta_{LN}^{\star}) \geq \bar{F}^{}(\beta_{TO}^{\star})$. 

Next we construct a feasible solution $(\beta_{TO}, z_{TO}, o)$ to the TO formulation. Given an optimal solution (${\beta_{LN}^\star}, z_{LN}^\star$, $\psi^\star$) for a continuous relaxation of the LN formulation, rank the $\psi^\star_j$ in non-descending order. Ties between $\psi$ values can be broken arbitrarily in this   ranking. Then, for each variable $j \in \{1, \dots, m\}$, $o_{jr}=1$ where $r$ denotes the rank of $\psi_j$ in a non-descending order (the first element is ranked 0). This mapping satisfies the two assignment constraints \eqref{TO-con5}-\eqref{TO-con6}. Let $\beta_{TO}= \beta_{LN}^*$, $z_{TO}= z_{LN}^*$. This gives a feasible solution for the TO formulation. Thus, $\bar{F}^{}(\beta_{LN}^{\star}) \geq \bar{F}^{}(\beta_{TO}^{\star}). $

\noindent  Case 2. $\ell_0$ regularization \vspace{0.1in}\\
Define $g^{\star}_{LN}=g_{{TO}}$. The rest of the proof is similar to the previous proofs.  \\ 

\noindent \textbf{Part C. $\bar{F}^{}(\beta_{CP}^\star) = \bar{F}^{}(\beta_{LN}^\star)$}. \vspace{0.1in}\\
To prove this, we first prove the following Lemma. \vspace{0.1in} \\
\noindent \textbf{Lemma}. \textit{The LO formulation is at least as strong as the CP formulation, that is $\mathcal{R}(LO) \subseteq \mathcal{R}(CP)$.} \\
  
\noindent \textbf{Proof.} Consider \eqref{LO-con2}-\eqref{LO-con4} in the linear ordering formulation given by 
\begin{eqnarray*}
	&& w_{jk} +w_{kj} = 1 \quad \forall \, \, \,(j,k) \in \overrightarrow{E}, \\
	&& w_{ij} +w_{jk} + w_{ki} \leq 2 \quad  \forall (i,j), (j,k), (k,i) \in \overrightarrow{E}.
\end{eqnarray*}

Consider the set of constraints in CP given by \eqref{CP-con2} as 

\begin{eqnarray*}
\sum_{(j,k) \in \, \mathcal{C}_A} g_{jk} \leq |\mathcal{C}_A|-1 \quad \forall \mathcal{C}_A \in \mathcal{C}, \\
\end{eqnarray*}

Consider the same mapping as in Proposition \ref{Prop3: LNLO}. Select an arbitrary constraint from \eqref{CP-con2}. Without loss of generality, consider $g_{1,2} + g_{2,3} + \dots + g_{p-1,p} + g_{p,1} \leq p-1$. We arrange the terms in \eqref{LO-con2}-\eqref{LO-con4} as 
\begin{eqnarray*}
	&& w_{1,2} +w_{2,3} + w_{3,1} \leq 2 \\
	&& w_{1,3} +w_{3,4} + w_{4,1} \leq 2 \\
	&& w_{1,4} +w_{4,5} + w_{5,1} \leq 2 \\
	&& \vdots \\
	%&& w_{1,p-3} +w_{p-3,p-2} + w_{p-2,1} \leq 2 \\
	&& w_{1, p-2} +w_{p-2, p-1} + w_{p-1,1} \leq 2 \\
	&& w_{1, p-1} +w_{p-1, p} + w_{p,1} \leq 2  
\end{eqnarray*}

Summing the above set of inequalities and substituting  $w_{ij}=1-w_{ji}$, when appropriate, gives $w_{1,2} + w_{2,3} + w_{3,4} + \dots + w_{p,1} \leq p-1$. Thus, we conclude that $\mathcal{R}(LO) \subseteq \mathcal{R}(CP)$. \hfill $\square$ 

Given this lemma and Proposition \ref{Prop3: LNLO}, we conclude that $\bar{F}^{}(\beta_{LN}^{\star}) \geq \bar{F}^{}(\beta_{TO}^{\star})$. \hfill $\square$ \\

%\newpage 
%Proof of Lemma. Recall \\

\noindent \textbf{PROOF OF PROPOSITION \ref{Prop5: BB}} 

This is a generalization of Proposition \ref{Prop4: Root}. Suppose we have branched on variable $w_{jk}$ in the LO formulation or correspondingly on variable $z_{jk}$ in the LN formulation. If $w_{jk}=0$, then $\beta_{jk}=0$. In this case, it is as if we now need to solve the original model with $(j,k) \notin \overrightarrow{E}$. Thus, Proposition \ref{Prop4: Root} (Part A) implies that both models attain the same continuous relaxation. On the other hand, if $w_{jk}=1$ in LO (or correspondingly $z_{jk}=1$ in LN), we define our mapping as \\

$
w_{jk}=w_{kj}=
\begin{cases}
	\frac{1}{2}, \quad z_{jk}^\star > 0, \, (j,k) \in \overrightarrow{E}, \\
	1, \quad (j,k) \in \mathcal{B} \\
	0, \quad \text{otherwise},
\end{cases}
$

\noindent where $\mathcal{B}$ is the set of $(j,k)$ for which $z_{jk}=1$. The rest of the proof follows from Proposition \ref{Prop4: Root} (Part A). 

The proofs for the TO and CP formulations are almost identical with Proposition \ref{Prop4: Root}. Suppose we have branched on variable $z_{jk}$ on any of the models (CP, LO, TO). If $z_{jk}=0$, then $\beta_{jk}=0$. In this case, it is as if we now need to solve the original model with $(j,k) \notin \overrightarrow{E}$. If $z_{jk}=1$, one defines the mapping in Proposition \ref{Prop4: Root}. The proof follows from Proposition \ref{Prop4: Root} parts B and C, respectively. \hfill $\square$

\begin{figure*}[]
\noindent \textbf{Appendix II} \\
	\begin{subfigure}[t]{0.49\textwidth}
		\centering
		\includegraphics[scale=0.22]{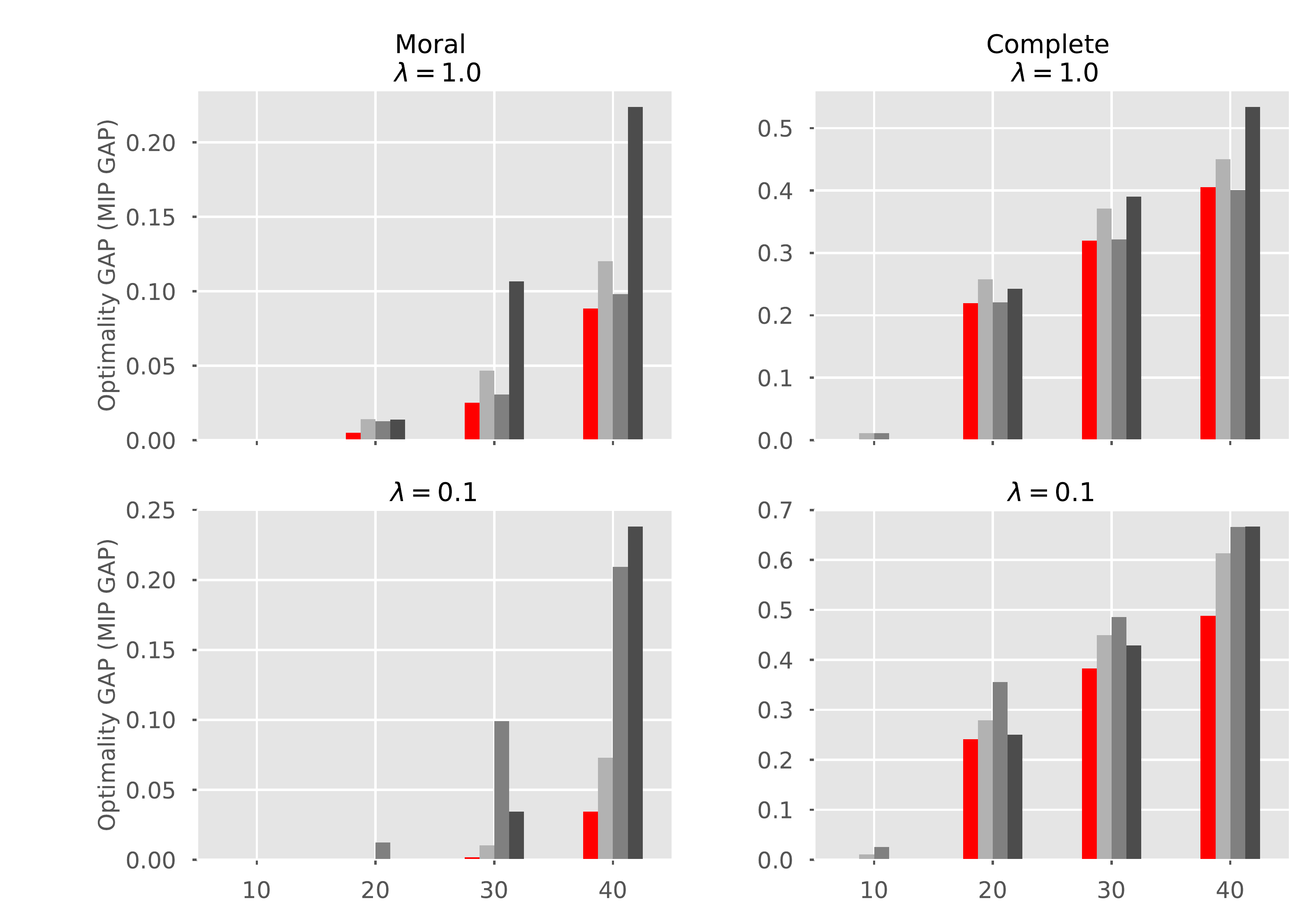}
		\caption{Optimality GAPs for MIQPs}
	\end{subfigure}%
	~ 
	\begin{subfigure}[t]{0.49\textwidth}
		\centering
		\includegraphics[scale=0.22]{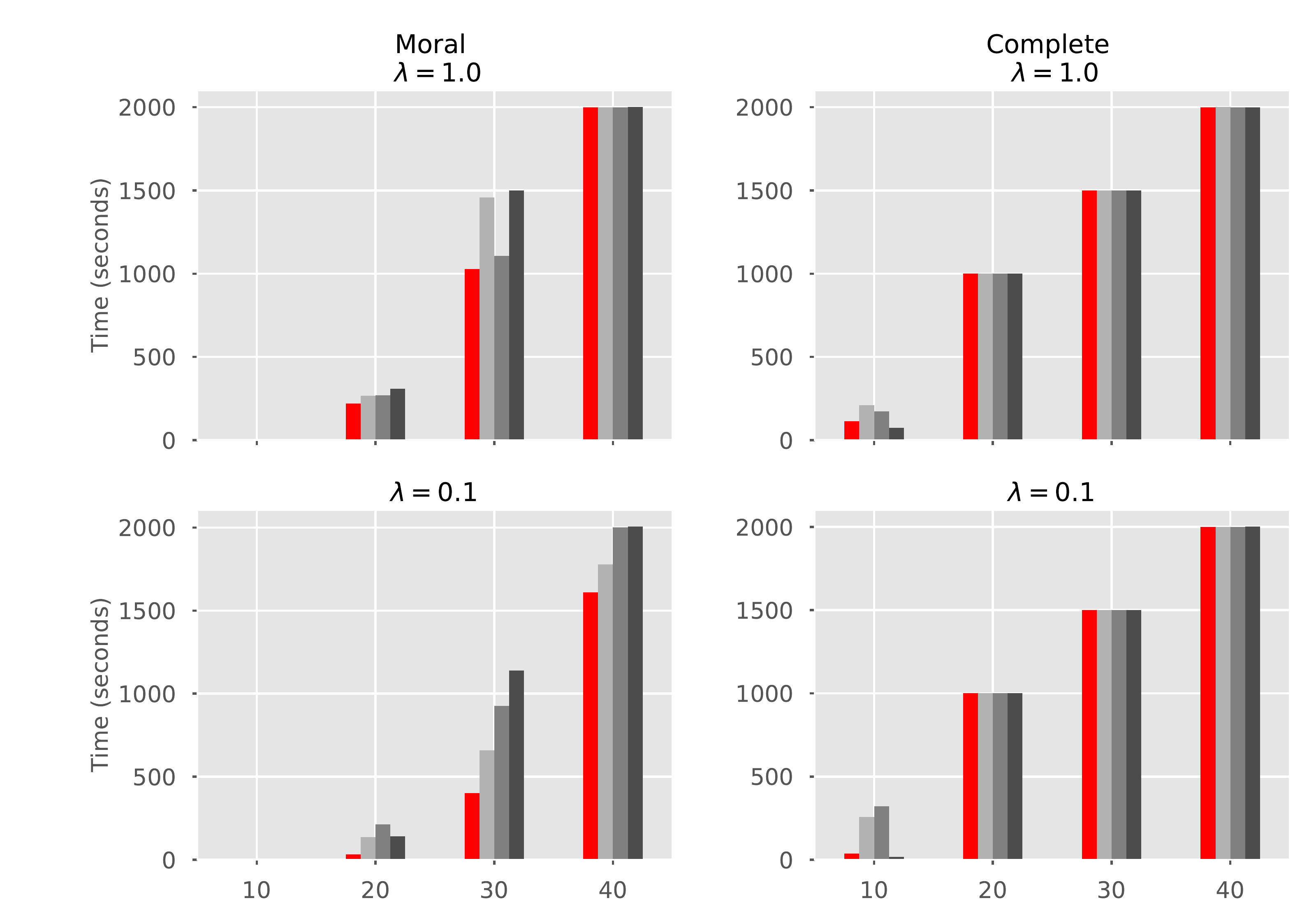}
		\caption{Time (in seconds) for MIQPs}
	\end{subfigure}
	~
	%\hspace*{-1in}
	\begin{subfigure}[t]{0.49\textwidth}
		\centering
		\includegraphics[scale=0.22]{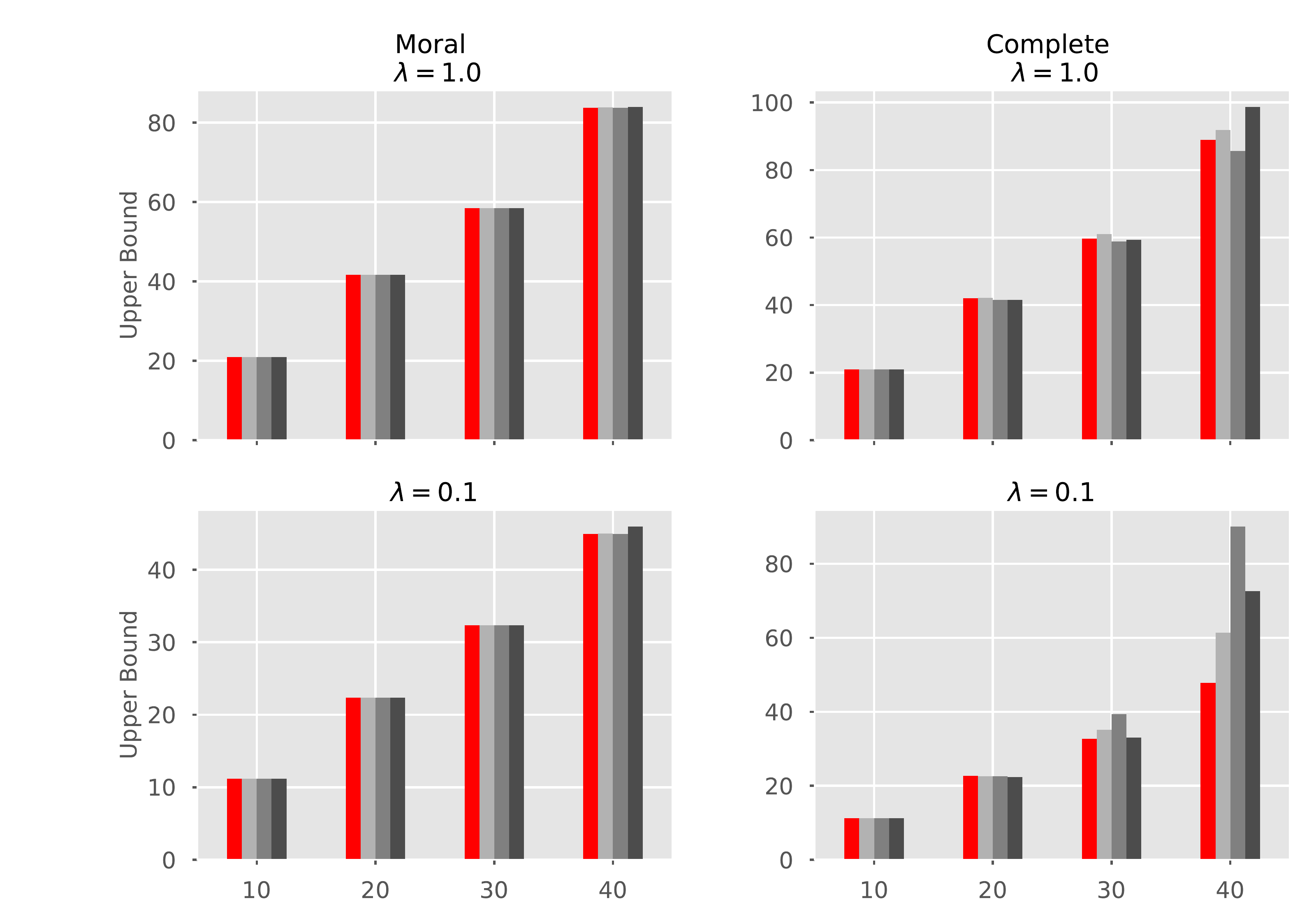}
		\caption{Best upper bounds for MIQPs}
	\end{subfigure}
	~ 
	\begin{subfigure}[t]{0.49\textwidth}
		\centering
		\includegraphics[scale=0.22]{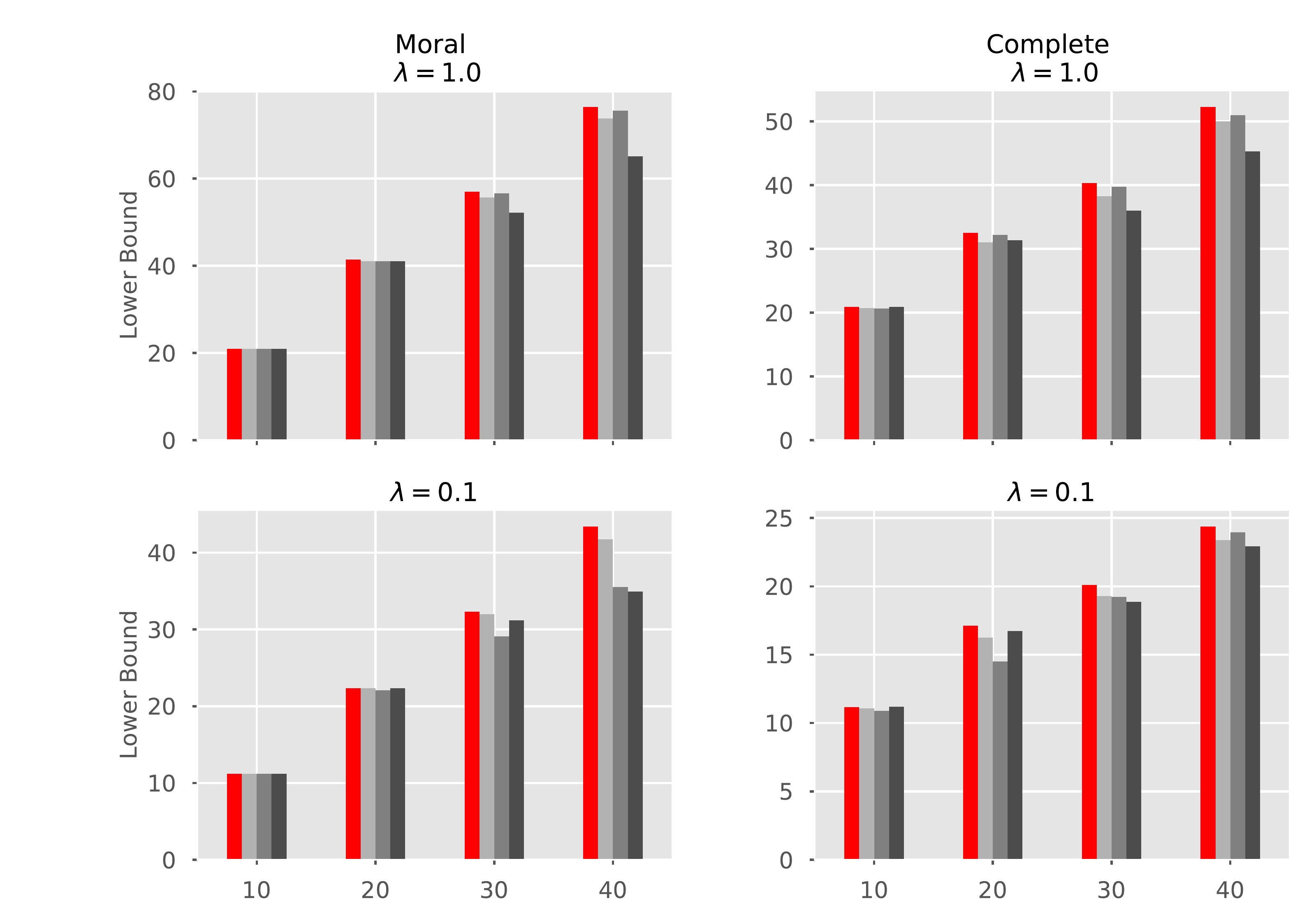}
		\caption{Best lower bounds for MIQPs}
	\end{subfigure}
	~
%	\hspace*{-1in}
	\begin{subfigure}[t]{0.49\textwidth}
		\centering
		\includegraphics[scale=0.22]{L0100Time_seconds_}
		\caption{Time (in seconds) for continuous root relaxation}
	\end{subfigure}
	~ 
%	\hspace{0.2in}
	\begin{subfigure}[t]{0.49\textwidth}
		\centering
		\includegraphics[scale=0.22]{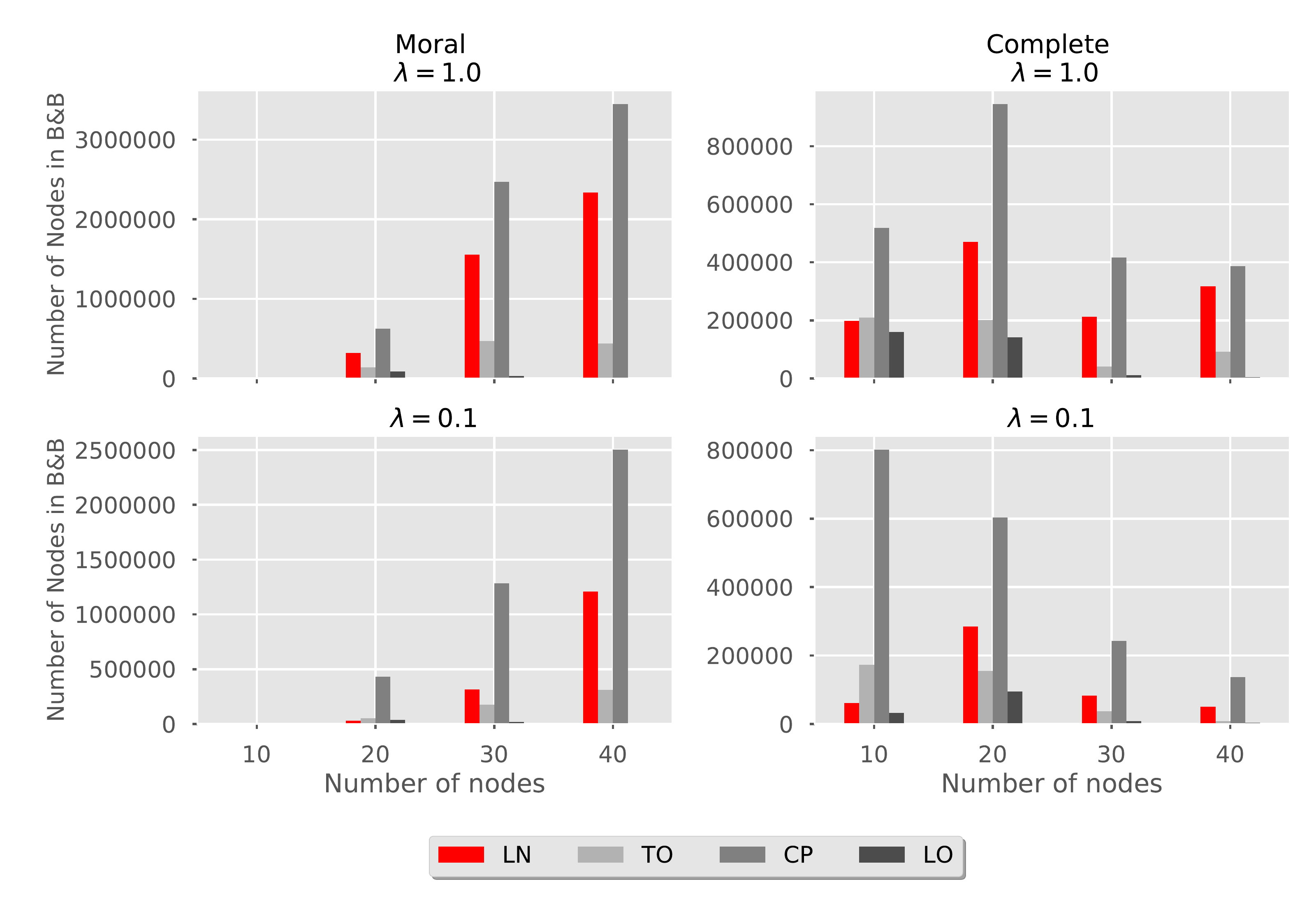}
		\caption{Number of explored nodes in B\&B tree}
	\end{subfigure}
	
	\caption{Optimization-based measures for MIQPs for $\ell_0$ model with number of samples $n=100$.}
	\label{Table0: IP_100}
\end{figure*}

\begin{figure*}[]
%	\hspace*{-1in}
%	\begin{subfigure}[t]{0.6\textwidth}
%		\centering
%		\includegraphics[scale=0.25]{Fig/L0_1000/StructuralHammingDistance}
%		\caption{Structural Hamming Distance for MIQPs}
%	\end{subfigure}%
%	~ 
%	\begin{subfigure}[t]{0.5\textwidth}
%		\centering
%		\includegraphics[scale=0.25]{Fig/L0_100/StructuralHammingDistance}
%		\caption{Structural Hamming Distance for MIQPs}
%	\end{subfigure}
%	~
%	\hspace*{-1in}
	\begin{subfigure}[t]{0.49\textwidth}
		\centering
		\includegraphics[scale=0.22]{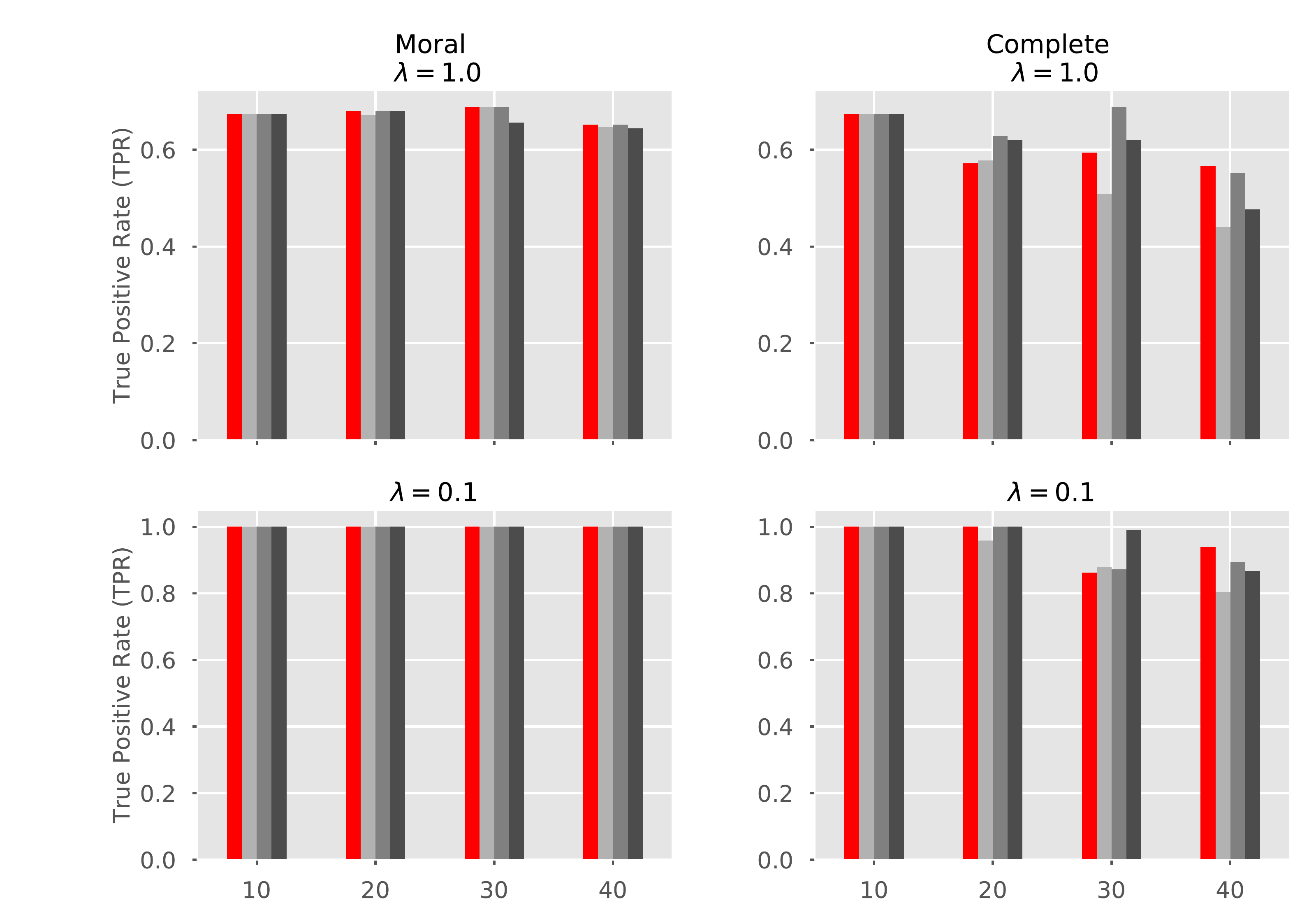}
		\caption{True Positive Rate (TPR) for MIQPs}
	\end{subfigure}
	~ 
	\begin{subfigure}[t]{0.49\textwidth}
		\centering
		\includegraphics[scale=0.22]{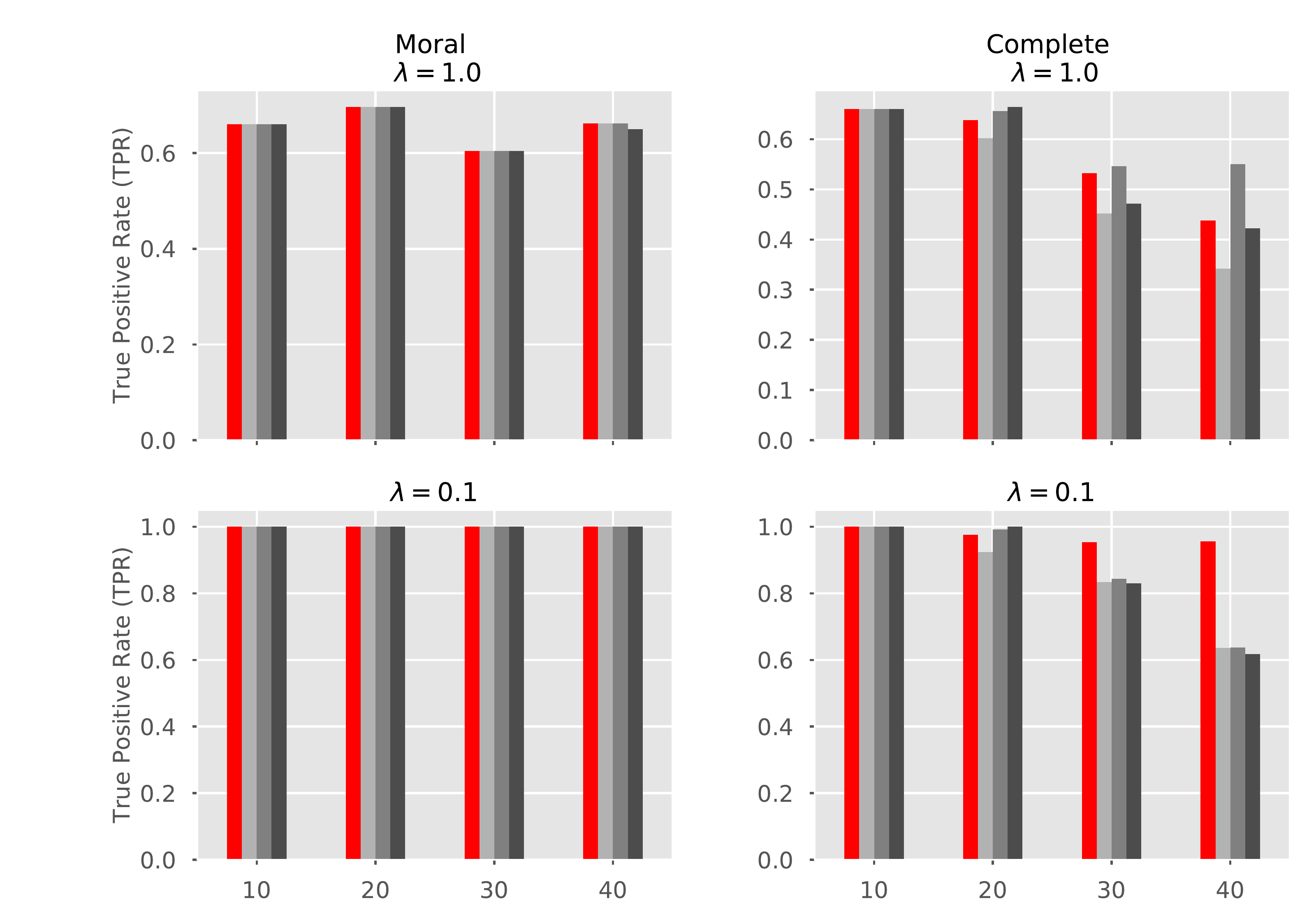}
		\caption{True Positive Rate(TPR) for MIQPs}
	\end{subfigure}
	~
%	\hspace*{-1in}
	\begin{subfigure}[t]{0.49\textwidth}
		\centering
		\includegraphics[scale=0.22]{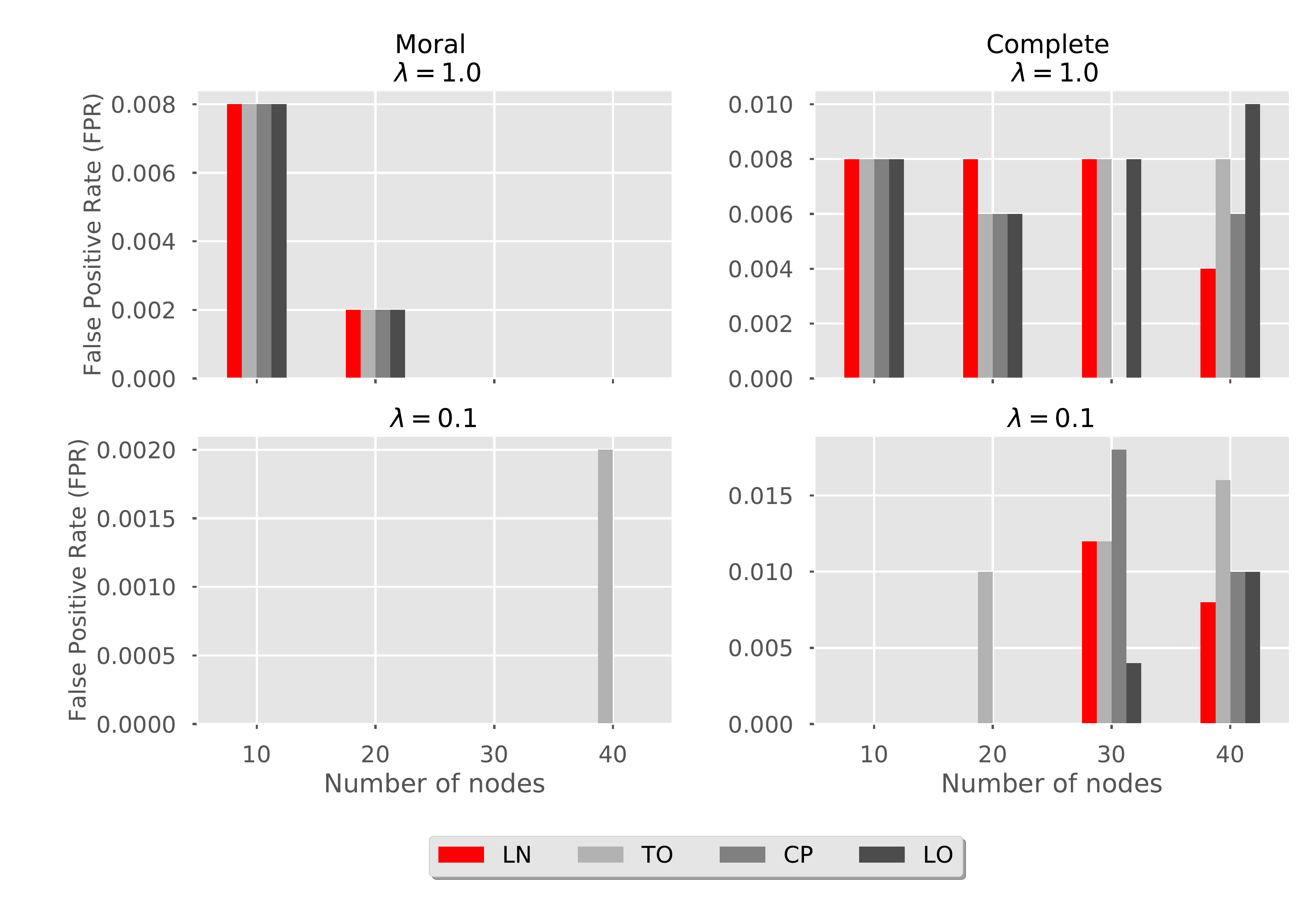}
		\caption{False Positive Rate(FPR) for MIQPs}
	\end{subfigure}
	~ 
%	\hspace{0.2in}
	\begin{subfigure}[t]{0.49\textwidth}
		\centering
		\includegraphics[scale=0.22]{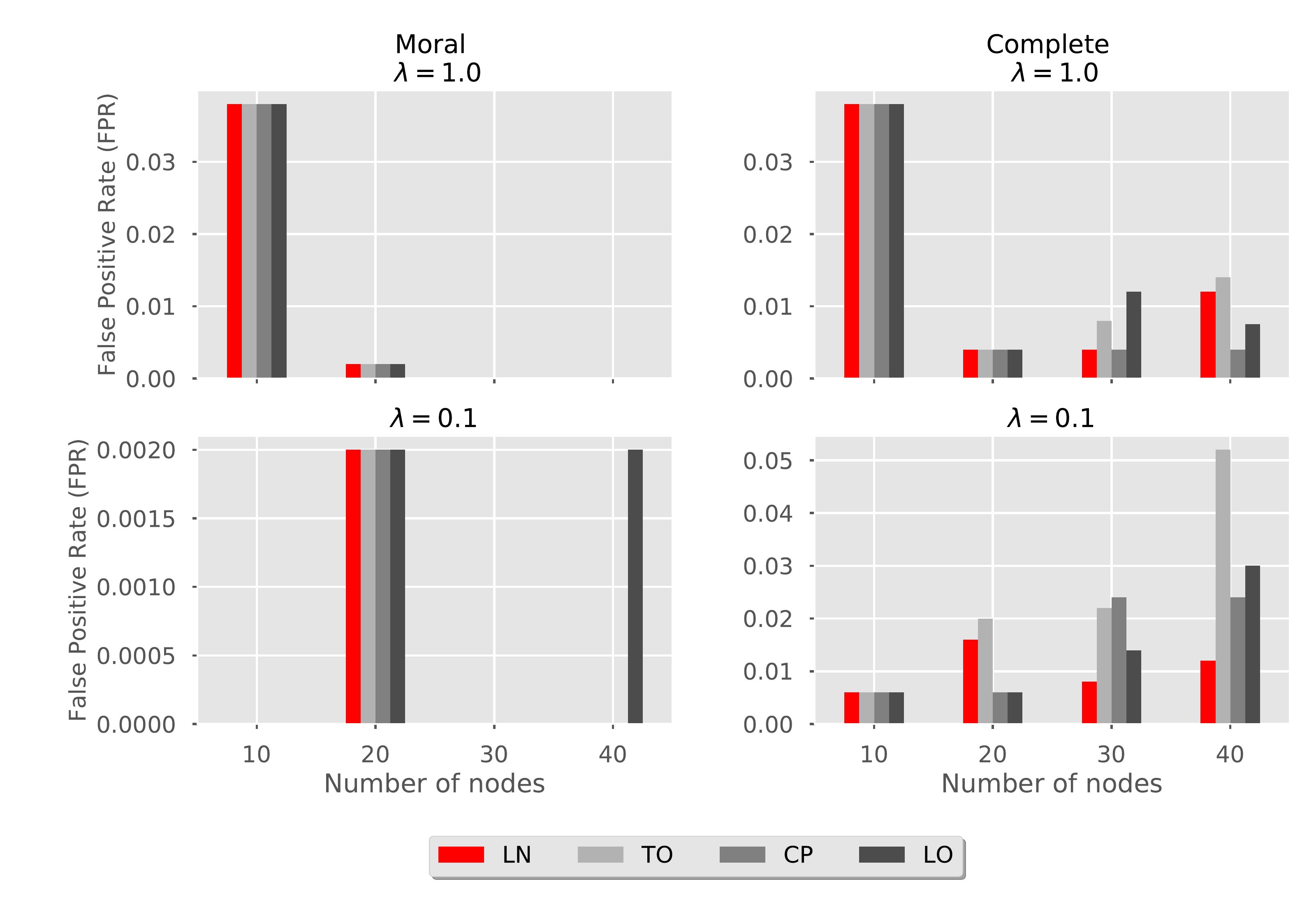}
		\caption{False Positive Rate(FPR) for MIQPs}
	\end{subfigure}
	\caption{Graph metrics for the MIQPs for $\ell_0$ regularization. Plots a, c (left) show graph metrics with the number of samples $n=1000$ and plots b, d (right) show the graph metrics with the number of samples $n=100$.}
	\label{Figure: Graph_1000}
\end{figure*}

\begin{figure*}[]
%	\hspace*{-1in}
	\begin{subfigure}[t]{0.49\textwidth}
		\centering
		\includegraphics[scale=0.22]{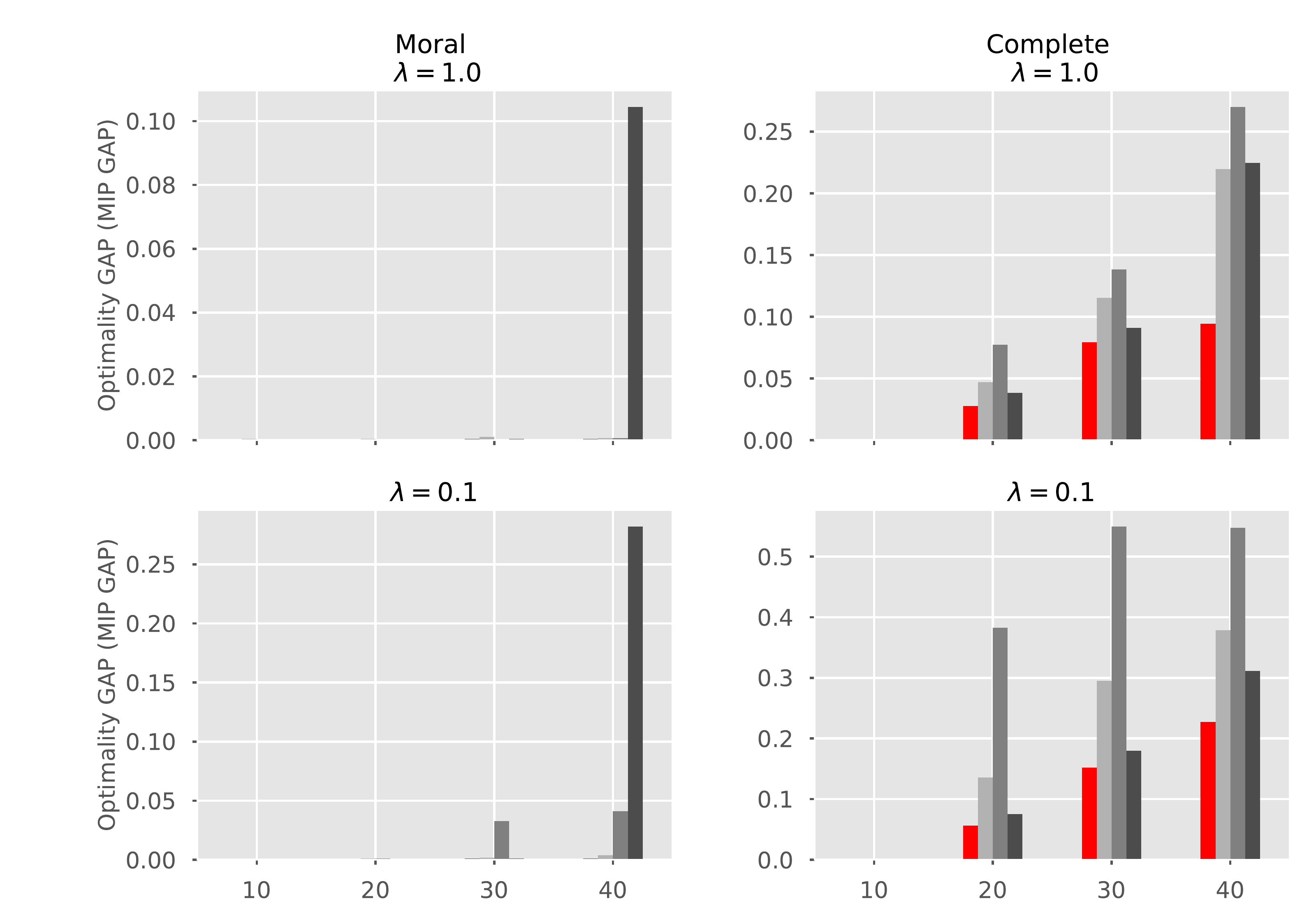}
		\caption{Optimality GAPs for MIQPs}
	\end{subfigure}%
	~ 
	\begin{subfigure}[t]{0.49\textwidth}
		\centering
		\includegraphics[scale=0.22]{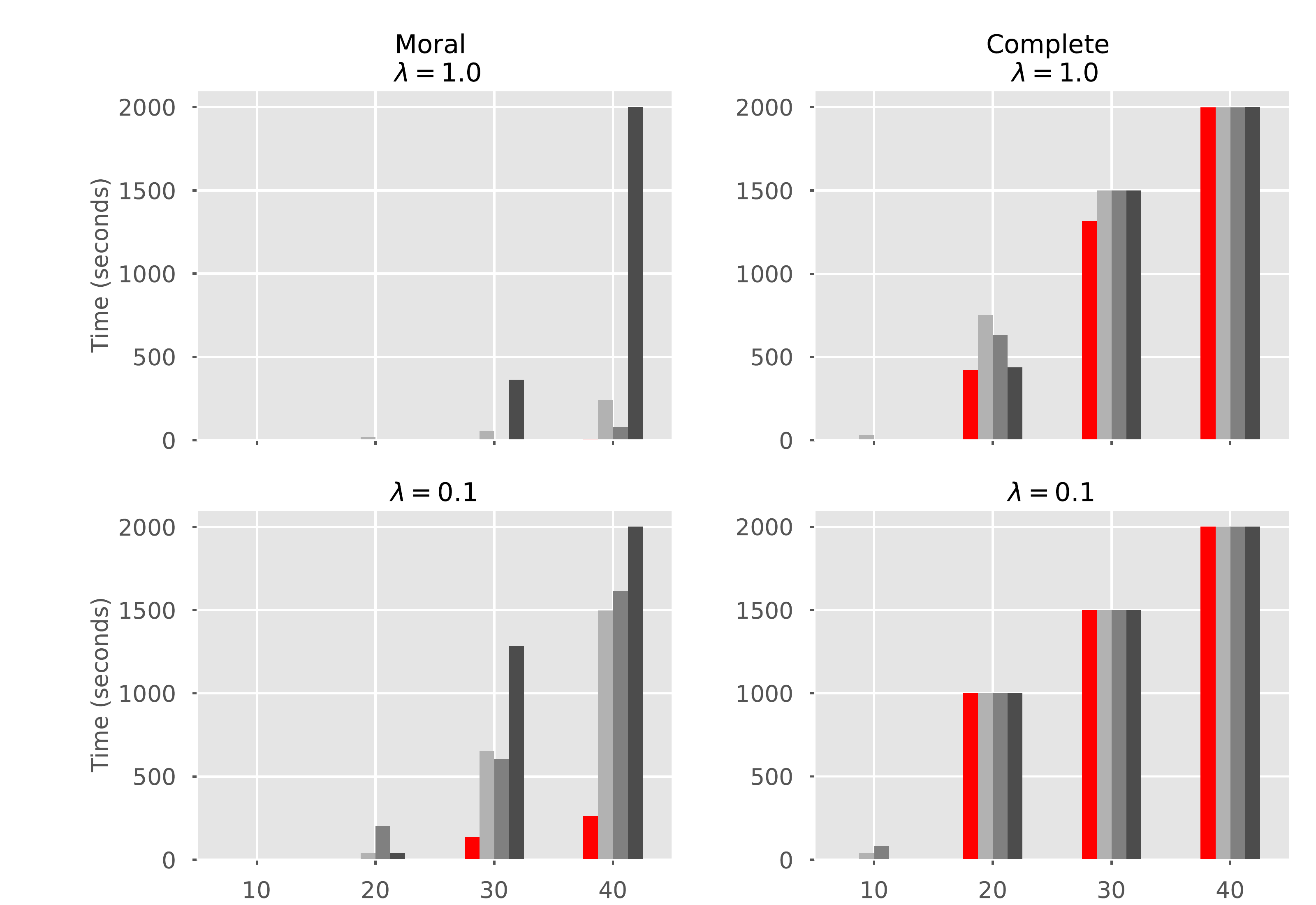}
		\caption{Time (in seconds) for MIQPs}
	\end{subfigure}
	~
%	\hspace*{-1in}
	\begin{subfigure}[t]{0.49\textwidth}
		\centering
		\includegraphics[scale=0.22]{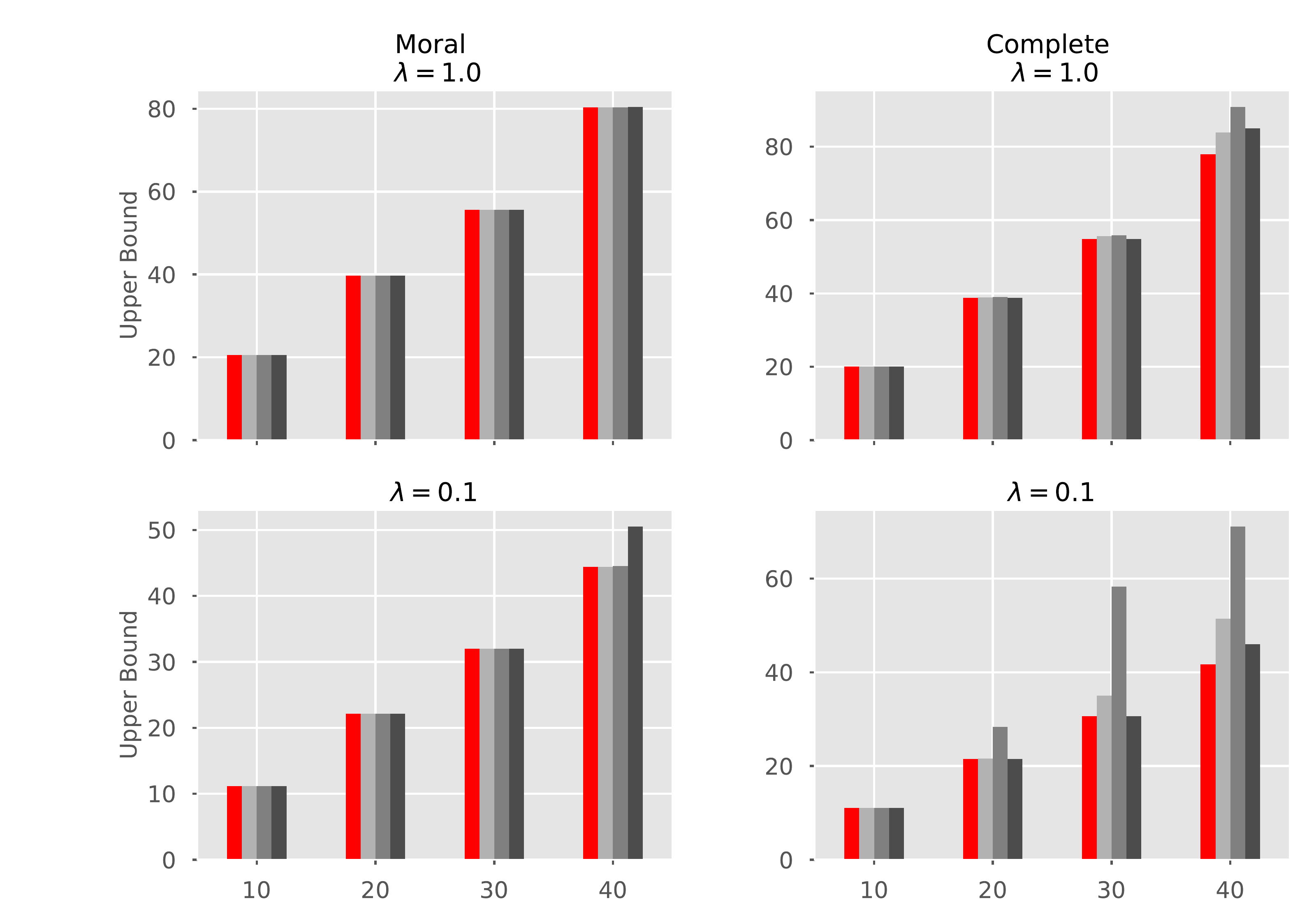}
		\caption{Best obtained upper bounds for MIQPs}
	\end{subfigure}
	~ 
	\begin{subfigure}[t]{0.49\textwidth}
		\centering
		\includegraphics[scale=0.22]{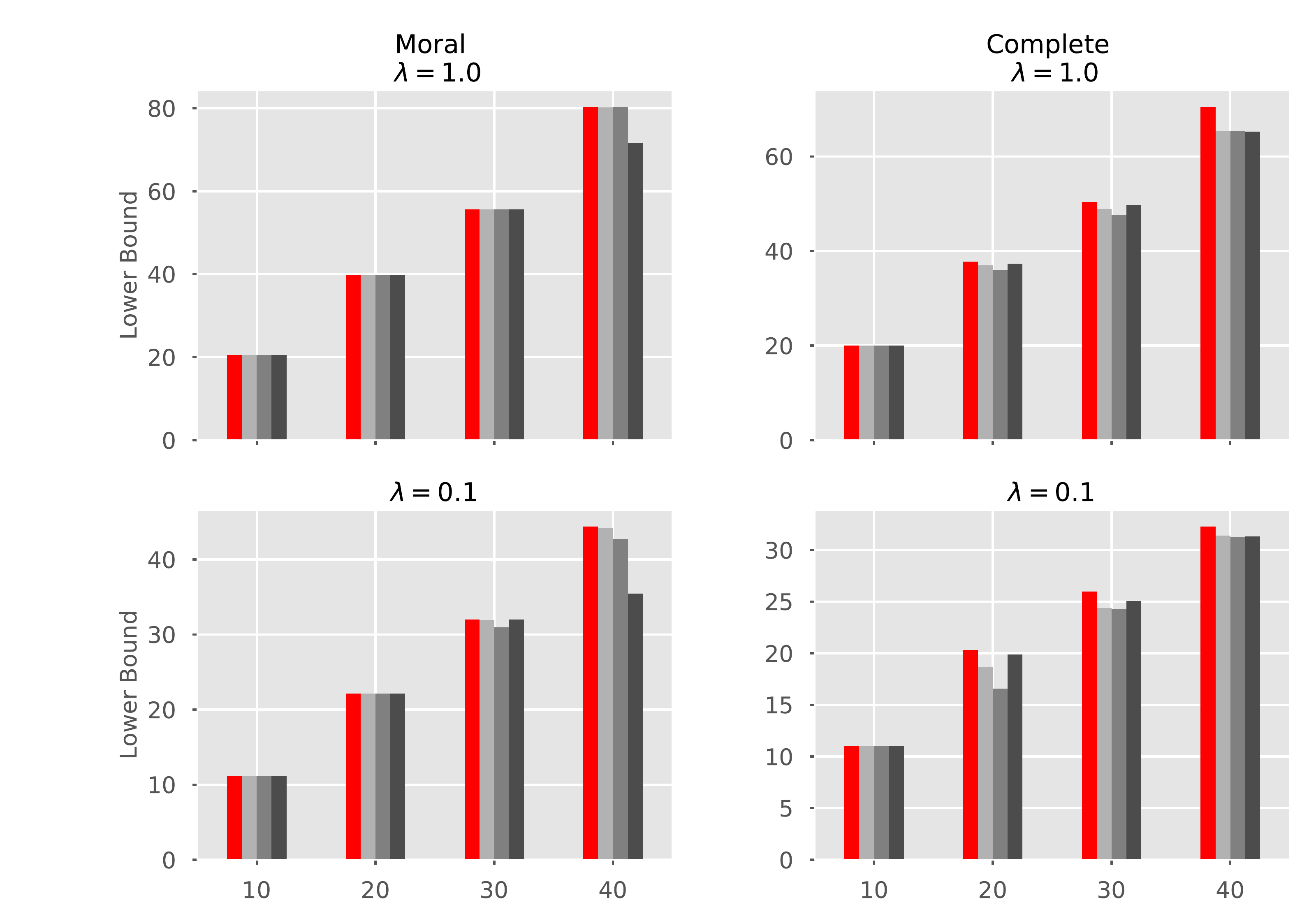}
		\caption{Best obtained lower bounds for MIQPs}
	\end{subfigure}
	~
%	\hspace*{-1in}
	\begin{subfigure}[t]{0.49\textwidth}
		\centering
		\includegraphics[scale=0.22]{L1100Time_seconds_}
		\caption{Time (in seconds) for continuous root relaxation}
	\end{subfigure}
	~ 
%	\hspace{0.2in}
	\begin{subfigure}[t]{0.49\textwidth}
		\centering
		\includegraphics[scale=0.22]{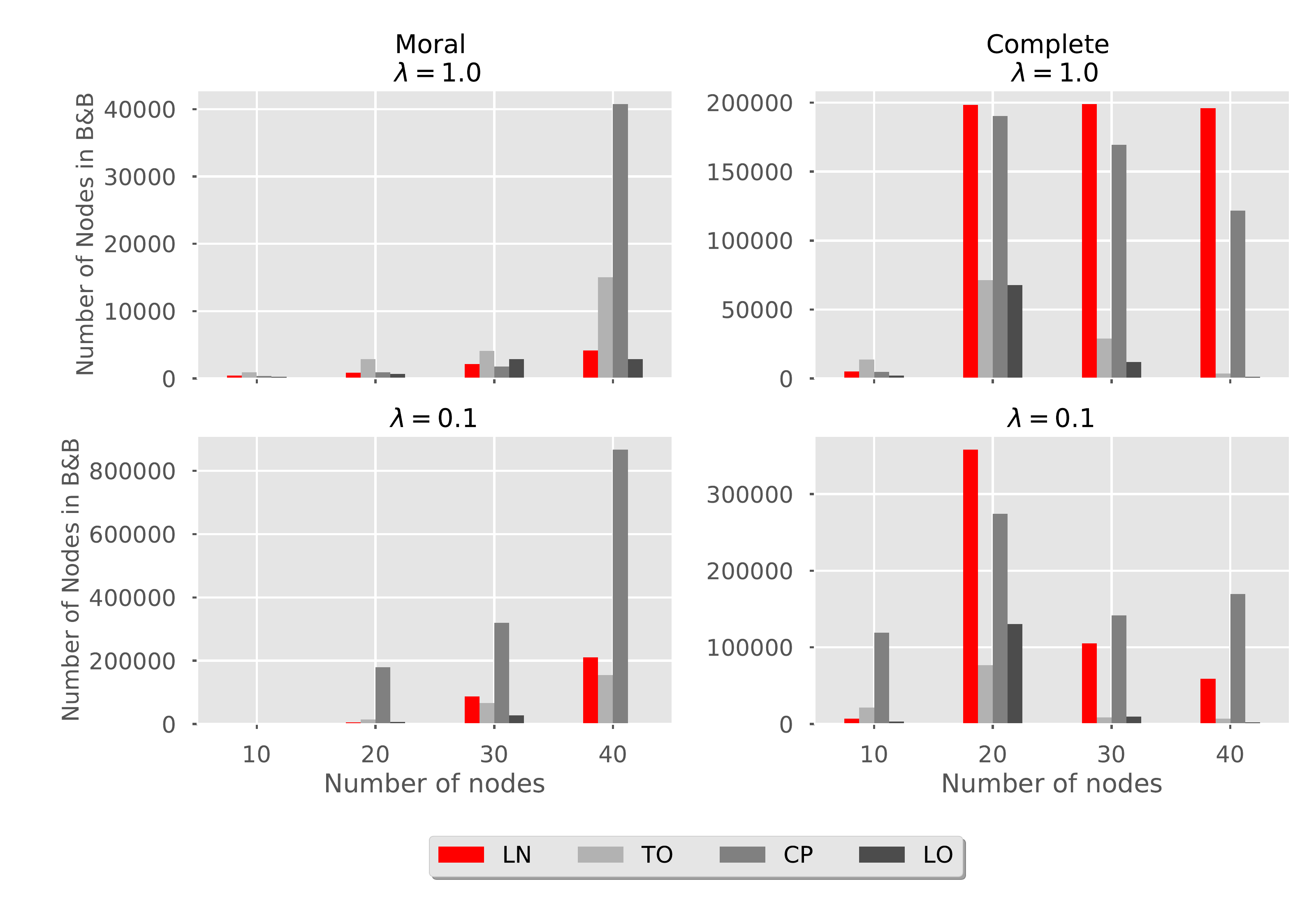}
		\caption{Number of explored nodes in B\&B tree}
	\end{subfigure}
	
	\caption{Optimization-based measures for MIQPs for $\ell_1$ model with the number of samples $n=100$.}
	\label{Table1: IP_100}
\end{figure*}

\begin{figure*}[]
	%	\hspace*{-1in}
	%\begin{subfigure}[t]{0.6\textwidth}
	%	\centering
	%		\includegraphics[scale=0.25]{Fig/L1_1000/StructuralHammingDistance}
	%	\caption{Structural Hamming Distance for MIQPs}
	%\end{subfigure}%
	%	~ 
	%\begin{subfigure}[t]{0.5\textwidth}
	%		\centering
	%		\includegraphics[scale=0.25]{Fig/L1_100/StructuralHammingDistance}
	%		\caption{Structural Hamming Distance for MIQPs}
	%	\end{subfigure}
	%	~
%	\hspace*{-1in}
	\begin{subfigure}[t]{0.49\textwidth}
		\centering
		\includegraphics[scale=0.22]{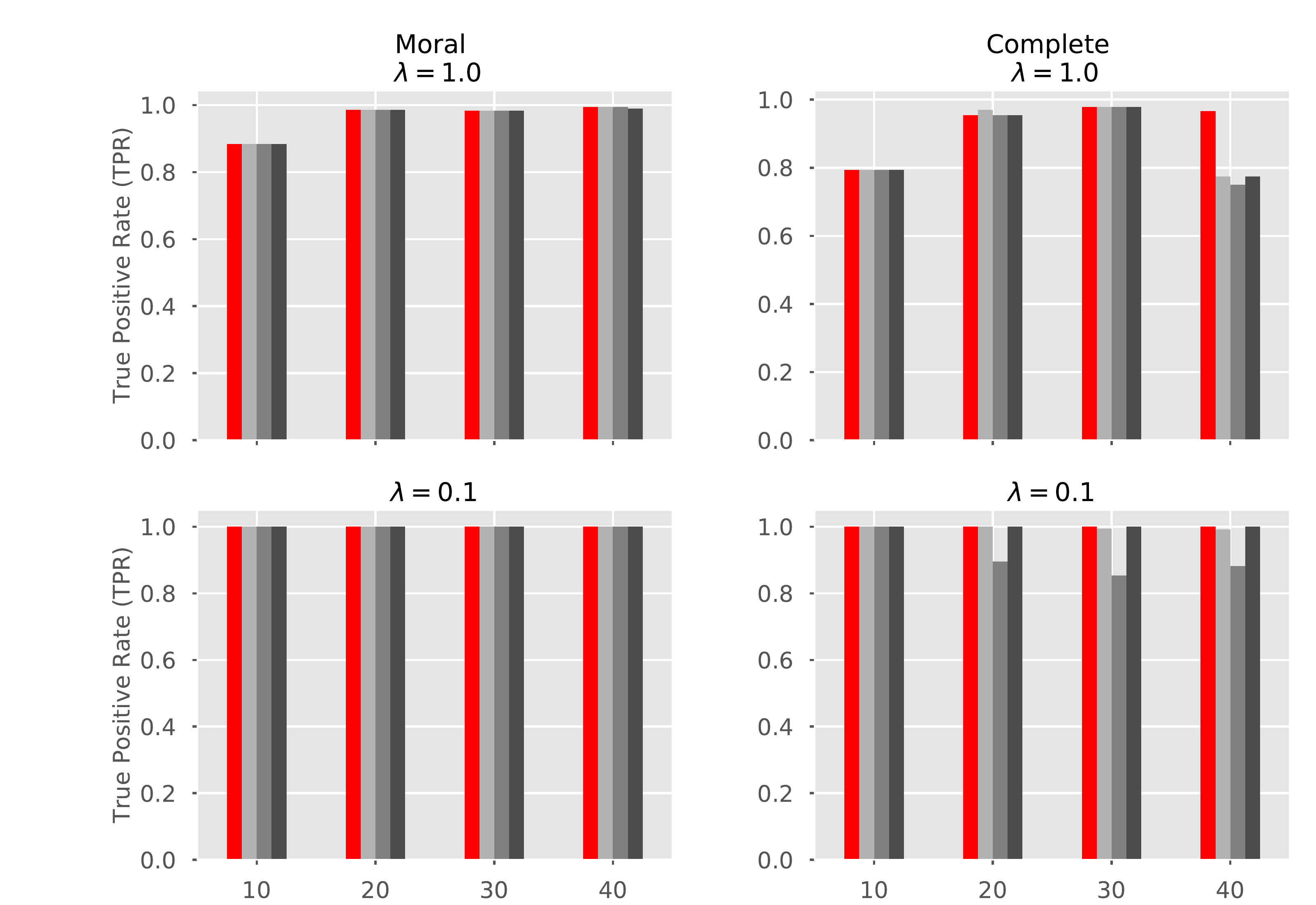}
		\caption{True Positive Rate (TPR) for MIQPs}
	\end{subfigure}
	~ 
	\begin{subfigure}[t]{0.49\textwidth}
		\centering
		\includegraphics[scale=0.22]{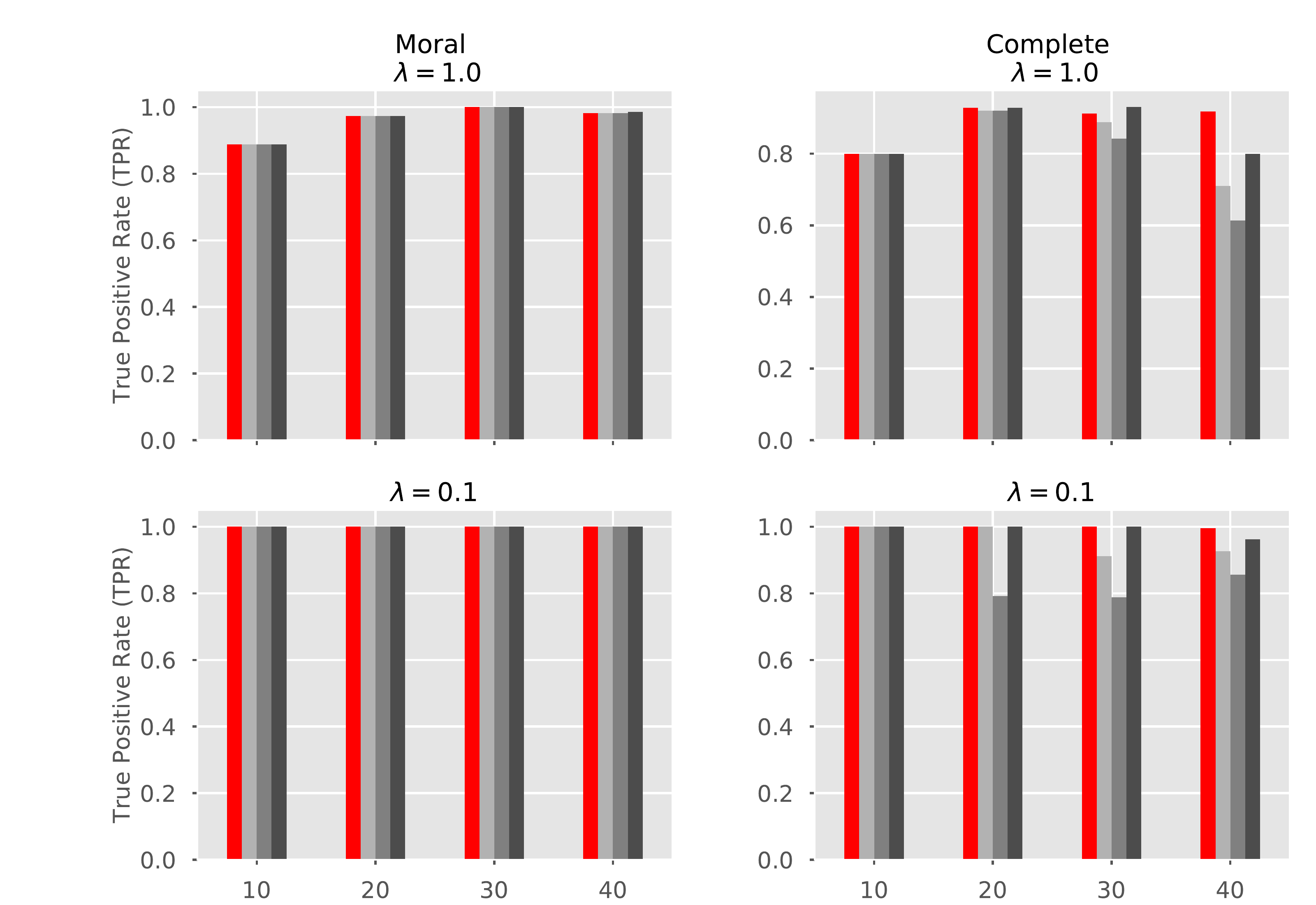}
		\caption{True Positive Rate (TPR) for MIQPs}
	\end{subfigure}
	~
%	\hspace*{-1in}
	\begin{subfigure}[t]{0.49\textwidth}
		\centering
		\includegraphics[scale=0.22]{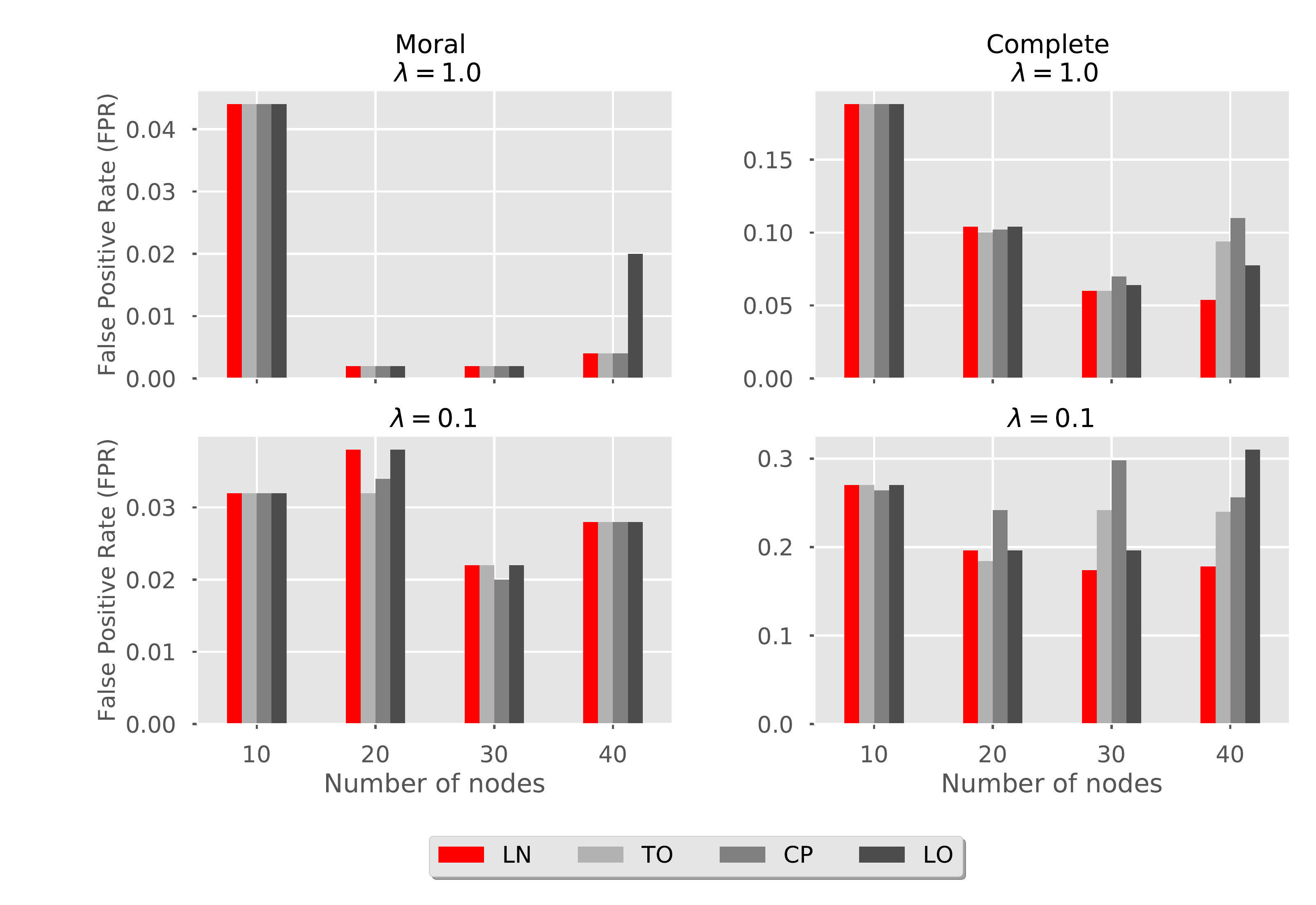}
		\caption{False Positive Rate (FPR) for MIQPs}
	\end{subfigure}
	~ 
%	\hspace{0.2in}
	\begin{subfigure}[t]{0.49\textwidth}
		\centering
		\includegraphics[scale=0.22]{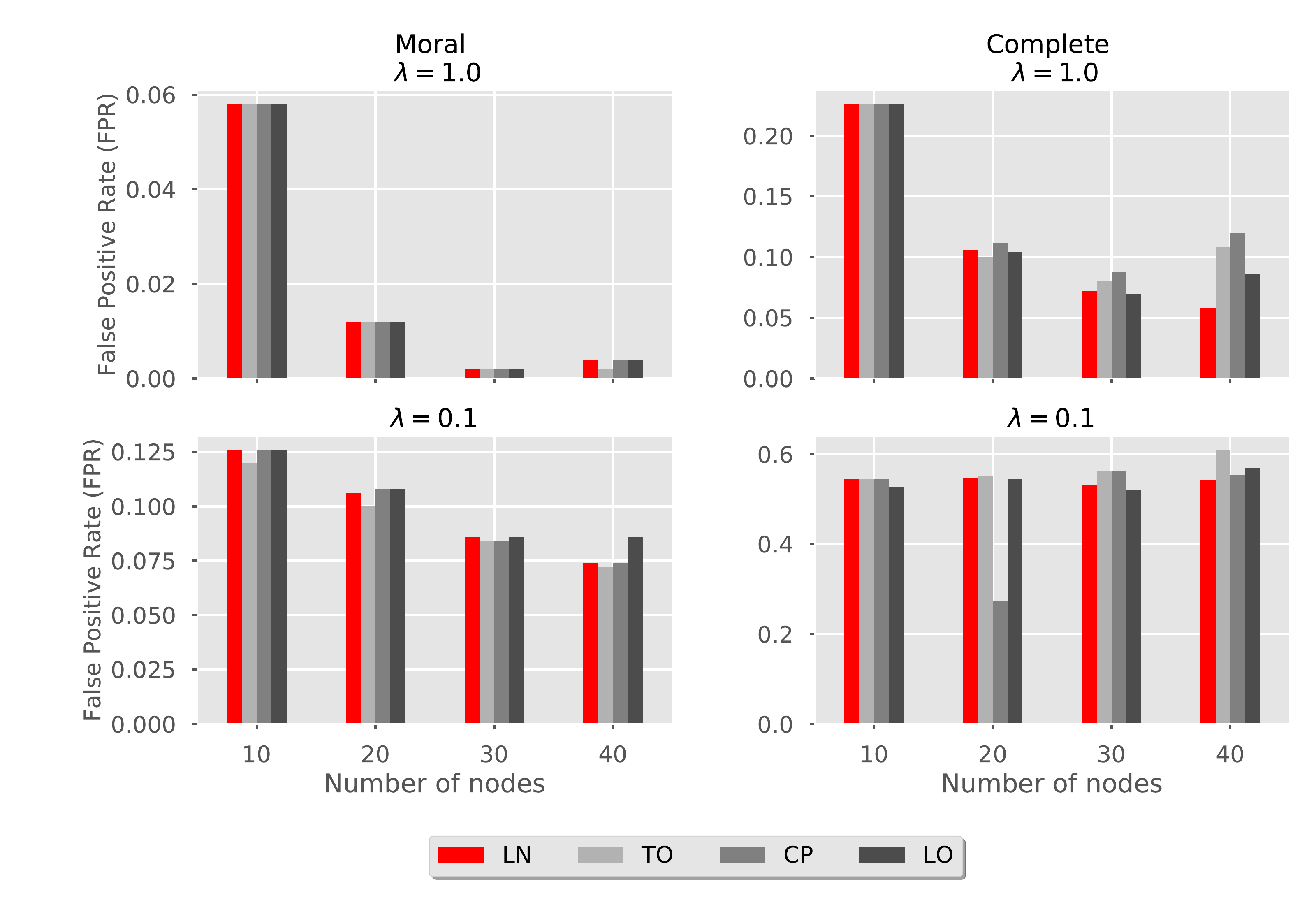}
		\caption{False Positive Rate (FPR) for MIQPs}
	\end{subfigure}
	
	\caption{Graph metrics for MIQPs for $\ell_1$ regularization. Plots a, c (left) show graph metrics with the number of samples $n=1000$. Plots b, d (right) show the graph metrics with the number of samples $n=100$.}
	\label{Figure: Graph_L1_100}
\end{figure*}

\restoregeometry

\end{document}